\documentclass[10pt,twocolumn,letterpaper]{article}

\usepackage{cvpr}              

\usepackage[utf8]{inputenc} 
\usepackage[T1]{fontenc}    
\usepackage{url}            
\usepackage{booktabs}       
\usepackage{amsfonts}       
\usepackage{nicefrac}       
\usepackage{microtype}      
\usepackage{graphicx}
\usepackage{amsmath}
\usepackage{amssymb}
\usepackage{booktabs}
\usepackage{multirow}
\usepackage{bm}
\usepackage{makecell}
\usepackage{CJKutf8}
\usepackage{algpseudocode}
\usepackage{algorithm}

\usepackage[pagebackref,breaklinks,colorlinks]{hyperref}

\usepackage[capitalize]{cleveref}
\crefname{section}{Sec.}{Secs.}
\Crefname{section}{Section}{Sections}
\Crefname{table}{Table}{Tables}
\crefname{table}{Tab.}{Tabs.}

\makeatletter
\def\blfootnote{\gdef\@thefnmark{}\@footnotetext}
\makeatother

\def\cvprPaperID{10837} 
\def\confName{CVPR}
\def\confYear{2023}

\begin{document}

\title{ERNIE-ViLG~2.0: Improving Text-to-Image Diffusion Model with \\ Knowledge-Enhanced Mixture-of-Denoising-Experts}

\author{
Zhida Feng$^{1,2,}$\thanks{Equal Contribution.}\;, Zhenyu Zhang$^{1,*}$, Xintong Yu$^{1,*}$, Yewei Fang$^{1}$, Lanxin Li$^{1}$, Xuyi Chen$^{1}$, Yuxiang Lu$^{1}$, \\ Jiaxiang Liu$^{1}$, Weichong Yin$^{1}$, Shikun Feng$^{1}$, Yu Sun$^{1}$, Li Chen$^{2}$, Hao Tian$^{1}$, Hua Wu$^{1}$, Haifeng Wang$^{1}$ \\
$^{1}$Baidu Inc.\\
$^{2}$School of Computer Science and Technology, Wuhan University of Science and Technology \\
{\tt\small \{fengzhida, zhangzhenyu07, yuxintong\}@baidu.com } \\
{\tt\small \{liujiaxiang, yinweichong, fengshikun01, sunyu02\}@baidu.com } \\
}

\twocolumn[{%
\renewcommand\twocolumn[1][]{#1}%
\maketitle
\begin{CJK*}{UTF8}{gbsn}
\begin{center}
    \centering
    \setlength{\tabcolsep}{1pt}
    \captionsetup{type=figure}
    \vspace{-1em}
    \begin{tabular}{cccc}
    \includegraphics[width=0.24\linewidth]{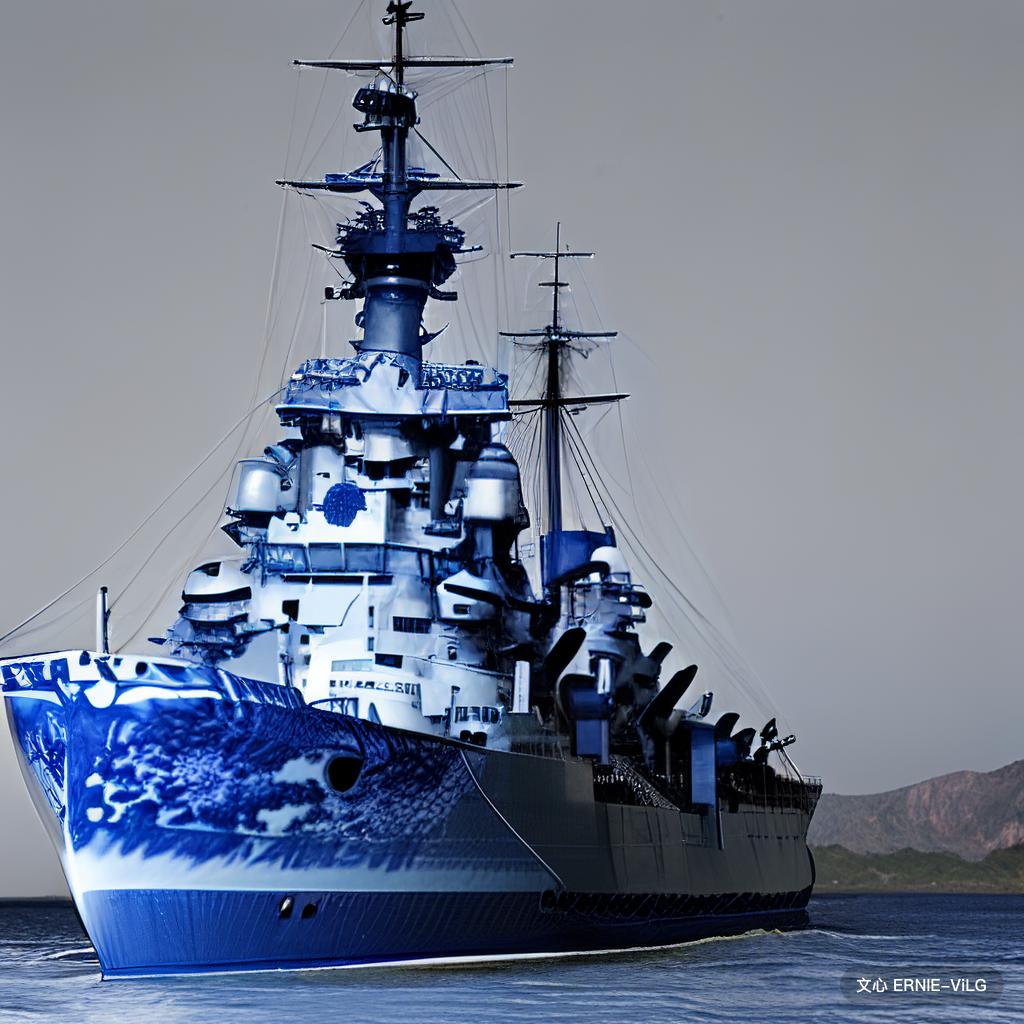} &
    \includegraphics[width=0.24\linewidth]{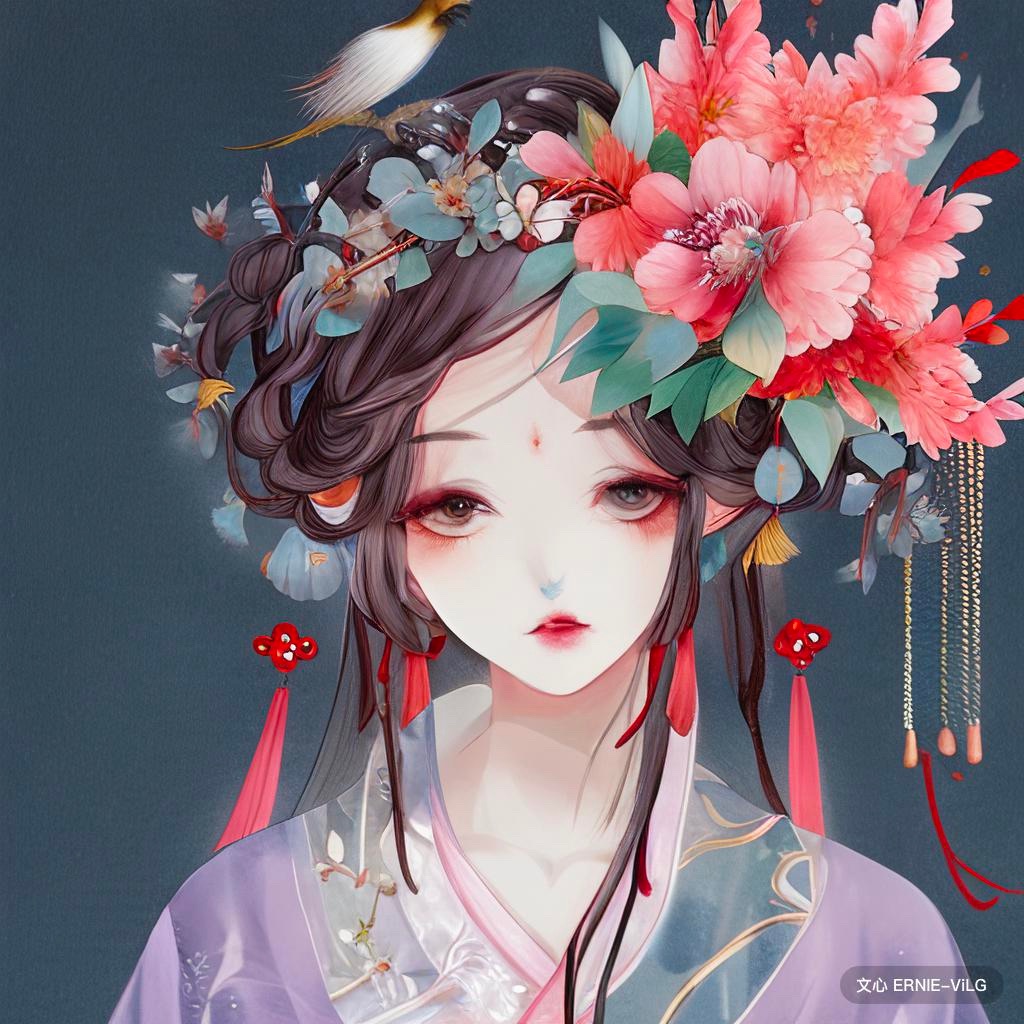} &
    \includegraphics[width=0.24\linewidth]{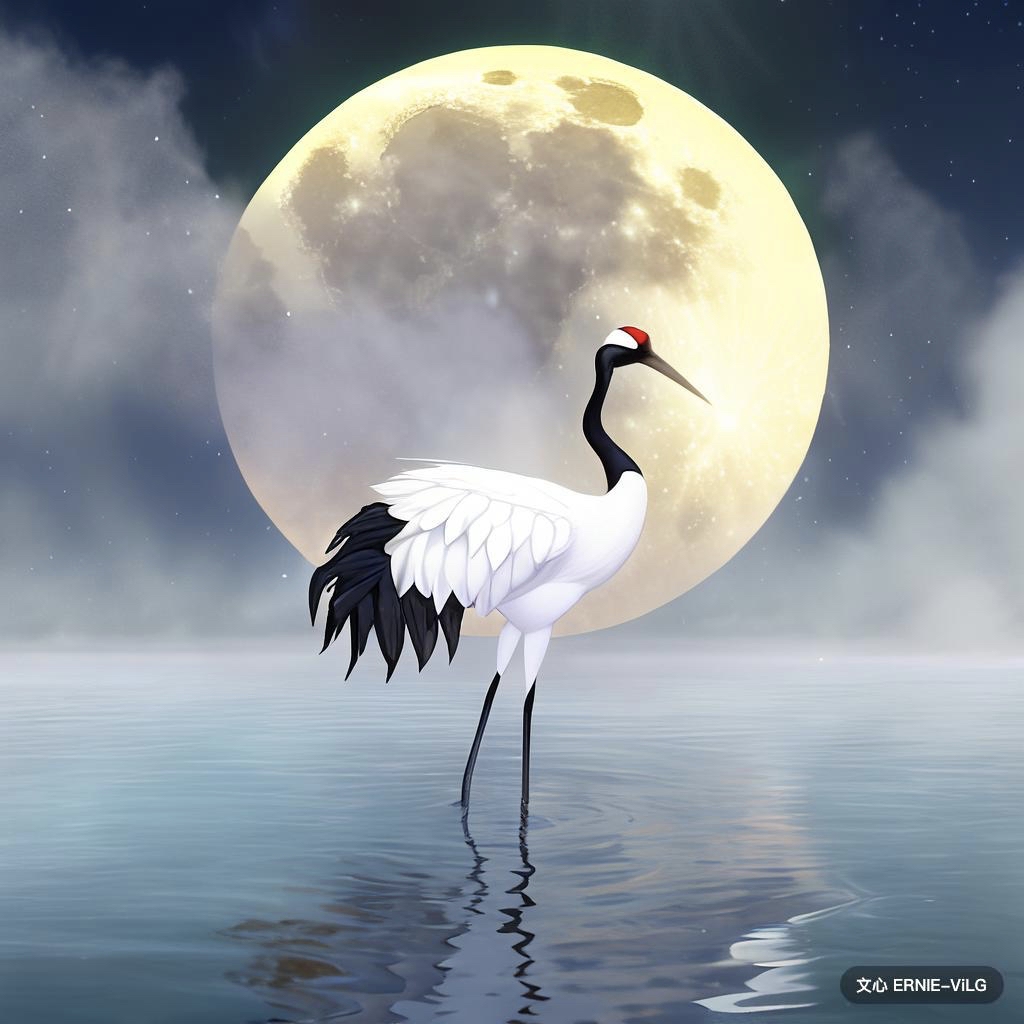}  &
    \includegraphics[width=0.24\linewidth]{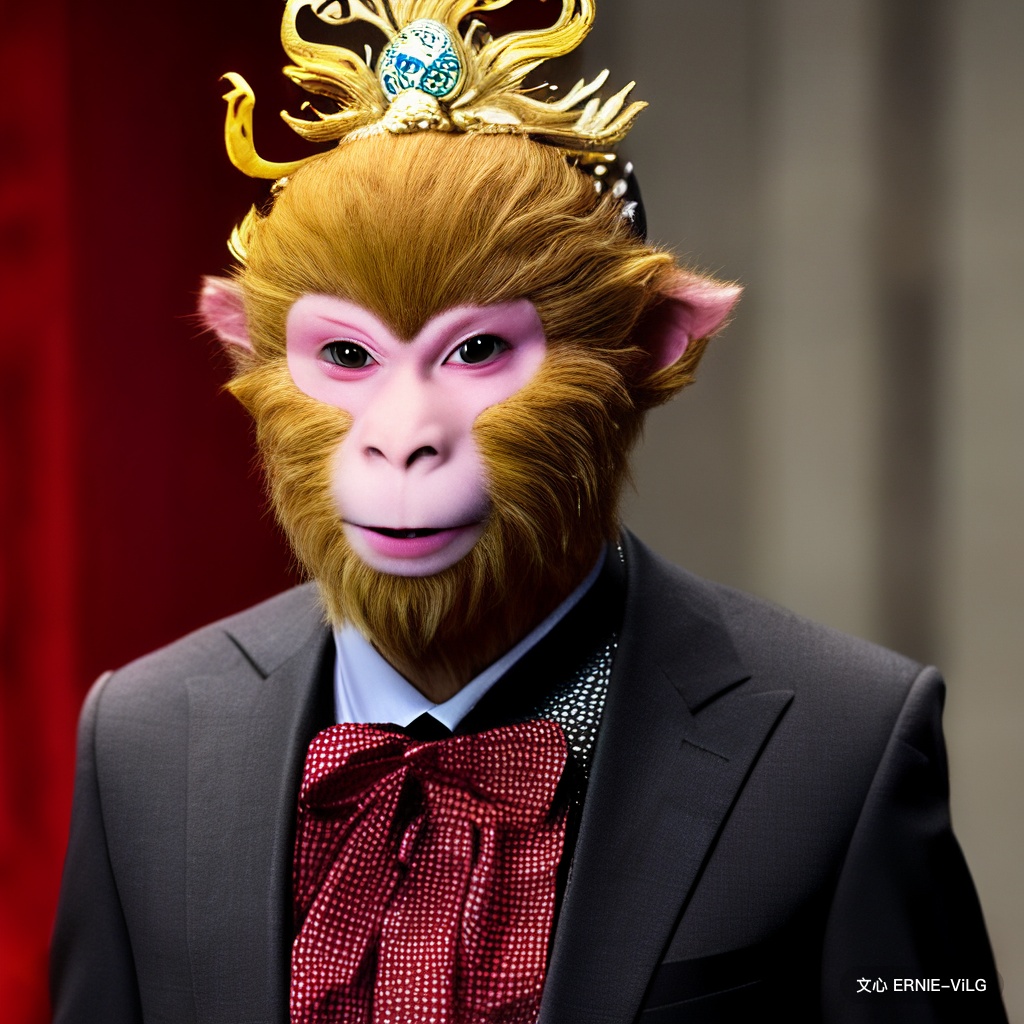} \\
    \scriptsize \makecell{一艘青花瓷质感的战舰} &
    \scriptsize \makecell{精致面容的古风少女，\\ 头戴百鸟鲜花头冠} &
    \scriptsize \makecell{一只仙鹤站在平静的湖面上，\\ 后面有一轮云雾缭绕的明月} &
    \scriptsize \makecell{穿西装的孙悟空的特写镜头} \\
    \scriptsize \makecell{A warship with blue and white \\ porcelain texture} &
    \scriptsize \makecell{An ancient style girl with delicate face, \\ wearing a crown of birds and flowers} &
    \scriptsize \makecell{A crane is standing on the calm lake,  \\ with a bright moon surrounded by \\ clouds in the background} &
    \scriptsize \makecell{The close up of Sun Wukong \\ in a suit} \\
    \includegraphics[width=0.24\linewidth]{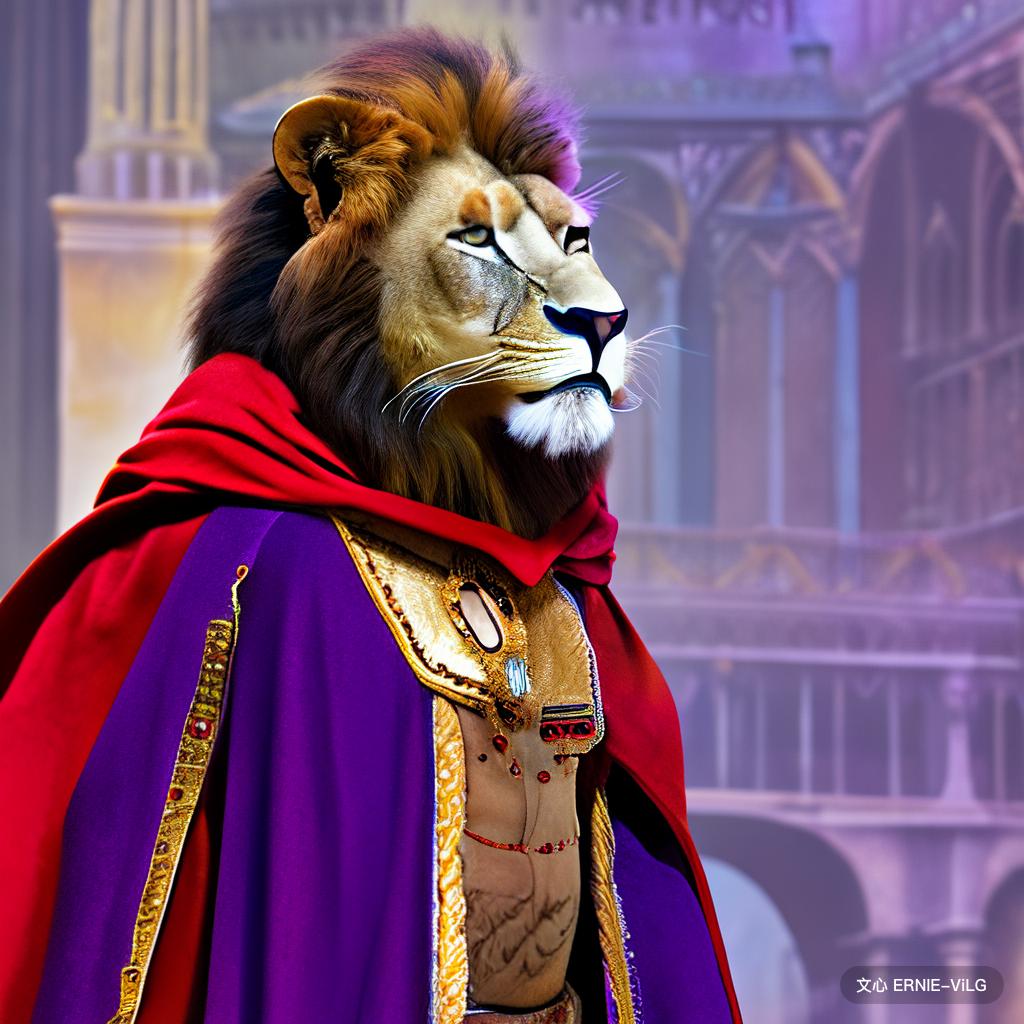} &
    \includegraphics[width=0.24\linewidth]{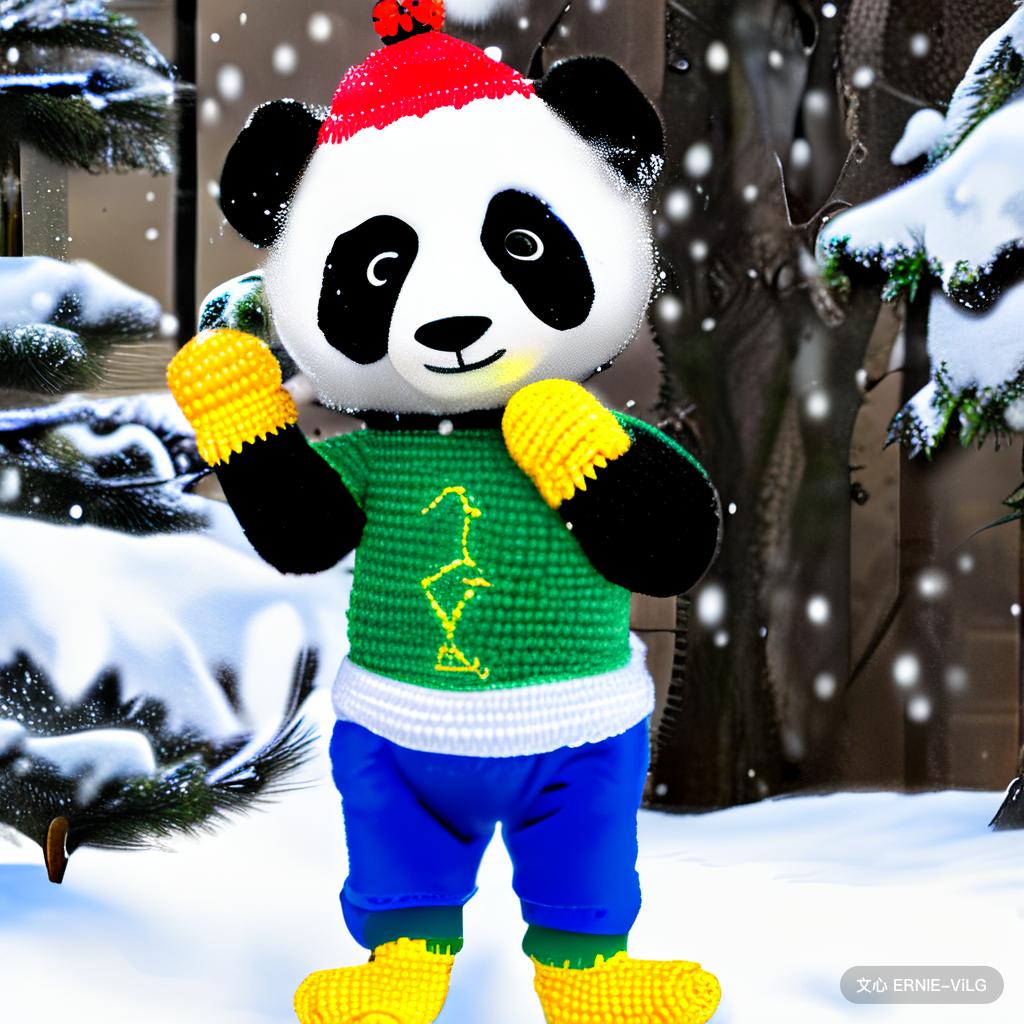} &
    \includegraphics[width=0.24\linewidth]{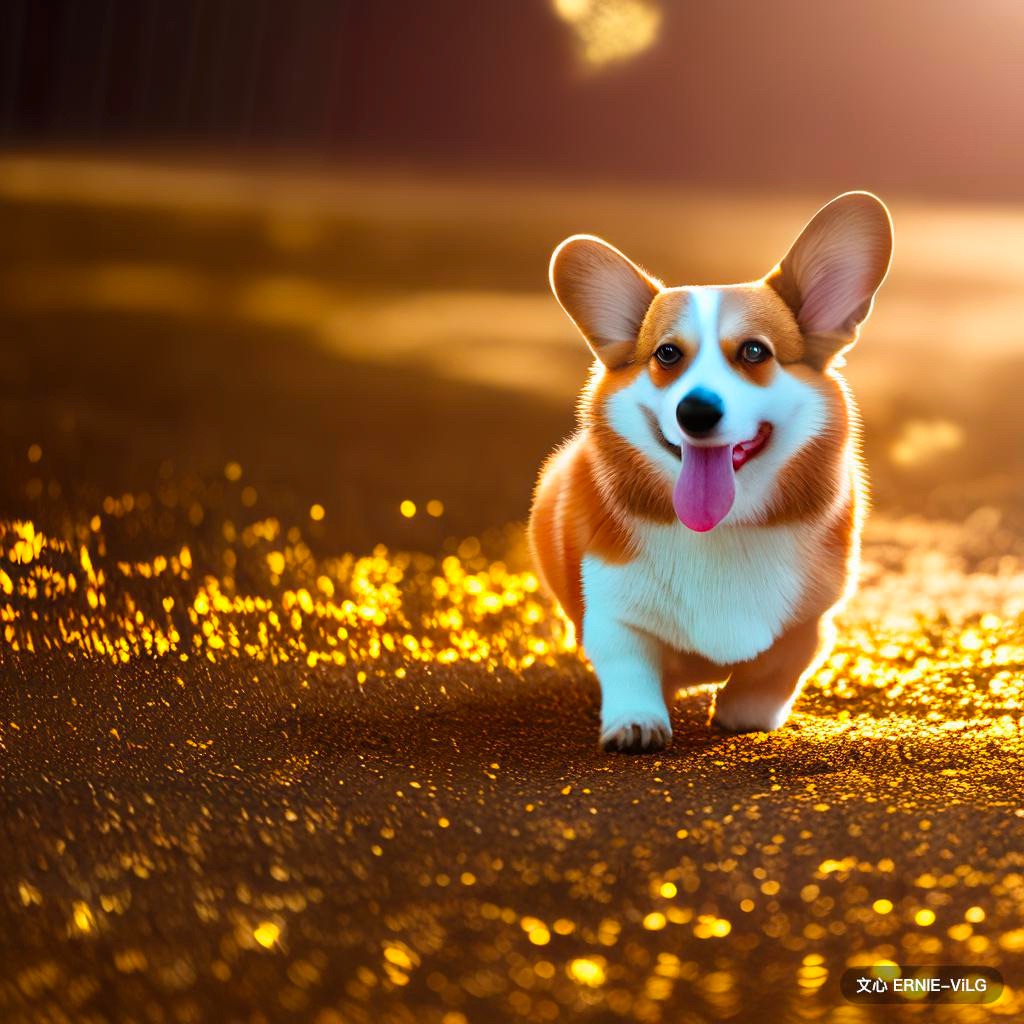}  &
    \includegraphics[width=0.24\linewidth]{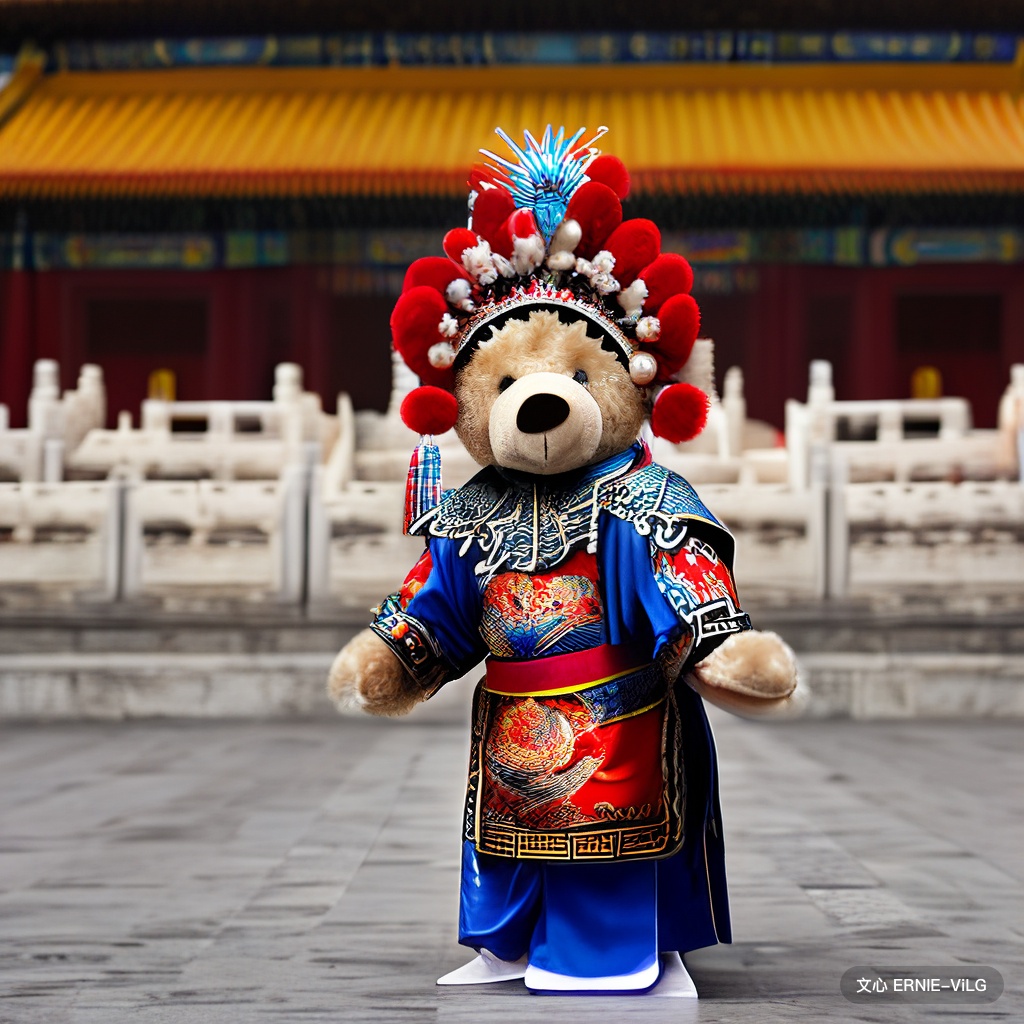}  \\
    \scriptsize \makecell{狮子王身穿紫色皇家大衣，身披红 \\ 色皇家斗篷， 正在发表史诗级的演说} &
    \scriptsize \makecell{雪地里的熊猫宝宝戴着红帽子、\\ 黄手套，穿着绿毛衣和蓝裤子} &
    \scriptsize \makecell{一只欢快的柯基行走在夕阳里，\\ 光从右边打来 铺成一地金黄} &
    \scriptsize \makecell{泰迪熊穿着戏服，站在太和殿前唱京剧} \\
    \scriptsize \makecell{The Lion King, wearing a purple \\ royal coat and a red royal cloak, \\ is delivering an epic speech} &
    \scriptsize \makecell{A baby panda in the snow, \\ wearing red hat, yellow gloves, \\ green sweater and blue pants} &
    \scriptsize \makecell{A cheerful Koji is walking in the sunset,\\ the golden light  comes from the right side \\ and spreads all over the ground} &
    \scriptsize \makecell{A teddy bear, wearing a costume, is \\ standing in front of the Hall of Supreme \\ Harmony and singing Beijing opera} \\
    \end{tabular}
    \captionof{figure}{Selected $1024 \times 1024$ samples with various text inputs, which shows that ERNIE-ViLG~2.0 has powerful capabilities of fine-grained semantic control and high-resolution image synthesis, as well as to produce high-quality creative images of different styles.}
    \label{fig:showcase}
\end{center}%
\end{CJK*}
}]

\blfootnote{\hspace{-2em}* denotes equal contribution. Work done during Feng's internship at Baidu.}

\begin{abstract}
Recent progress in diffusion models has revolutionized the popular technology of text-to-image generation.
While existing approaches could produce photorealistic high-resolution images with text conditions, there are still several open problems to be solved, which limits the further improvement of image fidelity and text relevancy.
In this paper, we propose ERNIE-ViLG~2.0, a large-scale Chinese text-to-image diffusion model, to progressively upgrade the quality of generated images by:
(1) incorporating fine-grained textual and visual knowledge of key elements in the scene,
and (2) utilizing different denoising experts at different denoising stages.
With the proposed mechanisms, ERNIE-ViLG~2.0\footnote{\url{https://wenxin.baidu.com/ernie-vilg}} not only achieves a new state-of-the-art on MS-COCO with zero-shot FID-30k score of 6.75, but also significantly outperforms recent models in terms of image fidelity and image-text alignment, with side-by-side human evaluation on the bilingual prompt set ViLG-300.
\end{abstract}

\section{Introduction}
\label{sec:intro}

Recent years have witnessed incredible progress in text-to-image generation.
With large-scale training data and model parameters, kinds of text-to-image generation models are now able to vividly depict the visual scene described by a text prompt, and enable anyone to create exquisite images without sophisticated drawing skills.
Among all types of image generation approaches, diffusion models~\cite{DBLP:conf/nips/HoJA20} are attracting increasing attention due to their ability to produce highly photorealistic images conditioned on text prompts.
Given a text prompt, the models transform a Gaussian noise into an image that conforms to the prompt through iterative denoising steps.
In the past years, text-to-image diffusion models such as LDM~\cite{DBLP:journals/corr/abs-2112-10752},  GLIDE~\cite{DBLP:conf/icml/NicholDRSMMSC22},  DALL-E~2~\cite{DBLP:journals/corr/abs-2204-06125}, and Imagen~\cite{DBLP:journals/corr/abs-2205-11487} have achieved impressive performance in both text relevancy and image fidelity. 
Despite these advances, the exploration of diffusion models by existing methods is still at the initial stage.
When we go deep into the principle and implementation of text-to-image diffusion models, there are still many opportunities to improve the quality of generated images further.

First, during the learning process of each denoising step, all text tokens interact with image regions and all the image regions contribute equally to the final loss function.
However, a visual scene of text and image contains many elements (i.e., textual words and visual objects), and different elements usually hold different importance for the expression of the scene semantics~\cite{DBLP:conf/aaai/0010TYSTW021}.
The indiscriminate learning process may cause the model to miss some key elements and interactions in the scene, thus facing the risk of text-image misalignment, such as the attribute confusion problem, especially for text prompts containing multiple objects with specific attributes~\cite{DBLP:journals/corr/abs-2204-06125}.
Second, when opening the horizon from individual step to the whole denoising process, we can found that the requirements of different denoising stages are also not identical.
In the early stages, the input images are highly noised, and the model is required to outline the semantic layout and skeleton out of almost pure noise. By contrast, in the later steps close to the image output, denoising mainly means improving the details based on an almost completed image~\cite{DBLP:journals/corr/abs-2112-10752}.
In practice, existing models usually use one U-Net for all steps, which means that the same set of parameters has to learn different denoising capabilities.

In this paper, we propose ERNIE-ViLG~2.0, an improved text-to-image diffusion model with knowledge-enhanced mixture-of-denoising-experts, to incorporate extra knowledge about the visual scene and decouple the denoising capabilities in different steps.
Specifically, we employ a text parser and an object detector to extract key elements of the scene in the input text-image pair, and then guide the model to pay more attention to their alignment in the learning process, so as to hope the model could handle the relationships among various objects and attributes.
Moreover, we divide the denoising steps into several stages and employ specific denoising ``experts'' for each stage.
With the mixture of multiple experts, the model can involve more parameters and learn the data distribution of each denoising stage better, without increasing the inference time, as only one expert is activated in each denoising step.

With the extra knowledge from the visual scene and the mixture-of-denoising-experts mechanism, we train ERNIE-ViLG~2.0 and scale up the model size to 24B parameters.
Experiments on MS-COCO show that our model exceeds previous text-to-image models by setting a new state-of-the-art of 6.75 zeros-shot FID-30k score, and detailed ablation studies confirm the contributions of each proposed strategy. 
Apart from automatic metrics, we also collect 300 bilingual text prompts that could assess the quality of generated images from different aspects and enable a fair comparison between English and Chinese text-to-image models.
The human evaluation results again indicate that ERNIE-ViLG~2.0 outperforms other recent methods, including DALL-E~2~\cite{DBLP:journals/corr/abs-2204-06125} and Stable Diffusion~\cite{DBLP:journals/corr/abs-2112-10752}, by a significant margin both in terms of image-text alignment and image fidelity.

\begin{figure*}[t]
    \centering
        \includegraphics[width=0.9\linewidth]{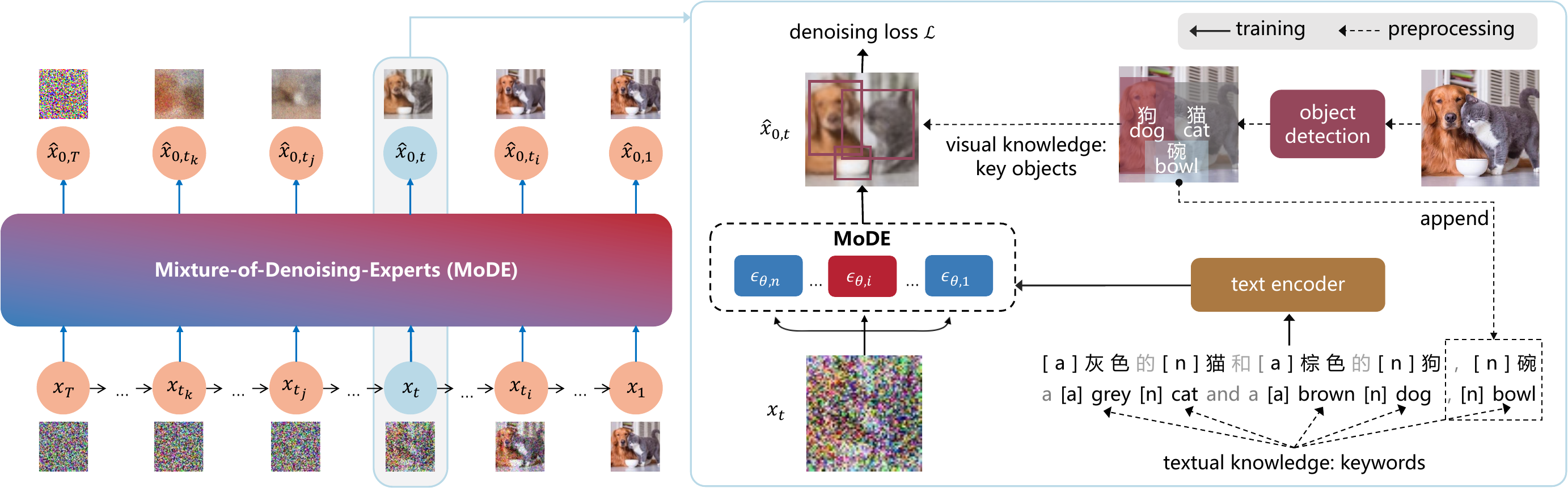}
    \caption{The architecture of ERNIE-ViLG~2.0, which incorporates fine-grained textual and visual knowledge of key elements in the scene and utilizes different denoising experts at different denoising stages.}
    \label{fig:model_structure}
\end{figure*}

To sum up, the main contributions of this work are: (1) We incorporate textual and visual knowledge into the text-to-image diffusion model, which effectively improves the ability of fine-grained semantic control and alleviates the problem of object-attribute mismatching in generated images. (2) We propose the mixture-of-denoising-experts mechanism to refine the denoising process, which can adapt to the characteristics of different denoising steps and scale up the model to 24B parameters, making it the largest text-to-image model at present. (3) ERNIE-ViLG~2.0 achieves the state-of-the-art zero-shot FID-30k score of 6.75 on MS-COCO, surpasses DALL-E~2 and Stable Diffusion in human evaluation on the Chinese-English bilingual prompt set ViLG-300.

\section{Method}

During the training process, the text-to-image diffusion model receives paired inputs $(x, y)$ consisting of an image $x$ with its text description $y$, and the ultimate goal is to generate $x$ based on $y$.
To achieve this, a text encoder $f_\theta(\cdot)$ first encodes $y$ as $f_\theta(y)$, then a denoising network $\epsilon_\theta(\cdot)$ conditioned on $f_\theta(y)$ learns to generate $x$ from a Gaussian noise.
To help the model generate high-quality images that accurately match the text description (i.e., text prompt), ERNIE-ViLG~2.0 enhances text encoder $f_\theta(\cdot)$ and denoising network $\epsilon_\theta(\cdot)$ with textual and visual knowledge of key elements in the scene.
Furthermore, ERNIE-ViLG~2.0 employs mixture-of-denoising-experts to refine the image generation process, where different experts are responsible for different generation steps in the denoising process.
The overall architecture of ERNIE-ViLG~2.0 is shown in Figure~\ref{fig:model_structure} and the details are described in the following subsections.

\subsection{Preliminary}
\label{sec:preliminary}

Denoising diffusion probabilistic models (DDPM) are a class of score-based generative models that have recently shown delightful talents in the field of text-to-image generation~\cite{DBLP:conf/nips/HoJA20}.
The diffusion process of DDPM aims to iteratively add diagonal Gaussian noise to the initial data sample $x$ and turn it into an isotropic Gaussian distribution after $T$ steps:
\begin{equation}
\small
    x_t = \sqrt{\alpha_t} x_{t-1} + \sqrt{1-\alpha_t} \epsilon_t, \quad t \in \{1,\dots,T\}
\end{equation}
where the sequence $\{x_t\}$ starts with $x_0 = x$ and ends with $x_T \sim \mathcal{N}(0,I)$, the added noise at each step is $\epsilon_t \sim \mathcal{N}(0,I)$, and $\{\alpha_t\}_{1 \dots T}$ is a pre-defined schedule~\cite{DBLP:conf/icml/Sohl-DicksteinW15,DBLP:conf/iclr/0011SKKEP21}.
The denoising process is the reverse of diffusion, which converts the Gaussian noise $x_T \sim \mathcal{N}(0,I)$ back into the data distribution $x_0$ through iterative denoising steps $t = T,\dots, 1$. 
During training, for a given image $x$, the model calculates $x_t$ by sampling a Gaussian noise $\epsilon \sim \mathcal{N}(0,I)$:
\begin{equation}
\label{eq:x0_to_xt}
\small
    x_t = \sqrt{\bar{\alpha}_t} x_0 + \sqrt{1-\bar{\alpha}_t} \epsilon,
\end{equation}
where $\bar{\alpha}_t = \prod_{s=1}^t\alpha_s$. Given $x_t$, the target of the denoising network $\epsilon_\theta(\cdot)$ is to restore $x_0$ by predicting the noise $\epsilon$. It is learned via the loss function
\begin{equation}
\label{eq:loss}
\small
    \mathcal{L} = \mathbb{E}_{x, \epsilon \sim \mathcal{N}(0,I), t} 
    \left[ || \epsilon - \epsilon_\theta(x_t, t) ||_2^2 \right].
\end{equation}

With the predicted $\epsilon_\theta(x_t, t)$, we can have the prediction of $x_0$ at step $t$ by converting Equation~(\ref{eq:x0_to_xt}):
\begin{equation}
\small
    \hat{x}_{0,t} = \frac{1}{\sqrt{\bar{\alpha}_t}}(x_t - \sqrt{1-\bar{\alpha}_t} \epsilon_\theta(x_t, t)).
\end{equation}
In Figure~\ref{fig:model_structure}, we visualize the sampled $x_t$ and the predicted $\hat{x}_{0,t}$ for several timesteps during training.
In the inference process of DDPM, $x_0$ is unknown, so the model iteratively generates $x_{t-1}$ based on $x_t$ and $\hat{x}_{0,t}$:
\begin{equation}
\small
\begin{aligned}
    x_{t-1} = &\frac{ 1 - \bar{\alpha}_{t-1}}{1 - \bar{\alpha}_t} \sqrt{\alpha_t} x_t + 
    \frac{1 - \alpha_t }{1 - \bar{\alpha}_t} \sqrt{\bar{\alpha}_{t-1}} \hat{x}_{0,t} \\
    &+ \sqrt{\frac{(1 - \bar{\alpha}_{t-1})(1-\alpha_t)}{1 - \bar{\alpha}_t}} \epsilon_t', 
    \quad  t \in \{T, \dots, 1\},
\end{aligned}
\end{equation}
where $\epsilon_t' \sim \mathcal{N}(0,I)$ is a sampled Gaussian noise.

The denoising network $\epsilon_\theta(\cdot)$ is typically implemented by U-Net~\cite{DBLP:conf/nips/HoJA20}. 
To allow $\epsilon_\theta(\cdot)$ to condition on text prompts, a text encoder $f_\theta(\cdot)$ first extracts the text representation $f_\theta(y) \in \mathbb{R}^{n_y \times d_y}$, which is then fed into $\epsilon_\theta(\cdot)$ via a cross-modal attention layer~\cite{DBLP:conf/icml/NicholDRSMMSC22}.
Formally, the U-Net representation $\varphi_i(x_t) \in \mathbb{R}^{n_x \times d}$ is concatenated with the text representation $f_\theta(y)$ after projection, and then goes through an attention layer to achieve cross-modal interaction, 
\begin{equation}
\small
\begin{aligned}
    Q &= \varphi_i(x_t) W_Q^{(i)}  , \\
    K &= [\varphi_i(x_t) W_{K_x}^{(i)} ; f_\theta(y) W_{K_y}^{(i)}], \\
    V &= [\varphi_i(x_t) W_{V_x}^{(i)} ; f_\theta(y) W_{V_y}^{(i)}],
\end{aligned}
\end{equation}
\begin{equation}
\label{eq:attention}
\small
    {\rm Attention}(Q, K, V) = {\rm softmax} \left( \frac{QK^\top}{\sqrt{d}} \right) V,
\end{equation}
where $i$ is the index for U-Net layers, $[;]$ is the concatenation operator, $W_Q^{(i)}, W_{K_x}^{(i)}, W_{V_x}^{(i)} \in \mathbb{R}^{d \times d}$ and $ W_{K_y}^{(i)}, W_{V_y}^{(i)} \in \mathbb{R}^{d_y \times d} $ are learnable projection layers, $n_x$ and $n_y$ are the length of encoded image and text, respectively.

During inference, given a text prompt $y$, the denoising U-Net $\epsilon_\theta(\cdot)$ predicts the image sample $x$ conditioned on the text $y$ with classfier-free guidance~\cite{DBLP:journals/corr/abs-2207-12598} and Denoising Diffusion Implicit Models (DDIM) sampling~\cite{DBLP:conf/iclr/SongME21}.

\subsection{Knowledge-Enhanced Diffusion Model}

The text-to-image model receives a text prompt that describes the scene of an image, then depicts it with crucial objects and corresponding attribute details.
In other words, both text and image are intended to express a visual scene, in which key elements have different expressions, such as keywords in text or salient regions in image.
However, naive diffusion model does not distinguish the importance of elements and indiscriminately iterates the denoising process. ERNIE-ViLG~2.0 incorporates extra text and visual knowledge into the learning stage, hoping to enhance the fine-grained semantic perception of diffusion model.

\noindent\textbf{Textual Knowledge.}
An ideal text-to-image model is expected to focus on all the critical semantics mentioned in the text prompt.
To distinguish function words and words describing key semantics, we adopt an off-the-shelf part-of-speech toolkit to extract lexical knowledge of the input prompt, and then improve the learning process by (1) inserting special tokens into the input sequence and (2) increasing the weight of tokens with specific part-of-speech tags in the attention layer.
Specifically, we selected 50\% of samples and inserted special tokens at the beginning of each word, in which each part-of-speech tag corresponds to a special token.
For the selected samples, we also strengthen the attention weight of keywords based on the lexical analysis results. 
In this way, Equation~(\ref{eq:attention}) is modified to,
\begin{equation}
\small
    {\rm Attention}(Q, K, V)' = {\rm softmax} \left( \frac{W_a \cdot (QK^{\top})}{\sqrt{d}} \right) V,
\end{equation}
where $W_a \in \mathbb{R}^{n_x \times (n_x+n_y)}$ is an auxiliary weight matrix that used to scale the vanilla attention, and
\begin{equation}
\small
    W_a^{ij} = \left\{ \begin{array}{cl} 1+w_a & { tok_i \in \{x\}, tok_j \in \{x, y_{key}\}} \\ 1 & {\rm otherwise.}   
    \end{array}\right.
\end{equation}
Here $w_a^{ij}$ is the scaling factor of the attention weight between token ${tok_i}$ and ${tok_j}$, $w_a$ is a hyper-parameter, $x$ refers to all the image tokens, and $y_{key}$ denotes the keywords in text\footnote{The keywords is defined as notional words in modern Chinese (i.e., nouns, verbs, adjectives, numerals, quantifiers, and pronouns).}. Figure~\ref{fig:model_structure} gives an example, where special tokens ``\verb|[a]|'' and ``\verb|[n]|'' are inserted for adjectives and nouns, respectively.

\noindent\textbf{Visual Knowledge.}
Similar to notional words in the text prompt, there are also salient regions in an image, such as people, trees, buildings, and objects mentioned in the input. 
To extract such visual knowledge, we apply an object detector~\cite{DBLP:conf/cvpr/00010BT0GZ18} to 50\% of training samples, and then select eye-catching objects from the results with heuristic strategies.
Since the loss function of the diffusion model directly acts on the image space, we can assign higher weights to corresponding regions by modifying Equation~(\ref{eq:loss}), thus promoting the model to focus on the generation of these objects:
\begin{equation}
\small
    \mathcal{L}' = \mathbb{E}_{z, \epsilon \sim \mathcal{N}(0,I), t} 
    \left[W_l \cdot || \epsilon - \epsilon_\theta(z_t, t) ||_2^2 \right],
\end{equation}
\begin{equation}
\small
    W_l^{ij} = \left\{ \begin{array}{cl} 1+w_l & {los_{ij} \in \{x_{key}\} } \\ 1 & {\rm otherwise.}    
    \end{array}\right.
\end{equation}
Here $W_l \in \mathbb{R}^{n_{h} \times n_{w}}$ is the weight matrix, $n_h$ and $n_w$ are the height and weight of image space, $w_l$ is a hyper-parameter, $los_{ij}$ is the loss item in $i$-th row and $j$-th column of image space, $x_{key}$ is the regions that corresponding to key objects. 
As Figure~\ref{fig:model_structure} illustrates, the regions of ``\verb|dog|'' and ``\verb|cat|'' are assigned with larger weights in the calculation of $\mathcal{L}'$.

Now a new problem arises: as a kind of fine-grained knowledge, the selected objects may not appear in the text prompt, thus perplexing the model in learning the alignment between words and objects.
An intuitive idea is first to obtain the object and attribute category of each region, then combine corresponding class labels with the original text prompt to achieve fine-grained description, thus ensuring the input contains both coarse and fine granularity information.
For instance, as shown in Figure~\ref{fig:model_structure}, the detected object ``\verb|bowl|'' is not included in the caption, so we append it to the original description.
Besides, we also employ an image captioning model~\cite{DBLP:conf/icml/WangYMLBLMZZY22} to generate text for images, and randomly replace the original prompt with generated captions, because the generated captions of many images are more concise and reveal more accurate semantics than original prompts.

Most notably, the above strategies are only limited to the training stage. By randomly selecting a part of samples to equip these additional enhancement strategies, the model is supposed to sense the hints of knowledge from various perspectives, and generate higher quality images for the given text in the inference stage, even without special tokens, attention strengthening, or text refinement.

\subsection{Mixture-of-Denoising-Experts}

Recall that the diffusion process is to iteratively corrupt the image with Gaussian noises by a series of diffusion steps $t = 1,\dots,T$, and DDPM~\cite{DBLP:conf/nips/HoJA20} are trained to revert the diffusion process by denoising steps $t = T,\dots,1$.
During the denoising process, all steps aim to denoise a noised input, and they together convert a completely random noise into a meaningful image gradually.
Although sharing the same goal, the difficulty of these denoising steps varies according to the noise ratio of input.
Figure~\ref{fig:model_structure} illustrates such difference by presenting some examples of $x_t$ and the denoising prediction $\hat{x}_{0,t}$ during training. For timesteps $t$ near $T$, such as $t=T, t_k, t_j$ in the figure, the input of the denoising network $x_t$ is highly noised, and the network of these steps mainly tackles a generation task, i.e., generating an image from a noise.
On the contrary, for timesteps $t$ near $1$, such as $t=t_i, 1$, the input $x_t$ is close to the original image, and the network of these steps needs to refine the image details.

DDPM makes no specific assumption on the implementation of denoising network, that is, the denoising process does not require the same denoising network for all steps in theory.
However, most of the previous text-to-image diffusion approaches~\cite{DBLP:journals/corr/abs-2112-10752,DBLP:conf/icml/NicholDRSMMSC22,DBLP:journals/corr/abs-2204-06125,DBLP:journals/corr/abs-2205-11487} follow the vanilla implementation to adopt a denoising network for the whole denoising process.
Considering that tasks of different timesteps are different, we conjecture that using the same set of parameters to learn different tasks might lead to suboptimal performance.

In view of this, we further propose Mixture-of-Denoising-Experts (MoDE), where the primary motivation is to employ multiple specialized expert networks to fit different tasks at different timesteps.
Since the inputs of adjacent timesteps are similar and so are the denoising tasks, we divide all the timesteps uniformly into $n$ blocks ($S_1, \cdots ,S_i, \cdots ,S_n$), in which each block consists of consecutive timesteps and is assigned to one denoising expert. In other words, the timesteps in the same block are denoised by the same group of network parameters.
In practice, we share the same text encoder for all denoising experts, and utilize different U-Net experts for different timestep blocks:
\begin{equation}
\small
    \epsilon_\theta(x_t, t) = \{\epsilon_{\theta,i}(x_t, t)\}, \quad t \in S_i, 
\end{equation}
where $\epsilon_{\theta,i}(x_t, t)$ is the $i$-th expert network.
Herein, MoDE improves the model performance by adopting expert networks to specially deal with different denoising stages.

Intuitively, when using more experts, each block contains fewer timesteps, so each expert could better focus on learning the characteristics of specific denoising steps assigned to it.
Meanwhile, as only one expert network is activated at each step, increasing the number of experts does not affect the computation overhead during inference. 
Therefore, ERNIE-ViLG~2.0 can flexibly scale up the parameters of diffusion model, allowing the experts to fit the data distribution better without increasing inference time.

\section{Experiments}

In this section, we first introduce the implementation details of ERNIE-ViLG~2.0. Then we present the comparison of models with automatic metrics and human evaluation. Last, we further analyze the results with quantitative ablation studies and qualitative showcases.

\subsection{Implementation Details}

To reduce learning complexity, we use diffusion models to generate the representations of images in latent space of an image auto-encoder following Latent Diffusion Models~\cite{DBLP:journals/corr/abs-2112-10752}.
We first pre-train an image encoder to transform an image $x \in \mathbb{R}^{n_{h} \times n_{w} \times 3}$ from pixel space into latent space $ \hat{x} \in \mathbb{R}^{n_{h}^{l} \times n_{w}^{l} \times 4}$ and an image decoder to convert it back. Here $n_{h}$/$n_{h}^{l}$ and $n_{w}$/$n_{w}^{l}$ denote the image's original/latent height and width, and we collectively refer to pixel space and hidden space as image space in this paper.
Then we fix the auto-encoder and train the diffusion model to generate $\hat{x}$ from text prompt $y$.
During inference, we adopt the pre-trained image decoder to turn $\hat{x}$ into pixel-level image output.

ERNIE-ViLG~2.0 contains a transformer-based text encoder with 1.3B parameters and 10 denoising U-Net experts with 2.2B parameters each, which totally add up to about 24B parameters.
For hyper-parameters to incorporate knowledge, the attention weight scale $w_a$ is set to $0.01$ and the loss weight scale $w_l$ is set to $0.1$ (both chosen from [$0.01$, $0.1$, $0.5$, $1$]).
For the MoDE strategy, all timesteps are divided into $10$ blocks.
The model is optimized by AdamW~\cite{DBLP:conf/iclr/LoshchilovH19}, with a fixed learning rate $0.9 \times 10^{-4}$, ${\beta}_1=0.9$, ${\beta}_2=0.999$, and weight decay of $0.01$.
We train ERNIE-ViLG~2.0 on 320 Tesla A100 GPUs for 18 days.
More training details are introduced in Appendix~\ref{sec:training}.

The training data consists of 170M image-text pairs, including publicly available English datasets like LAION~\cite{DBLP:journals/corr/abs-2111-02114} and a series of internal Chinese datasets. The image auto-encoder is trained on the same set. For images with English captions, we translate them with Baidu Translate API\footnote{\url{https://fanyi.baidu.com}} to get the Chinese version.

\begin{table}[t]
  \centering\small
  \caption{Comparison of ERNIE-ViLG~2.0 and representative text-to-image generation models on MS-COCO $256 \times 256$ with zero-shot FID-30k. We use classifier-free guidance scale 2.1 for our diffusion model and achieve the best performance.}
    \begin{tabular}{lc}
    \toprule
    \textbf{Model} & Zero-Shot FID-30k $\downarrow$  \\
    \midrule
    DALL-E~\cite{DBLP:conf/icml/RameshPGGVRCS21}          & 27.50 \\
    CogView~\cite{DBLP:conf/nips/DingYHZZYLZSYT21}        & 27.10 \\ 
    LAFITE~\cite{DBLP:journals/corr/abs-2111-13792}       & 26.94 \\
    LDM~\cite{DBLP:journals/corr/abs-2112-10752}          & 12.61 \\
    ERNIE-ViLG~\cite{DBLP:journals/corr/abs-2112-15283}   & 14.70 \\
    GLIDE~\cite{DBLP:conf/icml/NicholDRSMMSC22}           & 12.24 \\
    Make-A-Scene~\cite{DBLP:journals/corr/abs-2203-13131} & 11.84 \\
    DALL-E~2~\cite{DBLP:journals/corr/abs-2204-06125}     & 10.39 \\
    CogView2~\cite{DBLP:journals/corr/abs-2204-14217}     & 24.00 \\
    Imagen~\cite{DBLP:journals/corr/abs-2205-11487}       & 7.27 \\
    Parti~\cite{DBLP:journals/corr/abs-2206-10789}        & 7.23 \\
    \midrule
    ERNIE-ViLG~2.0 & \textbf{6.75} \\
    \bottomrule
    \end{tabular}%
  \label{tab:main}%
\end{table}%

\subsection{Results}

\noindent\textbf{Automatic Evaluation on MS-COCO.}
Following previous work~\cite{DBLP:journals/corr/abs-2112-10752,DBLP:journals/corr/abs-2204-06125,DBLP:journals/corr/abs-2205-11487}, we evaluate ERNIE-ViLG~2.0 on MS-COCO $256 \times 256$ with zero-shot FID-30k, where 30,000 images from the validation set are randomly selected and the English captions are automatically translated to Chinese.

Table~\ref{tab:main} shows that ERNIE-ViLG~2.0 achieves new state-of-the-art  performance of text-to-image generation, with 6.75 zero-shot FID-30k on MS-COCO. 
Inspired by DALL-E~\cite{DBLP:conf/icml/RameshPGGVRCS21} and Parti~\cite{DBLP:journals/corr/abs-2206-10789},
we rerank the batch-sampled images (with only 4 images per text prompt, comparing with 512 images used in DALL-E and 16 images used in Parti) based on the image-text alignment score, calculated by a pre-trained CLIP model~\cite{DBLP:conf/icml/RadfordKHRGASAM21}, in which the text is the initial English caption in MS-COCO and the image is generated from the auto-translated Chinese caption.
Besides, even without the reranking strategy, we find that ERNIE-ViLG~2.0 can also beat the latest diffusion-based models like DALL-E~2~\cite{DBLP:journals/corr/abs-2204-06125} and Imagen~\cite{DBLP:journals/corr/abs-2205-11487}, with the zero-shot FID-30k of 7.23. 
See also Appendix~\ref{sec:detailed_fid} for a detailed comparison with regard to model sizes and reranking strategies.

\begin{figure}
  \centering
  \begin{subfigure}{0.49\linewidth}
    \includegraphics[width=\linewidth]{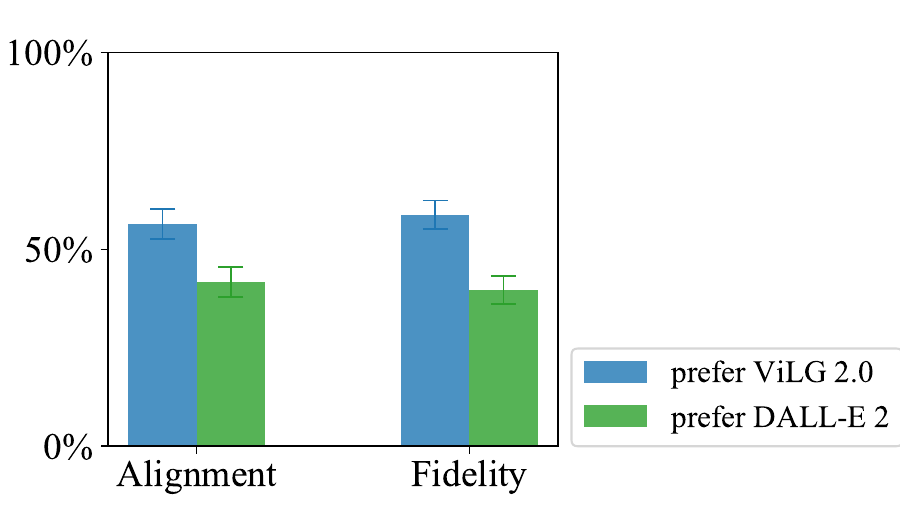}
    \caption{v.s. DALL-E~2.}
    \label{fig:human_dalle}
  \end{subfigure}
  \begin{subfigure}{0.49\linewidth}
    \includegraphics[width=\linewidth]{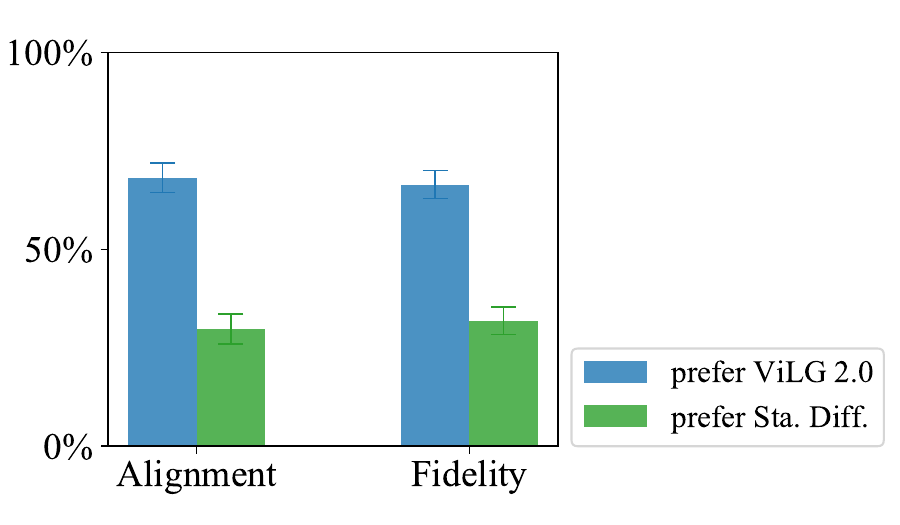}
    \caption{v.s. Stable Diffusion.}
    \label{fig:human_sd}
  \end{subfigure}
    \caption{Comparison of ERNIE-ViLG~2.0 and DALL-E~2/Stable Diffusion on ViLG-300 with human evaluation. We report the user preference rates with 95\% confidence intervals.}
  \label{fig:human}
\end{figure}

\begin{CJK*}{UTF8}{gbsn}
\begin{figure}[t]
    \centering
    \setlength{\tabcolsep}{2pt}
    \begin{tabular}{cccccccccc}
        \rotatebox{90}{\scriptsize\phantom{AA} ViLG~2.0} &
        \includegraphics[width=0.2\linewidth]{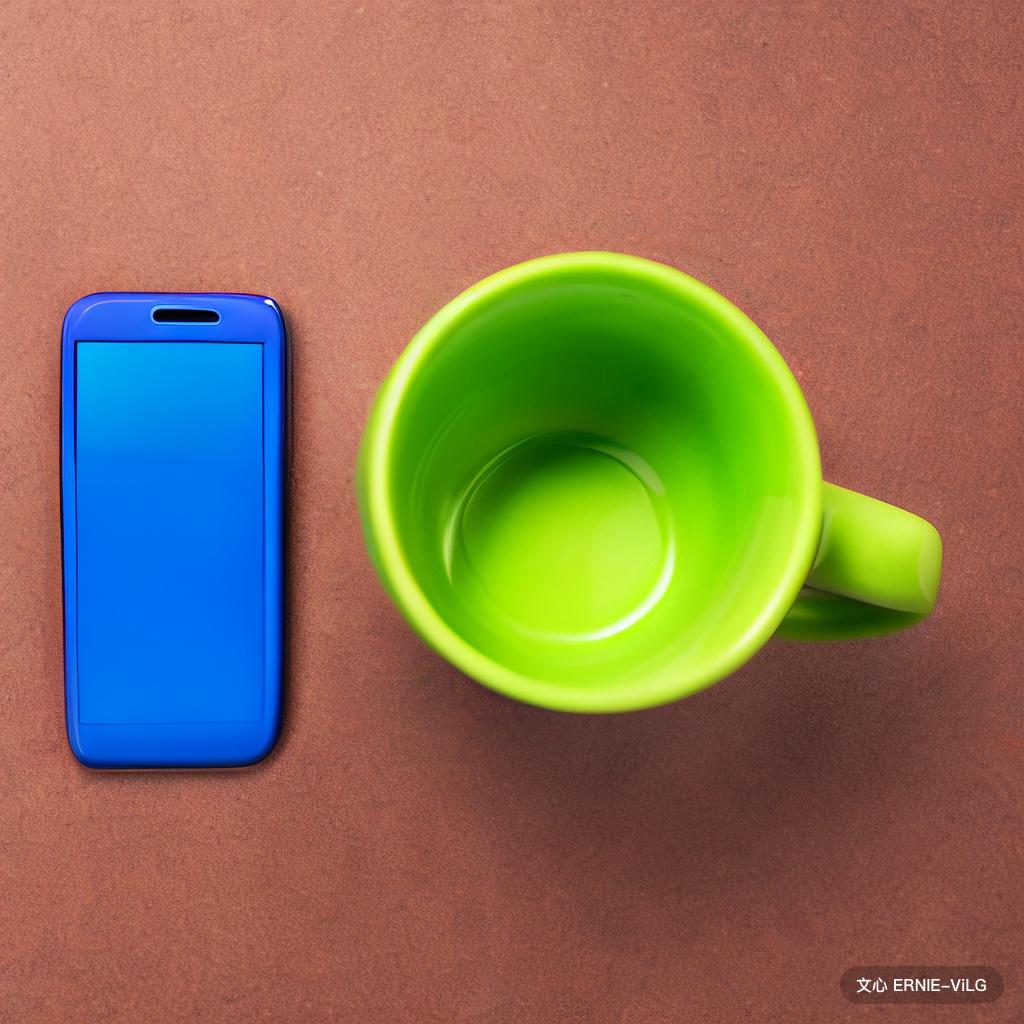} &
        \includegraphics[width=0.2\linewidth]{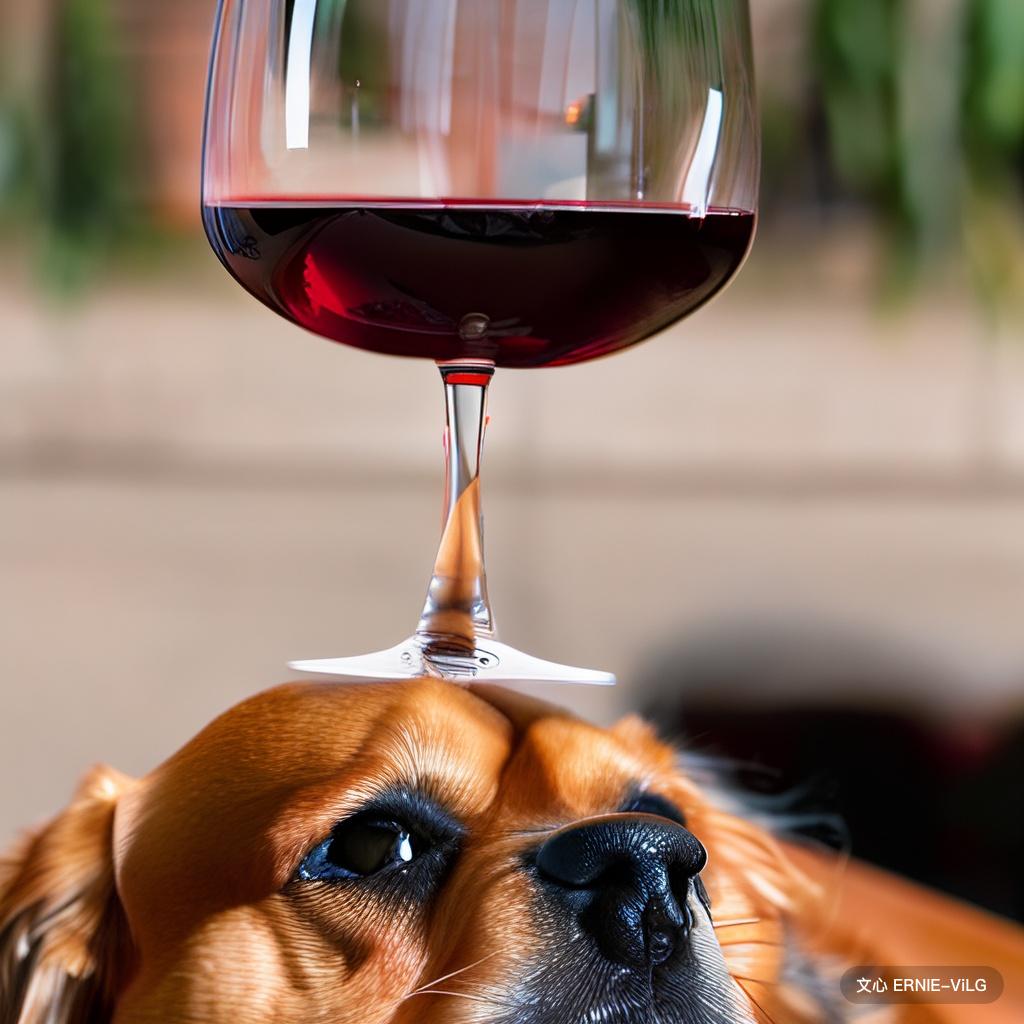} &
        \includegraphics[width=0.2\linewidth]{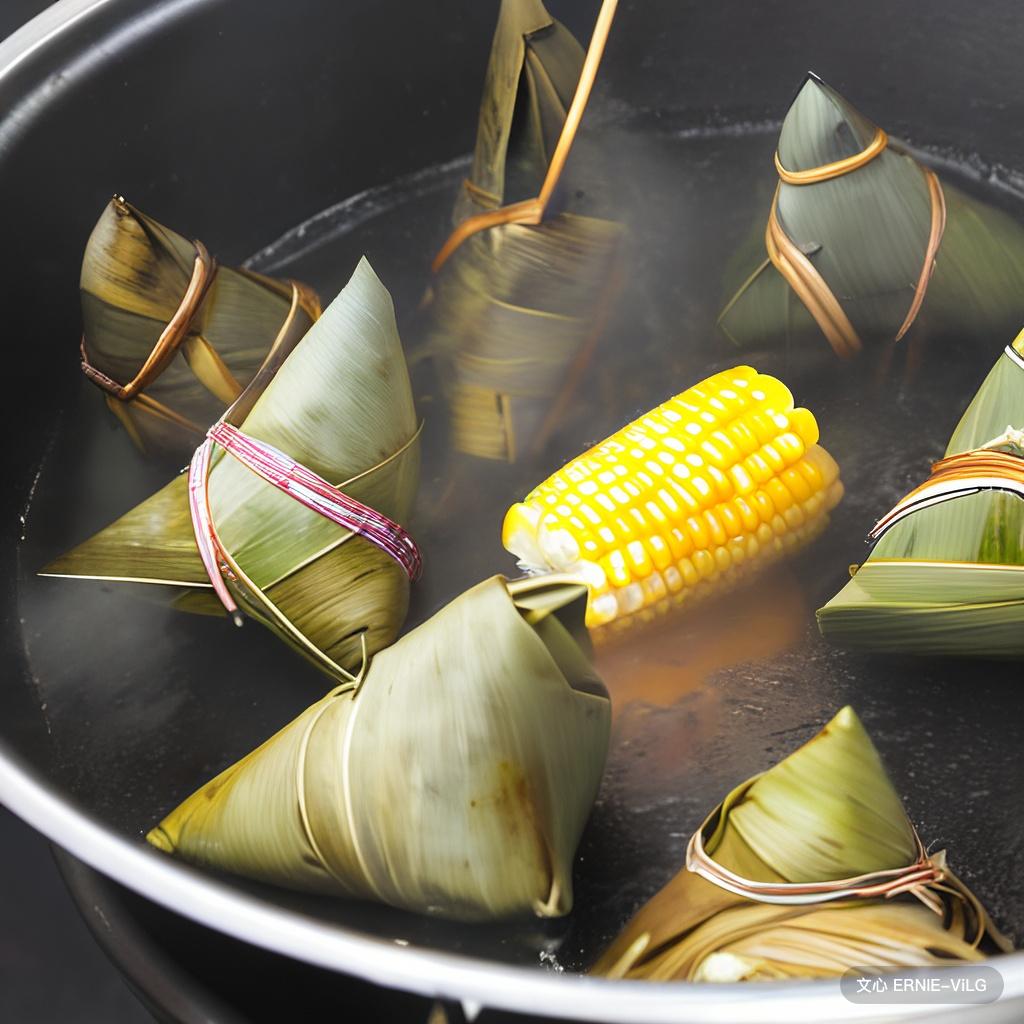} &
        \includegraphics[width=0.2\linewidth]{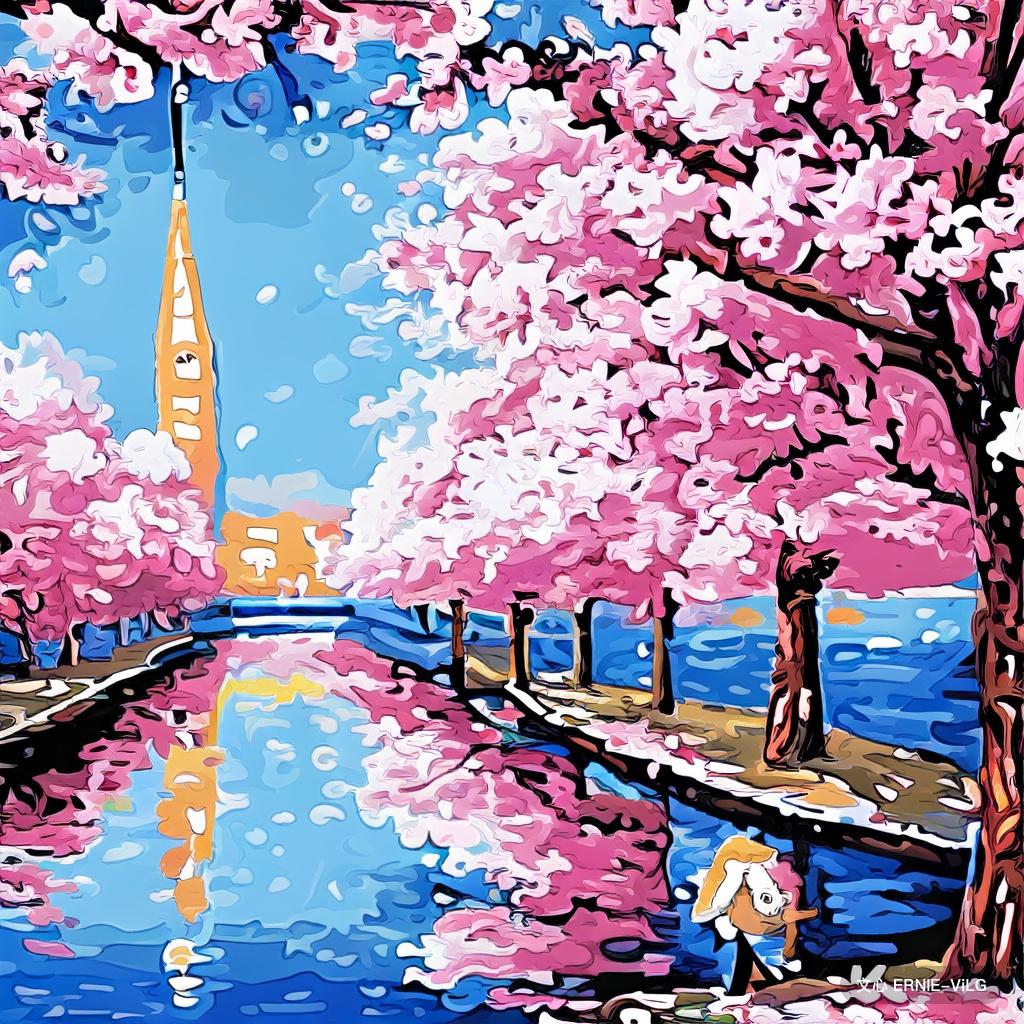}
        \\
        \rotatebox{90}{\scriptsize\phantom{A.} DALL-E~2} &
        \includegraphics[width=0.2\linewidth]{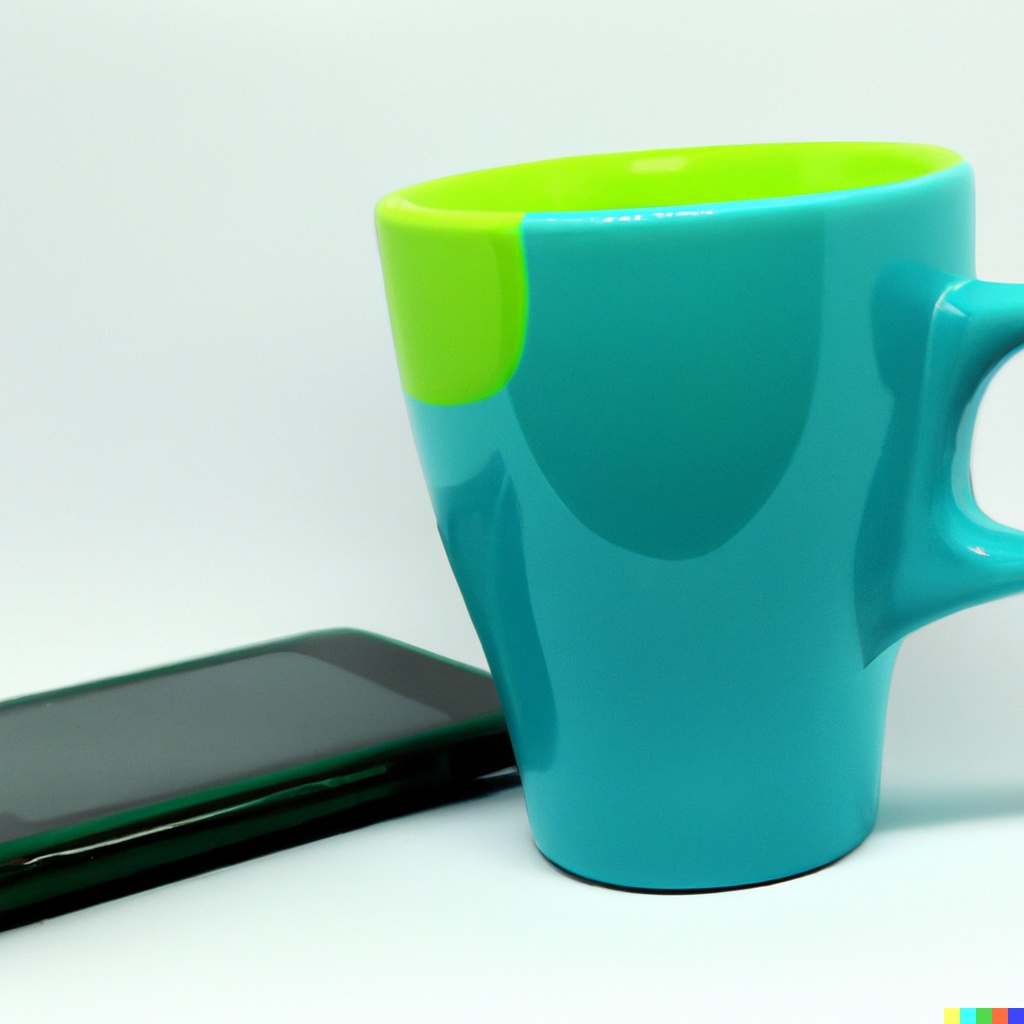} &
        \includegraphics[width=0.2\linewidth]{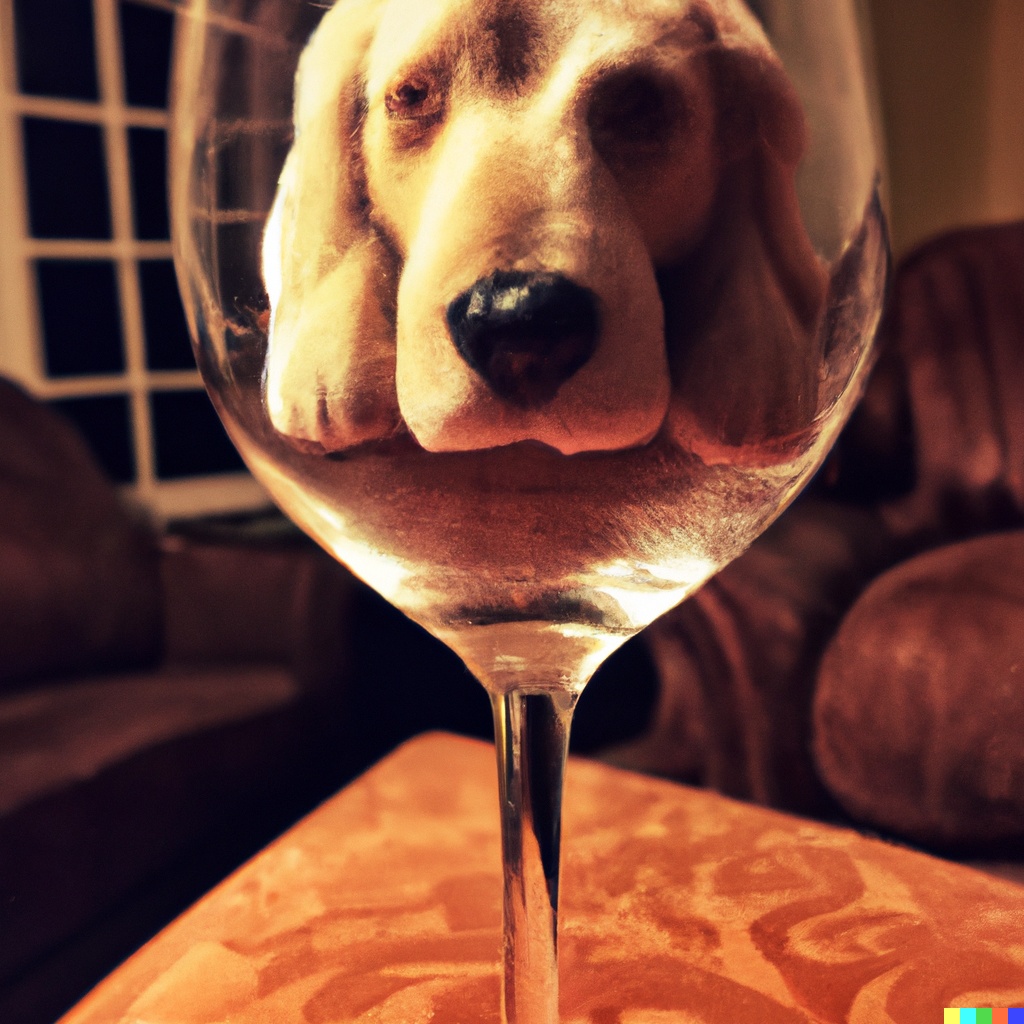} &
        \includegraphics[width=0.2\linewidth]{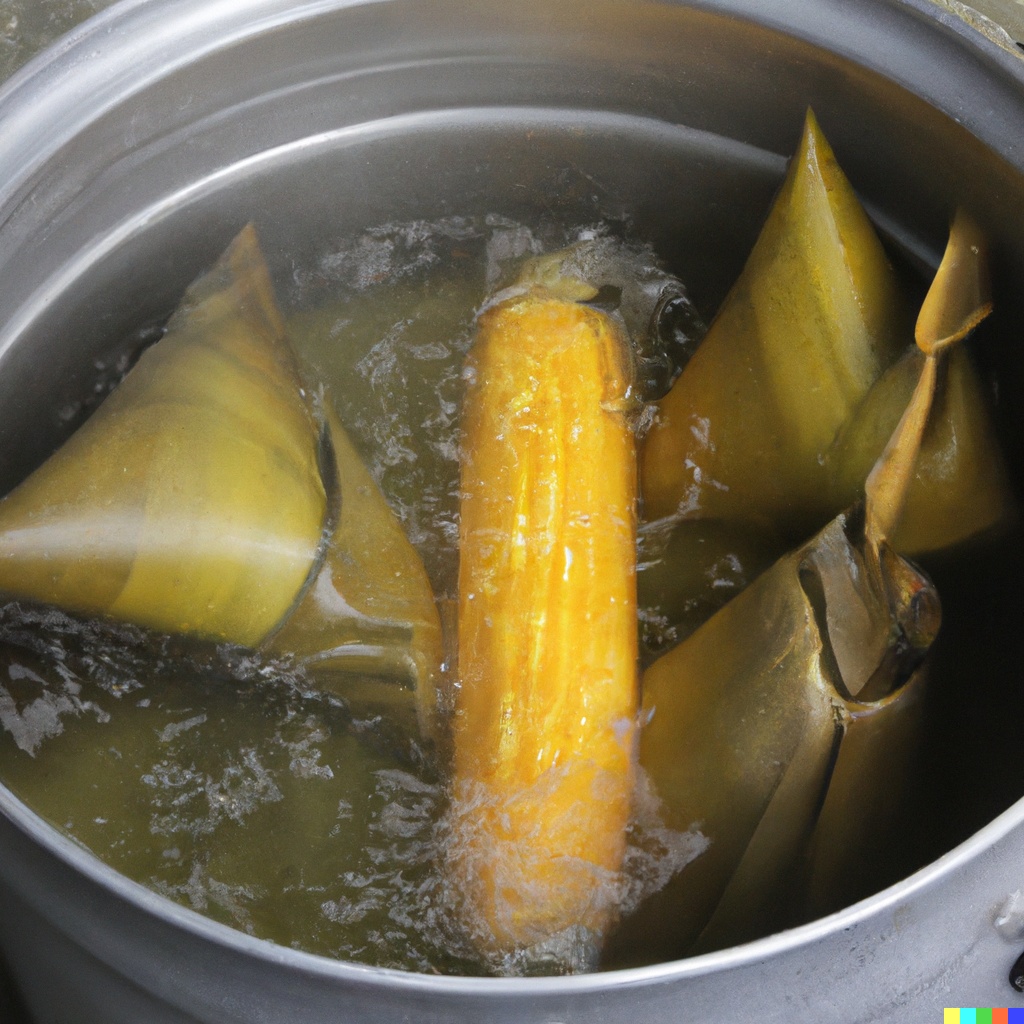} &
        \includegraphics[width=0.2\linewidth]{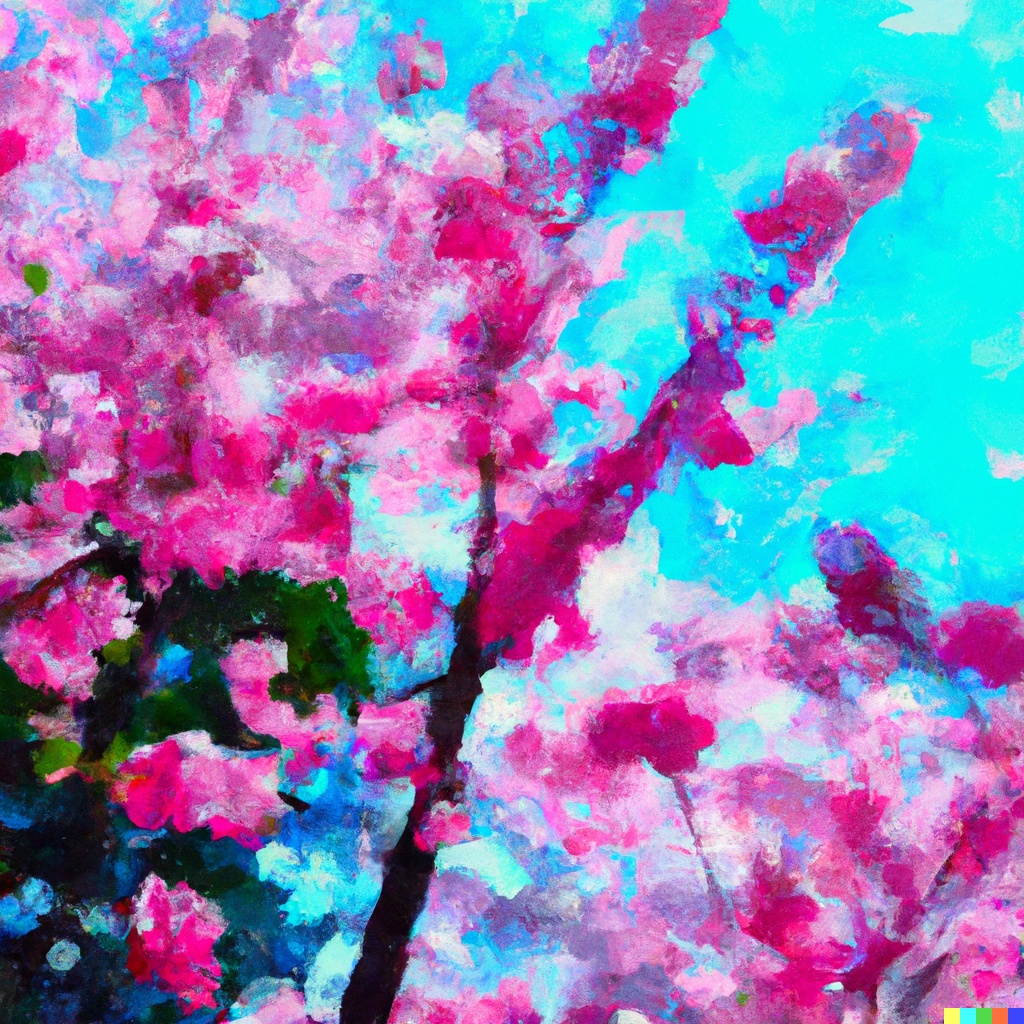}
        \\
        \rotatebox{90}{\scriptsize\phantom{AA} Sta. Diff.} &
        \includegraphics[width=0.2\linewidth]{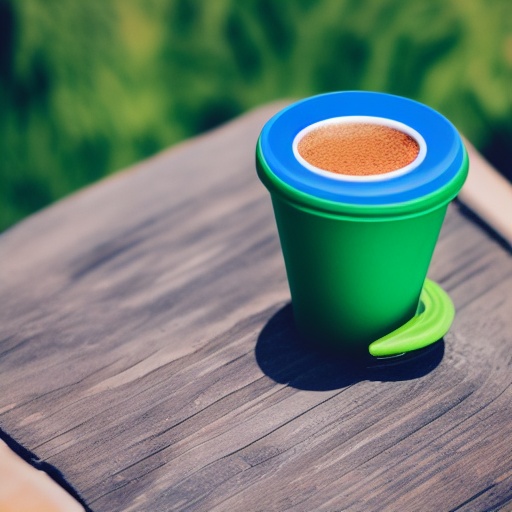} &
        \includegraphics[width=0.2\linewidth]{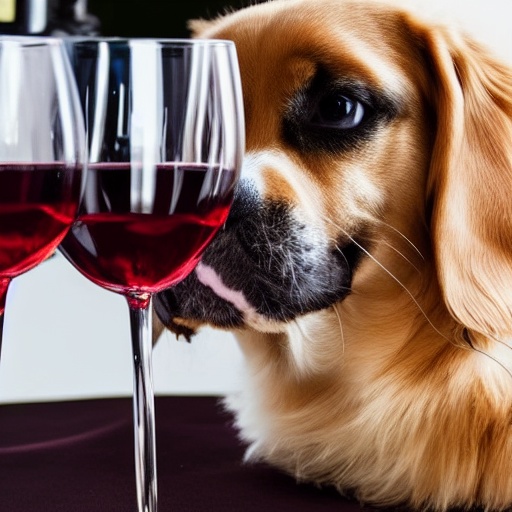} &
        \includegraphics[width=0.2\linewidth]{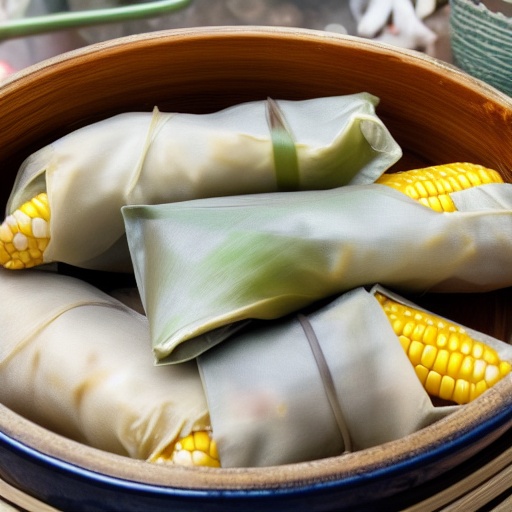} &
        \includegraphics[width=0.2\linewidth]{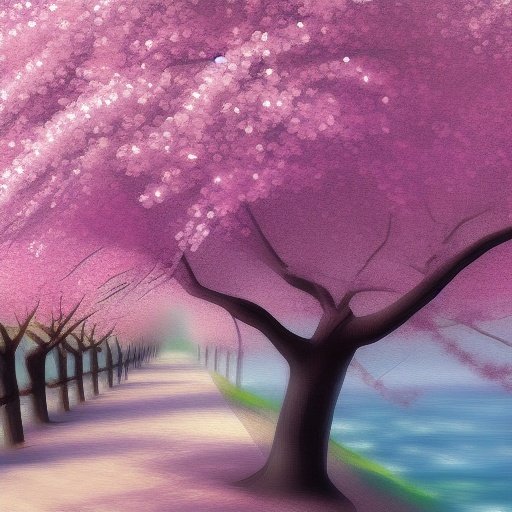}
        \\
        & \multicolumn{2}{l}{\scriptsize \makecell[l]{(1) A green cup and a blue cell phone \\ (3) Zongzi and corn boiled in the pot}}
        &
        \multicolumn{2}{l}{\scriptsize \makecell[l]{(2) A wine glass on top of a dog \\ (4) Cherry blossom, digital oil painting}}
    \end{tabular}
    \caption{Qualitative Comparison of ERNIE-ViLG~2.0 and DALL-E~2/Stable Diffusion on ViLG-300. For $4$ v.s. $4$ comparison, refer to the Figure~\ref{fig:case_drawbench} and \ref{fig:case_ernie_vilg} in appendix.}
    \label{fig:main_case}
\end{figure}
\end{CJK*}

\noindent\textbf{Human Evaluation on ViLG-300.}
ERNIE-ViLG~2.0 takes Chinese prompts as input and generates high-resolution images, unlike recent English-oriented text-to-image models.
Herein, we introduce ViLG-300\footnote{\url{https://github.com/PaddlePaddle/ERNIE/tree/ernie-kit-open-v1.0/Research/ERNIE-ViLG2/data/ViLG-300.csv}}, a bilingual prompt set that supports the systematic evaluation and comparison of Chinese and English text-to-image models.
ViLG-300 contains 300 prompts from 16 categories, composed of DrawBench~\cite{DBLP:journals/corr/abs-2205-11487} (in English) and the prompt set used in ERNIE-ViLG~\cite{DBLP:journals/corr/abs-2112-15283} (in Chinese). 
See Appendix~\ref{appx:vilg300} for more details about the construction process. 

With ViLG-300, we can make convincing comparisons between ERNIE-ViLG~2.0 and DALL-E~2\footnote{\url{https://openai.com/dall-e-2/}}, Stable Diffusion\footnote{\url{https://beta.dreamstudio.ai/dream}}\footnote{We use DALL-E~2 and Stable Diffusion interfaces to generate images on October 25, 2022, before the CVPR 2023 submission deadline.}.
For evaluation, five raters are presented with two sets of images generated by ERNIE-ViLG~2.0 and the compared model.
Next, they are asked to compare these images from two dimensions of image-text alignment and image fidelity, and then select the model they prefer, or respond that there is no measurable difference between two models. 
Throughout the process, raters are unaware of which model the image is generated from, and we do not apply any filtering strategy to the rating results.
Figure~\ref{fig:human} shows that human raters prefer ERNIE-ViLG~2.0 over all other models in both image-text alignment (56.5\%$\pm$3.8\% and 68.2\%$\pm$3.8\% when compared to DALL-E~2 and Stable Diffusion, respectively) and image fidelity (58.8\%$\pm$3.6\% to DALL-E~2,  66.5\%$\pm$3.5\% to Stable Diffusion, respectively), which again proves that ERNIE-ViLG~2.0 can generate high-quality images that conform to the text, with the help of knowledge enhancement and mixture-of-denoising-experts strategies.
We provide comparisons of separate categories in Appendix~\ref{appx:result}.
Beyond text relevancy and image fidelity, we also observe that ERNIE-ViLG~2.0 can generate images with better sharpness and textures than baseline models. 
See also Appendix~\ref{appx:definition} for detailed discussions.

\begin{figure}
  \centering
  \begin{subfigure}{0.48\linewidth}
    \includegraphics[width=\linewidth]{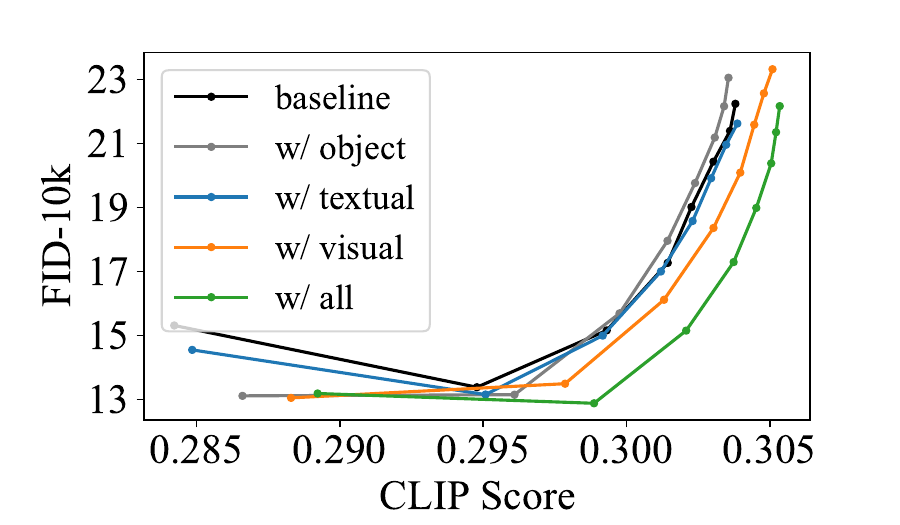}
    \caption{Knowledge enhancement.}
    \label{fig:ke_eval}
  \end{subfigure}
  \begin{subfigure}{0.48\linewidth}
    \includegraphics[width=\linewidth]{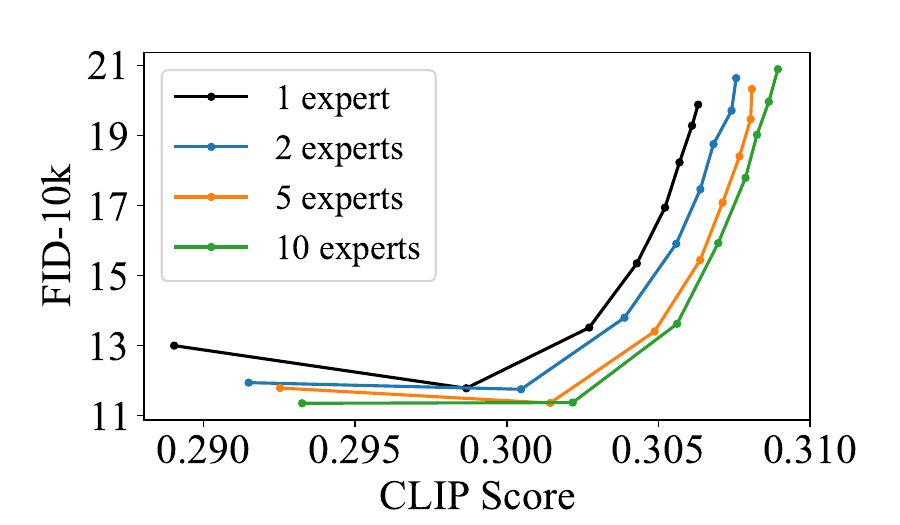}
    \caption{Mixture-of-denoising-experts.}
    \label{fig:moe_eval}
  \end{subfigure}
  \caption{Performance with various strategies in ERNIE-ViLG~2.0. Here we draw pareto curves with guidance scale [2,3,4,5,6,7,8,9].}
  \label{fig:ablation}
\end{figure}

\begin{CJK*}{UTF8}{gbsn}
\begin{figure}[t]
    \centering
    \setlength{\tabcolsep}{1.25pt}
    \begin{tabular}{cccccccccc}
        \rotatebox{90}{\scriptsize baseline} &
        \includegraphics[width=0.105\linewidth]{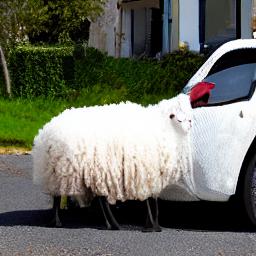} &
        \includegraphics[width=0.105\linewidth]{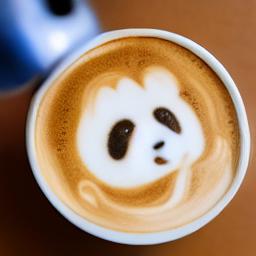} &
        \includegraphics[width=0.105\linewidth]{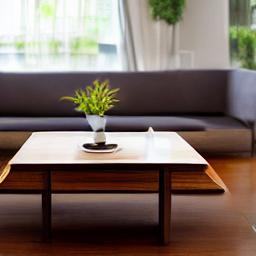} &
        \includegraphics[width=0.105\linewidth]{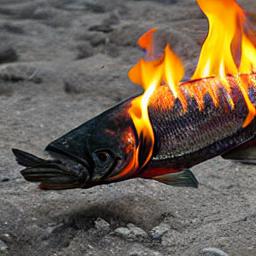} &
        \rotatebox{90}{\scriptsize\phantom{.} 1 exp.} &
        \includegraphics[width=0.105\linewidth]{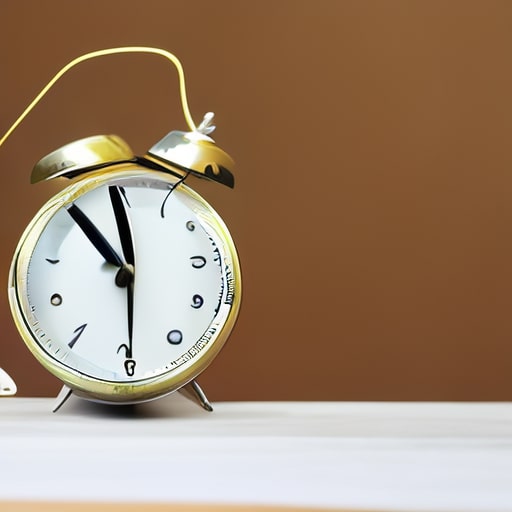} &
        \includegraphics[width=0.105\linewidth]{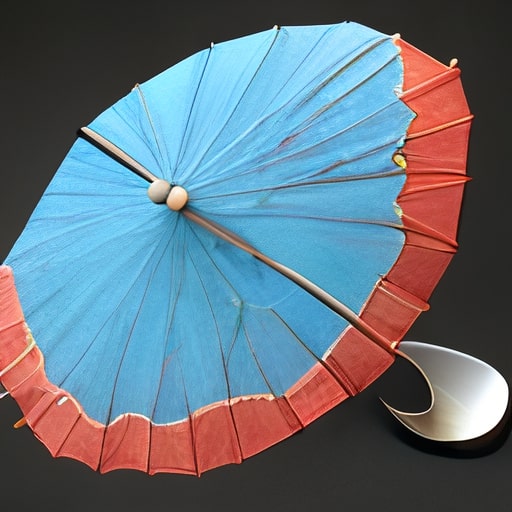} &
        \includegraphics[width=0.105\linewidth]{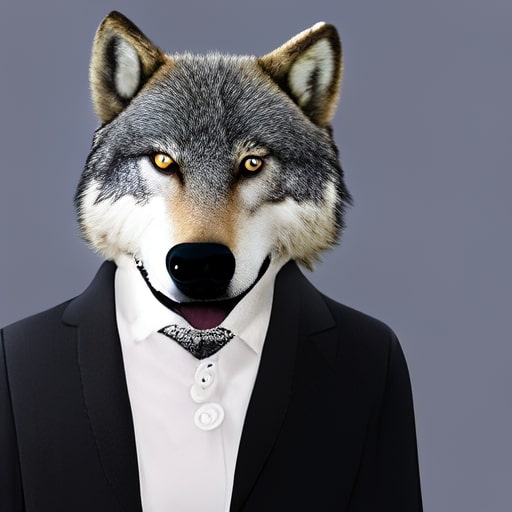} &
        \includegraphics[width=0.105\linewidth]{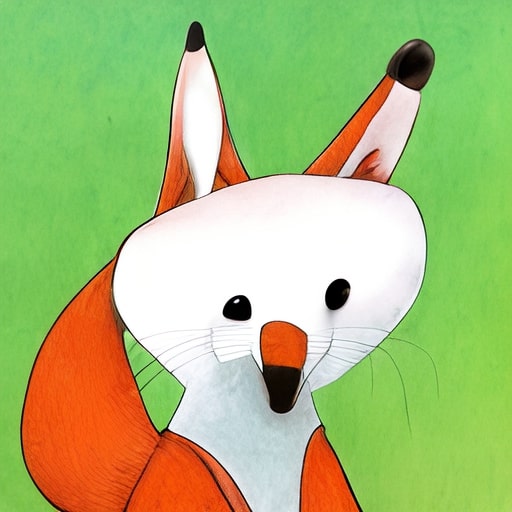}
        \\
        \rotatebox{90}{\scriptsize\phantom{.} w/ txt.} &
        \includegraphics[width=0.105\linewidth]{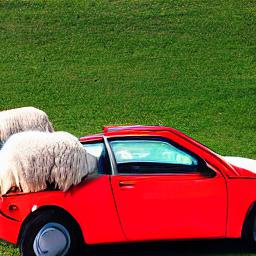} &
        \includegraphics[width=0.105\linewidth]{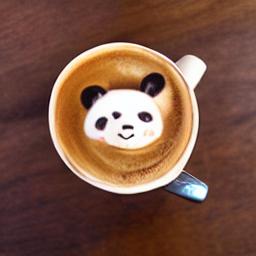} &
        \includegraphics[width=0.105\linewidth]{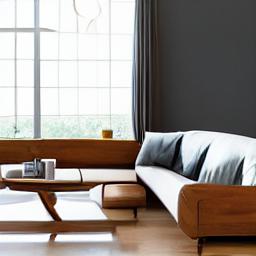} &
        \includegraphics[width=0.105\linewidth]{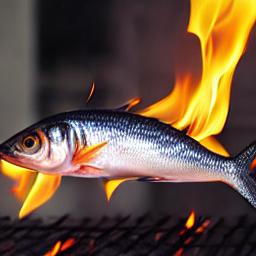} &
        \rotatebox{90}{\scriptsize\phantom{.} 2 exp.} &
        \includegraphics[width=0.105\linewidth]{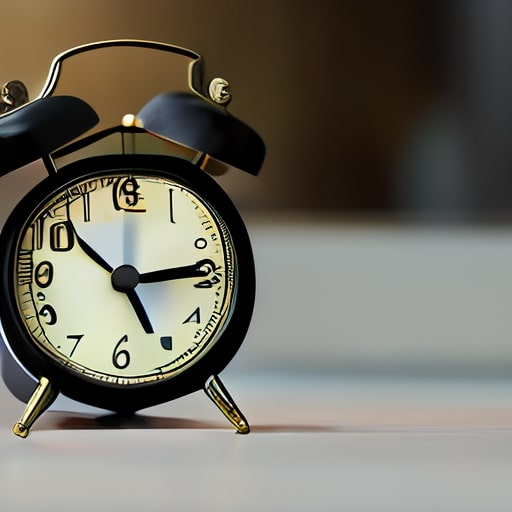} &
        \includegraphics[width=0.105\linewidth]{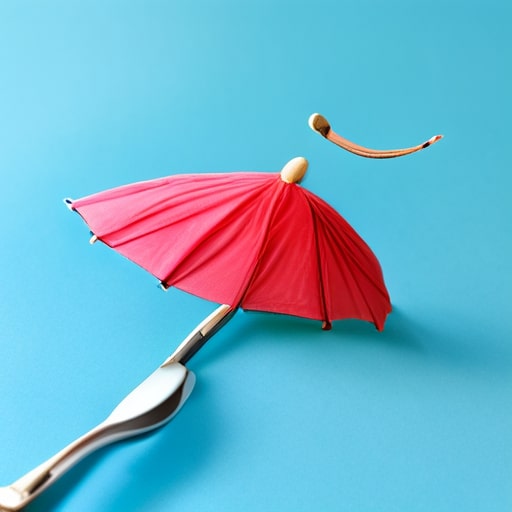} &
        \includegraphics[width=0.105\linewidth]{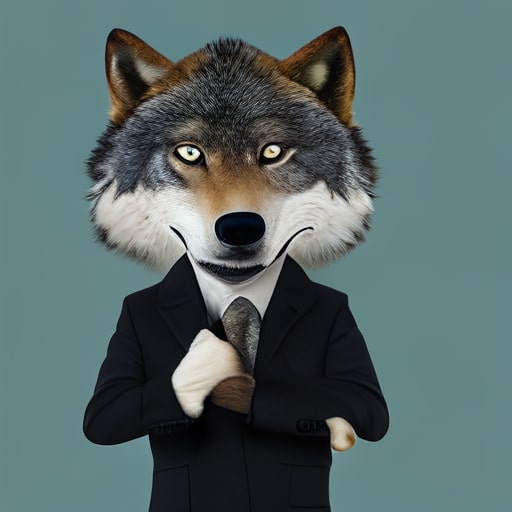} &
        \includegraphics[width=0.105\linewidth]{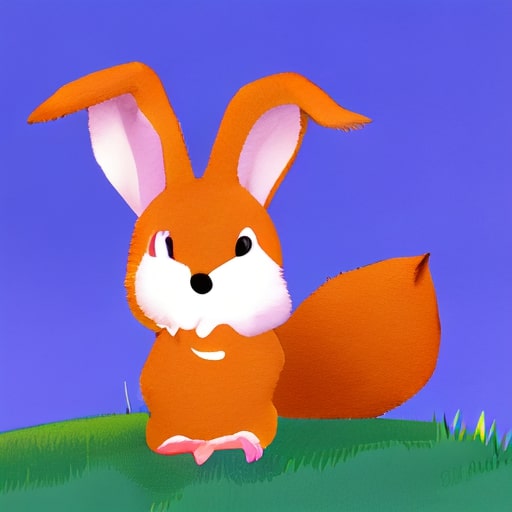}
        \\
        \rotatebox{90}{\scriptsize\phantom{.} w/ vis.} &
        \includegraphics[width=0.105\linewidth]{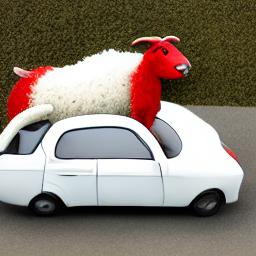} &
        \includegraphics[width=0.105\linewidth]{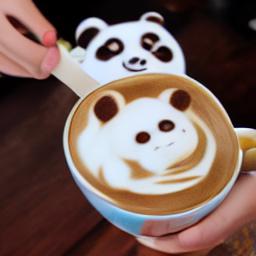} &
        \includegraphics[width=0.105\linewidth]{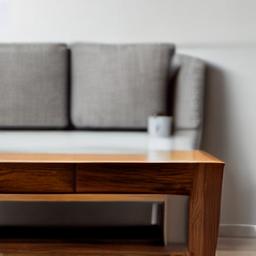} &
        \includegraphics[width=0.105\linewidth]{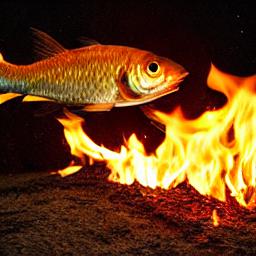} &
        \rotatebox{90}{\scriptsize\phantom{.} 5 exp.} &
        \includegraphics[width=0.105\linewidth]{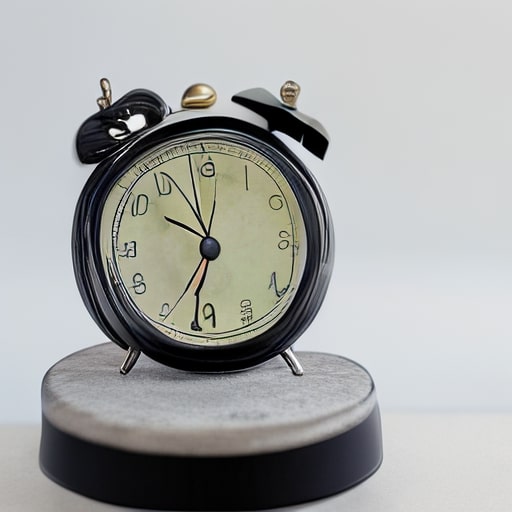} &
        \includegraphics[width=0.105\linewidth]{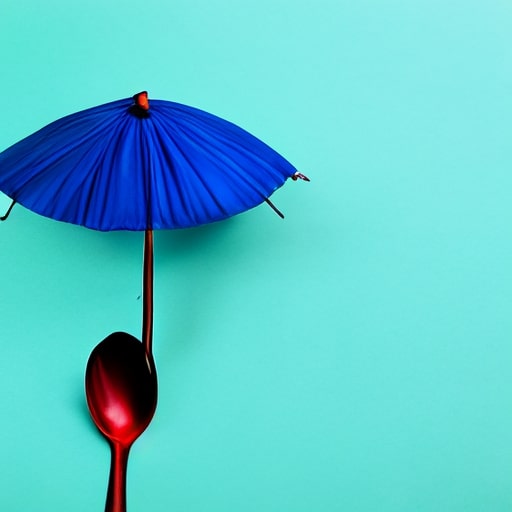} &
        \includegraphics[width=0.105\linewidth]{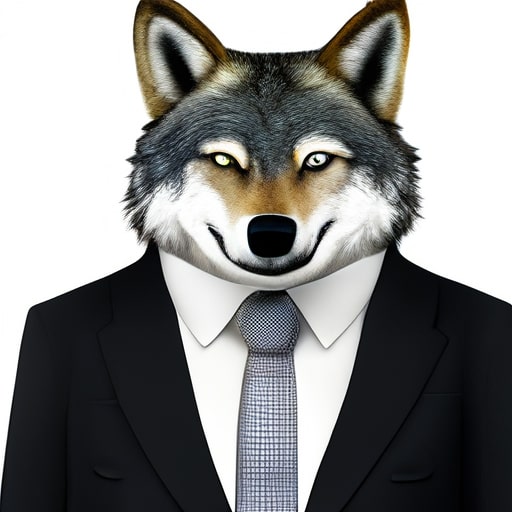} &
        \includegraphics[width=0.105\linewidth]{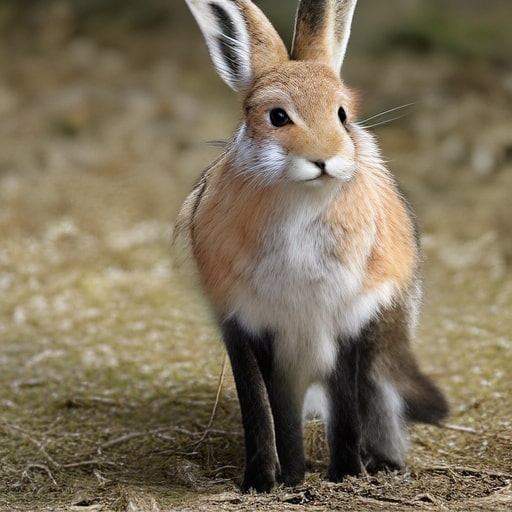}
        \\
        \rotatebox{90}{\scriptsize\phantom{.} w/ all} &
        \includegraphics[width=0.105\linewidth]{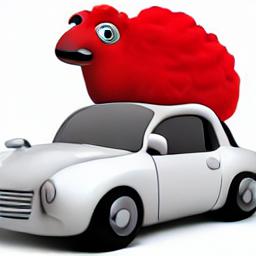} &
        \includegraphics[width=0.105\linewidth]{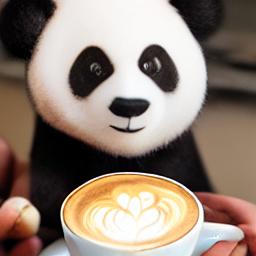} &
        \includegraphics[width=0.105\linewidth]{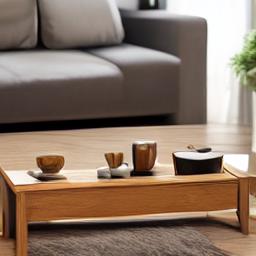} &
        \includegraphics[width=0.105\linewidth]{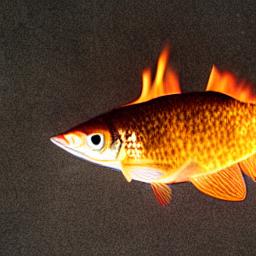} &
        \rotatebox{90}{\scriptsize\phantom{} 10 exp.} &
        \includegraphics[width=0.105\linewidth]{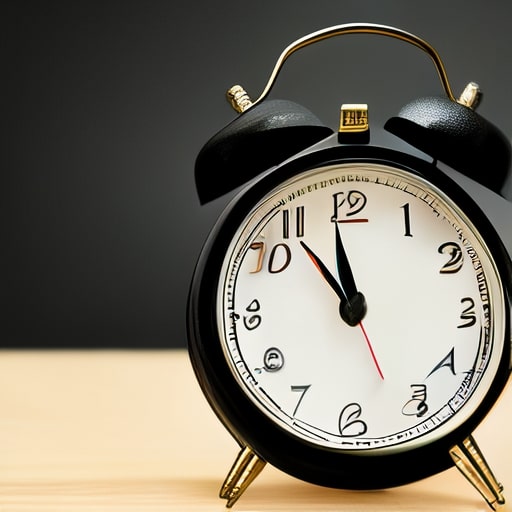} &
        \includegraphics[width=0.105\linewidth]{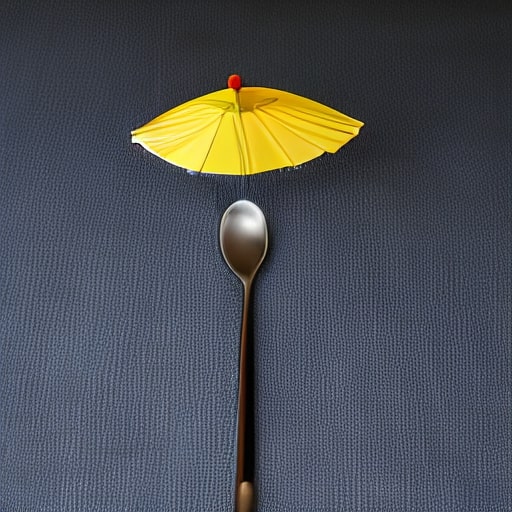} &
        \includegraphics[width=0.105\linewidth]{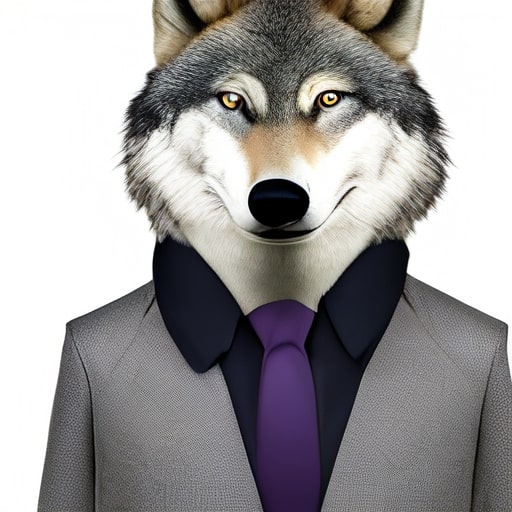} &
        \includegraphics[width=0.105\linewidth]{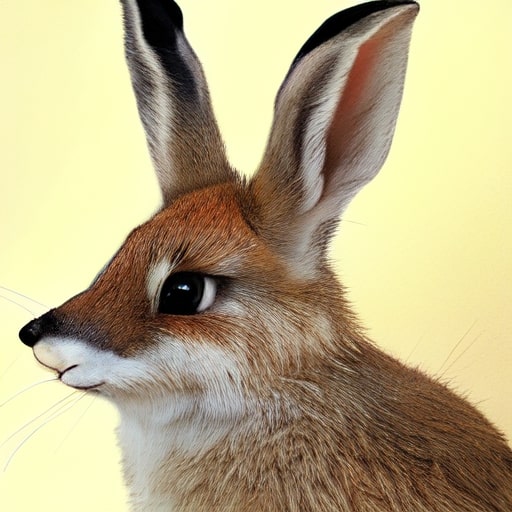}
        \\
        \multicolumn{5}{l}{\scriptsize \makecell[l]{(1) A white car and a red sheep \\ (2) A panda making latte art \\ (3) A small red ball in a big green block \\ (4) A burning fish}} &
        \multicolumn{5}{l}{\scriptsize \makecell[l]{(1) A single clock is sitting on a table \\ (2) An umbrella on top of a spoon \\ (3) Wolf in a suit \\ (4) A rabbit with a fox's head}}
    \end{tabular}
    \caption{Samples from ViLG-300 with different knowledge enhancement strategies (left) and different number of experts (right). For more samples, refer to the Figure~\ref{fig:ke_case} and \ref{fig:mode_case} in supplement.}
    \label{fig:ablation_case}
\end{figure}
\end{CJK*}

\subsection{Analysis}
\label{sec:MoDE-ablation}

To examine the effectiveness of our design philosophy, we conduct two groups of ablation studies.
Similar to the main experiment, we also provide both automatic metrics and intuitive showcases to demonstrate the advantages of each strategy in ERNIE-ViLG~2.0 here.

\noindent\textbf{Knowledge Enhancement Strategies.}
In this part, we focus on the impact of various knowledge enhancement strategies by training a series of lightweight models, with 500M text encoders, 870M U-Nets, and 500M training samples.
The pareto curves in Figure~\ref{fig:ke_eval} and convergence curves in Figure~\ref{fig:ke_conv} (Appendix~\ref{appx:ke_ablation}) demonstrate that incorporating knowledge in the learning process brings significant performance gains in image fidelity, image-text alignment, and convergence speed.
Specifically, (1) the benefits of textual knowledge are mainly reflected in precise fine-grained semantic control (\verb|w/ textual|), (2) only utilizing object knowledge may not be able to steadily promote the performance (\verb|w/ object|), while taking synthetic descriptions into consideration is an effective solution to make full use of visual knowledge (\verb|w/ visual|).
Figure~\ref{fig:ablation_case} provides more visual comparisons to intuitively demonstrate the changes brought by each strategy.
When handling complex prompts, \verb|baseline| model faces problems such as the absence of key objects or incorrect assignment of attributes.
At this point, textual knowledge helps the model accurately understand the attributes of each object, but the generated images sometimes fall into a new problem of distortion.
Complementarily, visual knowledge promotes the generation of high-fidelity images, but it cannot well understand specific entities in text. Eventually, the combination of two kinds of knowledge 
harmoniously promotes the model from single- and multi-modal views,
which ensures high fidelity and boost the image-text alignment in fine-grained visual scene.
In Appendix~\ref{appx:ke_ablation}, we also quantitatively measure the CLIP score improvement brought by different knowledge sources, which corroborates the qualitative observation.

\begin{CJK*}{UTF8}{gbsn}
\begin{figure}[t]
    \centering
        \includegraphics[width=\linewidth]{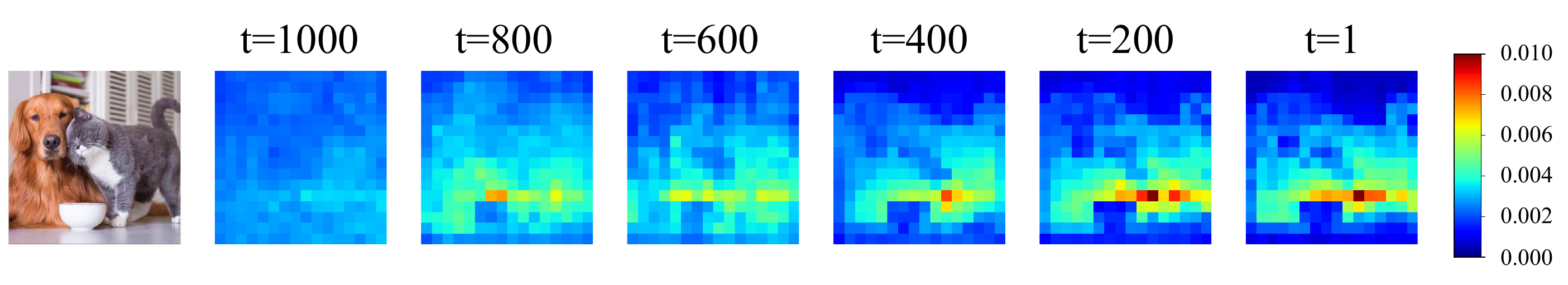}
    \caption{The visualization of cross-attention maps in different denoising timesteps, where each value in the image space is the average of attention from this image token to all text tokens.}
    \label{fig:attn_vis}
\end{figure}
\end{CJK*}

\noindent\textbf{Mixture-of-Denoising-Experts Strategies.}
Based on the above lightweight settings, we further train the \verb|baseline| model with 500M samples, and then train 200M samples for each denoising expert.
Figure~\ref{fig:moe_eval} shows that with the increasing number of denoising experts, the overall performance is gradually improved, proving that scaling the size of U-Net is also an effective solution to achieve better image quality.
More showcases are provided in Figure~\ref{fig:ablation_case}. 
When the number of experts increases from 1 to 10, the model can not only better handle the coupling between different elements but also generate images with more natural textures. For instance, the numbers on clocks become clearer, the proportion of wolf and suit becomes more harmonious, and the model can generate more photorealistic pictures instead of cartoon drawings.
We also tried to analyze the impact of the amount of denoising experts and training samples, and found that using more expert networks has better performance than using a network to train more samples (see also Appendix~\ref{appx:moe_ablation}).

Figure~\ref{fig:attn_vis} further visualizes the cross-attention maps from image features to text representations in denoising experts during 1,000-step denoising process, where these steps shown are denoised by different experts.
As shown in the illustration, attentions of different denoising timesteps vary. 
Specifically, the attention maps of timesteps $t$ near 1,000 are almost evenly distributed over the whole image, which is because the input of these steps is close to Gaussian noise and the image layout is unclear, so all the image tokens have to attend to the text prompt to generate image skeleton.
When the timesteps are close to 1, attention maps concentrate more on foreground objects. For these timesteps, the input to denoising network is close to the final image and the layout is clear, and only a few parts of the image need to focus on the text to fill in the details of object.
These observations again illustrate the difference among denoising timesteps and demonstrate the need to disentangle different timesteps with multiple experts.

\section{Related Work}

\noindent\textbf{Text-to-Image Generation.} 
Text-to-image generation is the task of synthesizing images according to natural language descriptions. 
Early works adopted generative adversarial networks~\cite{DBLP:conf/nips/GoodfellowPMXWOCB14} to produce images based on text~\cite{DBLP:conf/cvpr/XuZHZGH018,DBLP:conf/cvpr/ZhuP0019,DBLP:conf/cvpr/Tao00JBX22,DBLP:conf/cvpr/0010KBLY21}.
Inspired by the success of transformers in generation tasks~\cite{DBLP:conf/nips/VaswaniSPUJGKP17}, models such as ERNIE-ViLG~\cite{DBLP:journals/corr/abs-2112-15283}, DALL-E~\cite{DBLP:conf/icml/RameshPGGVRCS21}, Cogview~\cite{DBLP:conf/nips/DingYHZZYLZSYT21}, Make-A-Scene~\cite{DBLP:journals/corr/abs-2203-13131}, and Parti~\cite{DBLP:journals/corr/abs-2206-10789} have also explored text-to-image generation as a sequence-to-sequence problem, with auto-regressive transformers as generators and text/image tokens as input/output sequences.
Recently, another line of works have applied diffusion models~\cite{DBLP:conf/icml/Sohl-DicksteinW15}, shaping it as an iterative denoising task~\cite{DBLP:conf/nips/HoJA20,DBLP:conf/iclr/SongME21,DBLP:conf/iclr/SalimansH22}. 
By adding text condition in the denoising steps, practices such as LDM~\cite{DBLP:journals/corr/abs-2112-10752}, DALL-E~2~\cite{DBLP:journals/corr/abs-2204-06125}, and Imagen~\cite{DBLP:journals/corr/abs-2205-11487} constantly set new records in text-to-image generation.
Based on diffusion models as the backbone, ERNIE-ViLG~2.0 proposes incorporating knowledge of scene and mixture-of-denoising-experts mechanism into the denoising process.

\noindent\textbf{Knowledge-Enhanced Pre-trained Models.}
While transformers benefit from pre-training on large-scale data, many attempts have been adding knowledge to guide them to focus on key elements during learning.
For language-based tasks, knowledge-enhanced models used knowledge masking strategy~\cite{DBLP:journals/corr/abs-1904-09223,DBLP:journals/tacl/JoshiCLWZL20} or knowledge-aware pre-training tasks~\cite{DBLP:conf/aaai/SunWLFTWW20,DBLP:journals/corr/abs-2107-02137} to understand the language data distribution.
As for vision-language multi-modal discrimination models,
OSCAR~\cite{DBLP:conf/eccv/Li0LZHZWH0WCG20}, ERNIE-ViL~\cite{DBLP:conf/aaai/0010TYSTW021} and ERNIE-Layout~\cite{DBLP:journals/corr/abs-2210-06155} leveraged object tags, scene graphs, and document layouts as extra knowledge to help the models better align language and vision modalities.
Among multi-modal generation models, Make-A-Scene~\cite{DBLP:journals/corr/abs-2203-13131} emphasized the importance of object and face regions by integrating domain-specific perceptual knowledge.
While current text-to-image diffusion models suffer from attribute misalignment problems~\cite{DBLP:journals/corr/abs-2204-06125}, they have not employed any specific knowledge of objects.
Herein, ERNIE-ViLG~2.0 utilizes the knowledge of key elements in images and text to enhance diffusion models, leading to better fine-grained image-text alignment in generated pictures.

\noindent\textbf{Mixture-of-Expert.}
Mixture-of-Experts (MoE) in neural networks means dividing specific parts of the parameters into subsets, each of which is called an expert~\cite{DBLP:conf/iclr/ShazeerMMDLHD17,DBLP:journals/corr/abs-2209-01667}. During the forward pass, a router assigns experts to different input, and each input only interacts with the experts assigned to.
The router is a critical part of MoE.
In language tasks, the most common strategy is a matching algorithm that assigns each text token to several experts in the linear feed-forward layer~\cite{DBLP:conf/iclr/LepikhinLXCFHKS21,DBLP:journals/corr/abs-2101-03961,DBLP:conf/nips/RollerSSW21,DBLP:conf/naacl/GururanganLHSZ22}. 
While most practices formulate multiple experts in only the linear layers, some works also use an entire language model as an expert~\cite{DBLP:journals/corr/abs-2208-03306}.
Beyond the natural language processing tasks, the idea of MoE have also been applied to vision models~\cite{DBLP:conf/iclr/PuigcerverRMRPG21} and Mixture-of-Modality-Expert in multi-modal transformers~\cite{DBLP:journals/corr/abs-2111-02358,DBLP:journals/corr/abs-2208-10442,DBLP:journals/corr/abs-2206-02770}.
In ERNIE-ViLG~2.0, the MoDE mechanism takes multiple denoising U-Nets as experts. It uses the denoising step index as the fixed router to determine which expert to use.

\section{Risks, Limitations, and Future Work}\label{sec:risk}

\noindent\textbf{Model Usage and Data Bias.}
Text-to-image generation models trained by large-scale image-text data have all faced similar risks regarding to inappropriate usage of generation and data bias~\cite{DBLP:journals/corr/abs-2112-10752,DBLP:journals/corr/abs-2204-06125,DBLP:journals/corr/abs-2205-11487}.
Considering that text-to-image models help people realize their imagination with less effort, the malicious use of models may result in unexpected deceptive or harmful outcomes. 
Moreover, since the models are trained on datasets consisting of images and their alt-text crawled from websites, the generated images may exhibit social and cultural bias in the datasets and websites.

\begin{figure}[t]
    \centering
    \setlength{\tabcolsep}{1pt}
    \begin{tabular}{cccc}
    \includegraphics[width=0.30\linewidth]{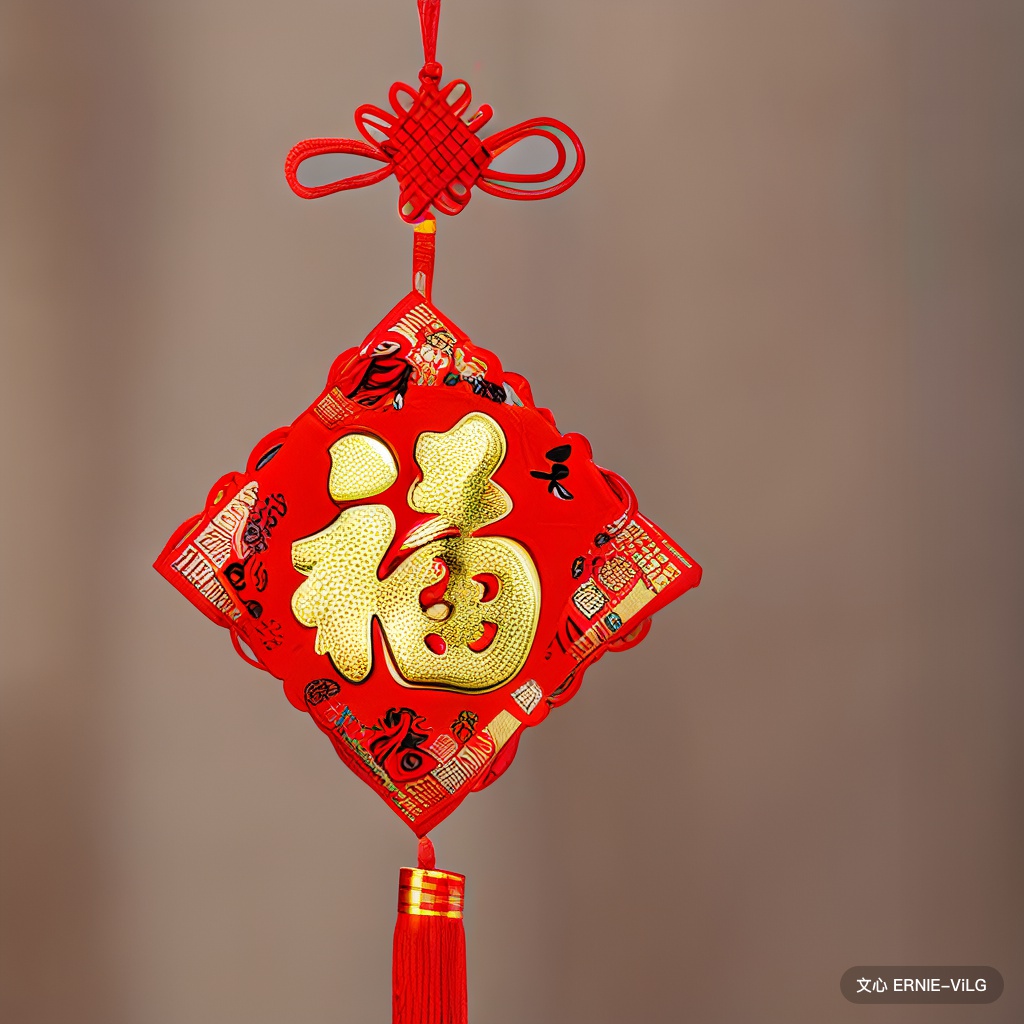} &
    \includegraphics[width=0.30\linewidth]{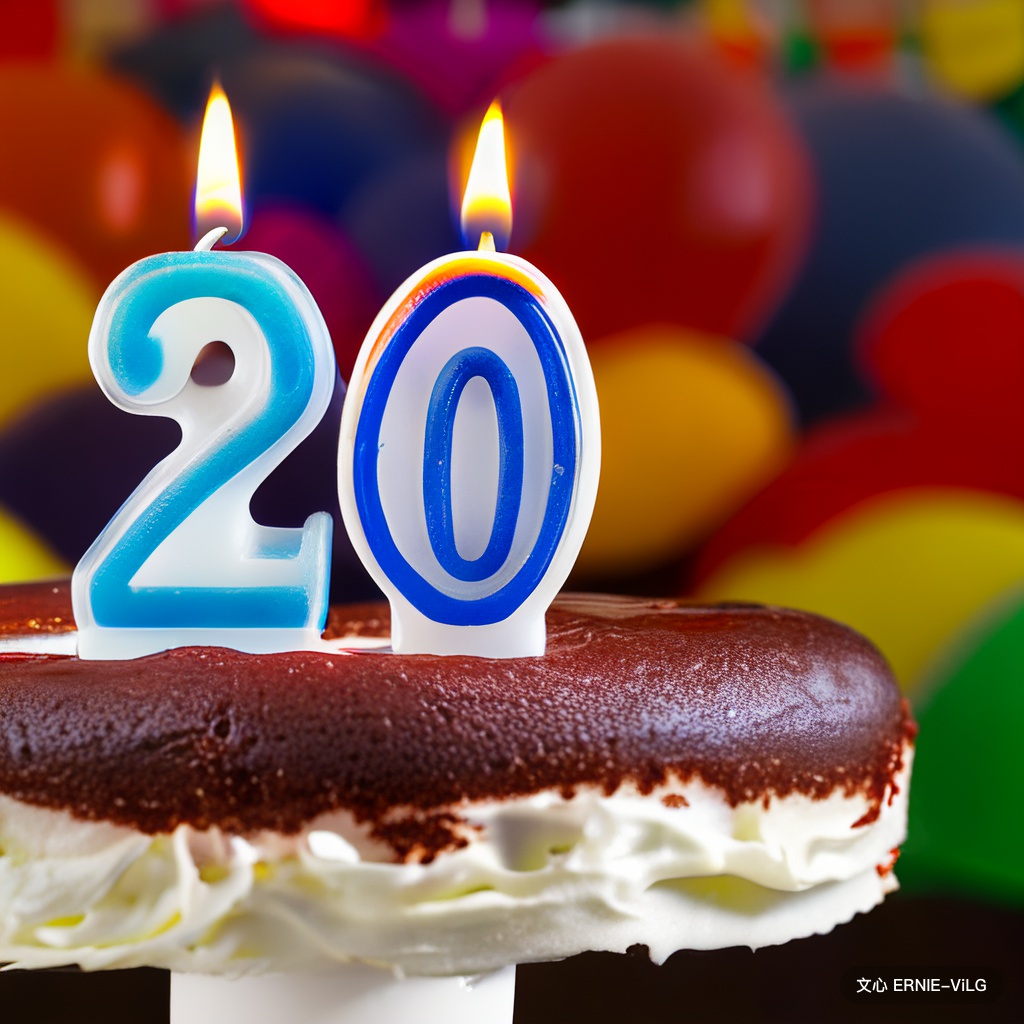} &
    \includegraphics[width=0.30\linewidth]{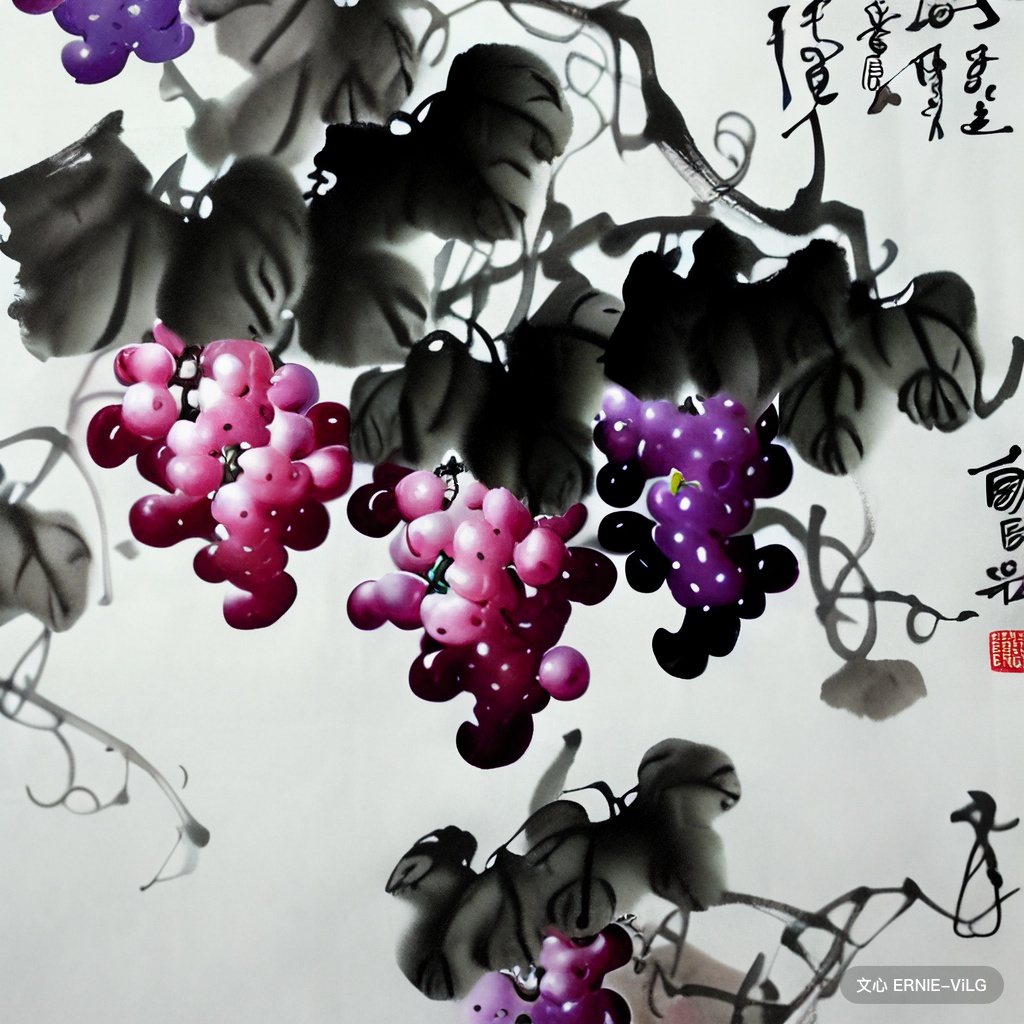} \\
    \scriptsize \makecell{(a) 国风福字挂饰} &
    \scriptsize \makecell{(b) 生日蛋糕上有蜡烛“20”} &
    \scriptsize \makecell{(c) 国画葡萄} \\
    \scriptsize \makecell{A hanging ornament with \\ ``福'' in Chinese fashion} &
    \scriptsize \makecell{A birthday cake with \\ candles of ``20'' on it} &
    \scriptsize \makecell{Chinese painting of grapes} \\
    \end{tabular}
    \caption{Examples of character rendering. 
    The model successfully renders simple characters specified in the prompt, while for more difficult cases, the model only learns the position for now.
    }
    \label{fig:chinese_character} 
\end{figure}

\begin{CJK*}{UTF8}{gbsn}
\noindent\textbf{Character Rendering.}
Figure~\ref{fig:chinese_character} shows two successful character rendering cases (a, b) and one failure case (c).
Character rendering is a challenging task for ERNIE-ViLG~2.0 for two reasons.
First, the training data contains both Chinese text-image pairs and English text-image pairs translated into Chinese. When a text prompt mentions characters, the characters in the image could be in Chinese or English, and it is hard for the model to learn corresponding characters in both languages simultaneously.
In the cases of successful character rendering that we observed, the characters could be words that are common in Chinese and do not have an exact match in English, such as ``福'' (``blessing, happiness, good luck'' in English) in Figure~\ref{fig:chinese_character}a, or numbers which are the same in English and Chinese images, such as ``20'' in Figure~\ref{fig:chinese_character}b.
The second reason that makes character rendering difficult is probably that Chinese characters are complex combinations of strokes without basic components like English letters.
In Figure~\ref{fig:chinese_character}c, the model does learn that it should write some Chinese characters in the top right corner, but it only paints meaningless strokes there.
\end{CJK*}

\begin{CJK*}{UTF8}{gbsn}
\end{CJK*}

\noindent\textbf{Variation of Mixture-of-Denoising-Experts.}
Section~\ref{sec:MoDE-ablation} shows that using more denoising experts leads to better model performance. It indicates that using parallel U-Net experts is an effective way to augment the denoising network. Due to the computation limitation, we only try using up to 10 experts in this work, while we believe that exploring more denoising experts and multiple text encoders as experts is a meaningful future direction. 
Herein, we can further scale up the model and allow it to learn data distribution better with similar inference time.

\section{Conclusions}
We present ERNIE-ViLG~2.0, the first Chinese large-scale text-to-image generation model based on diffusion models.
To improve the fine-grained control of scene semantics, we incorporate visual and textual knowledge of the scene into diffusion models. To disentangle the model parameters for different denoising timesteps, we introduce MoDE and scale up the model parameters to 24B with a relatively short inference time.
Experiments show that ERNIE-ViLG~2.0 achieves state-of-the-art on MS-COCO and each proposed mechanism contributes to the final results.
To allow fair comparisons between Chinese and English text-to-image models, we collect a bilingual prompt set ViLG-300, and human evaluation indicates that ERNIE-ViLG~2.0 is preferred over strong baselines in both text relevancy and image fidelity.
Further analysis suggests that different knowledge sources improve the generation in different aspects, and using more experts results in better image quality.
In the future, we intend to enrich external image-text alignment knowledge and expand the usage of multiple experts to advance the generation.
See also Appendix for more details on training and evaluation.

\paragraph{Acknowledgement}
This work was partly supported by National Natural Science Foundation of China (62271359).
We thank Yehan Zheng for help on dataset collection; Weixin Liu and Bin Shan for discussions on paper.


{\small
\bibliographystyle{ieee_fullname}
\bibliography{egbib}
}

\newpage

\appendix


\begin{algorithm}[t]
\footnotesize
\caption{Training Process}\label{alg:train}
\begin{algorithmic}[1]
\Require Paired (image, text) inputs, POS toolkit, object detector.
\State Find keywords in text and salient regions in image. \Comment{Pre-processing}
\State For training step $s$ in $(0, 350000]$: \Comment{First stage}
\State \quad Sample a denoising timestep $t$ from $[0, 1000)$,
\State \quad Optimize both the text encoder and the denoising network.
\State Init $10$ denoising networks with current parameters.
\State For training step $s$ in $(350000, 440000]$: \Comment{Second stage}
\State \quad Sample a denoising timestep $t$ from $[0, 1000)$,
\State \quad Optimize the $\lfloor \frac{t}{100} \rfloor$-th denoising network.
\end{algorithmic}
\end{algorithm}

\section{Detailed Training Process}\label{sec:training}

Algorithm~\ref{alg:train} shows the pseudo code for training ERNIE-ViLG 2.0.
With (image, text) pairs as input, we first find the keywords in texts with an open-source POS toolkit ``jieba'' and salient regions in images with an object detector~\cite{DBLP:conf/cvpr/00010BT0GZ18}. These additional information are then used in the knowledge-enhanced training.
The training process consists of two stages. In the first stage, we train a U-Net with 2.2B parameters and a text encoder with 1.3B parameters for $350,000$ steps. In the second stage, the text encoder is shared, and we train $10$ denoising experts for $90,000$ steps that inherit U-Net parameters from the first stage.
The unaccomplished hyper-parameters we use for ERNIE-ViLG~2.0 is provided in Table~\ref{tab:config}.

\begin{table}[t]
  \centering\small
  \caption{Hyperparameters and Configuration of ERNIE-ViLG~2.0.}
    \begin{tabular}{lc}
    \toprule
    \multicolumn{2}{c}{Text Encoder (Transformer)} \\
    \midrule
    Vocab size & 21128 \\
    Text encoder context & 77 \\
    Text encoder width & 2,048 \\
    Text encoder depth & 24 \\
    Text encoder heads & 32 \\
    \bottomrule
    \toprule
    \multicolumn{2}{c}{Denoising Network (U-Net)} \\
    \midrule
    Noise schedule & linear \\
    Diffusion steps & 1,000 \\
    Sampling steps & 50 \\
    Model channels & 512 \\
    Head channels & 64 \\
    Channel multiplier & [1,2,3,4] \\ 
    Attention resolutions & [2,4,8] \\
    ResNet number & 3 \\
    Dropout & 0 \\
    \bottomrule
    \end{tabular}%
  \label{tab:config}%
\end{table}%

\begin{table*}[t]
  \centering\small
  \caption{Detailed comparison of ERNIE-ViLG 2.0 and representative text-to-image generation models on MS-COCO $256 \times 256$ with zero-shot FID-30k.}
    \begin{tabular}{cccc}
    \toprule
    Model & \#params & FID w/o reranking & FID/\#reranking images \\
    \midrule
    DALL-E~\cite{DBLP:conf/icml/RameshPGGVRCS21} & 12B   & 34.6  & 27.5/512 \\
    LDM~\cite{DBLP:journals/corr/abs-2112-10752}  & 1.45B & 12.61 & - \\
    GLIDE~\cite{DBLP:conf/icml/NicholDRSMMSC22}   & 6B    & 12.24 & - \\
    Make-A-Scene~\cite{DBLP:journals/corr/abs-2203-13131} & 4B    & 11.84 & - \\
    DALL-E~2~\cite{DBLP:journals/corr/abs-2204-06125} & 4.5B  & 10.39 & - \\
    Imagen~\cite{DBLP:journals/corr/abs-2205-11487}   & 6.6B  & 7.27  & - \\
    Parti~\cite{DBLP:journals/corr/abs-2206-10789} & 20B   & -     & 7.23/16 \\
    \midrule
    ERNIE-ViLG 2.0 w/ 1 denoising expert & 3.5B  & 8.07  & 7.62/4 \\
    ERNIE-ViLG 2.0 w/ 10 denoising experts & 24B   & \textbf{7.23} &  \textbf{6.75}/4 \\
    \bottomrule
    \end{tabular}%

  \label{tab:fid}%
\end{table*}%

\section{Detailed Automatic Evaluation}\label{sec:detailed_fid}

Table~\ref{tab:fid} presents a detailed comparison on the automatic evaluation scores,  including model sizes and reranking strategies of models.
At the end of the first training stage, the FID-30K metric of our 3.5B model with only one denoising expert is 8.07 (w/o reranking), which is better than DALL-E 2~\cite{DBLP:journals/corr/abs-2204-06125} (10.39) with a similar model size, and worse than our final 24B model with 10 experts (7.23 w/o reranking). 
After the complete training process, our 24B model (6.75 w/ 4 reranking images) outperforms Parti~\cite{DBLP:journals/corr/abs-2206-10789} (7.23 w/ 16 reranking images) with a similar number of parameters and a smaller number of reranking images.
These comparisons indicate that both extra knowledge and model scaling contribute to the final performance of our model.

\begin{table*}[t]
  \centering\small
  \caption{Detailed categories and statistics of ViLG-300.}
    \begin{tabular}{lllr}
    \toprule
    Source & Category & Description & Number \\
    \midrule
    \multirow{8}{*}{DrawBench~\cite{DBLP:journals/corr/abs-2205-11487}} 
    & Color & objects with specified colors  & 22 \\ 
    & Counting & objects with specified numbers & 18 \\ 
    & Positional & objects with specified spatial positioning & 16 \\
    & Conflicting & objects with conflicting interactions & 10 \\
    & Description & complex and long prompts describing an objects & 20 \\
    & DALL-E case & prompts from DALL-E~\cite{DBLP:conf/icml/RameshPGGVRCS21} & 19 \\
    & Marcus & prompts from Marcus et al.~\cite{DBLP:journals/corr/abs-2204-13807} & 9 \\
    & Reddit & prompts from DALL-E~2 Reddit & 36 \\
    \midrule
    \multirow{8}{*}{ERNIE-ViLG~\cite{DBLP:journals/corr/abs-2112-15283}} 
    & Simple & single-object with specified attributes & 18 \\
    & Complex & multi-objects with specified attributes and relationships & 23 \\
    & Counterfactual & objects with impossible interactions or negative words & 23 \\
    & Geography & specific geographic entities & 24 \\
    & View & objects with specified view angles & 16 \\
    & Scene & objects with specified time and scenes & 14 \\
    & Style & objects with specified styles & 16 \\
    & Cartoon & anthropomorphic animals or cartoon characters & 16 \\
    \bottomrule
    \end{tabular}%
  \label{tab:vilg300}%
\end{table*}%

\begin{table*}[t]
  \small
  \centering
  \caption{Top five ViLG-300 categories with the maximum CLIP Score improvement for each knowledge enhancement strategy.}
    \begin{tabular}{ccccccc}
    \toprule
    \multicolumn{1}{c}{\multirow{2}[4]{*}{No.}} & \multicolumn{2}{c}{w/ textual} & \multicolumn{2}{c}{w/ visual} & \multicolumn{2}{c}{w/ all} \\
\cmidrule{2-7}    \multicolumn{1}{c}{} & Prompt category & \multicolumn{1}{c}{$\Delta$CLIP Score} & Prompt category & \multicolumn{1}{c}{$\Delta$CLIP Score} & Prompt category & $\Delta$CLIP Score \\
    \midrule
    1     & Counterfactual & 0.0051  & Counterfactual & 0.0080  & Complex & 0.0074  \\
    2     & Color  & 0.0041  & Counting & 0.0047  & Counterfactual & 0.0073  \\
    3     & Marcus & 0.0038  & Color  & 0.0035  & Cartoon & 0.0069  \\
    4     & Style & 0.0022  & Cartoon & 0.0032  & Color  & 0.0066  \\
    5     & Positional & 0.0018  & Complex & 0.0016  & Style & 0.0061  \\
    \bottomrule
    \end{tabular}%
  \label{tab:vilg300-clip}%
\end{table*}%

\section{Detailed Human Evaluation}\label{appx:human}

In this section, we supplement the part about human evaluation omitted in main content, including the construction process of ViLG-300, the performance comparison on various categories, and the example qualitative comparison of different models.

\subsection{The Construction of ViLG-300}\label{appx:vilg300}
To construct ViLG-300, we first remove the language-related prompts in DrawBench~\cite{DBLP:journals/corr/abs-2205-11487} (\verb|text| \verb|rendering|, \verb|rare| \verb|words|, \verb|misspelled| prompts) and \verb|MS|-\verb|COCO| prompts in ERNIE-ViLG~\cite{DBLP:journals/corr/abs-2112-15283}, leaving 162 and 398 prompts, respectively, then randomly sampled 150 prompts from these two parts, manually translated and proofread these prompts to achieve the final parallel Chinese and English set. 
Specifically, we remove language-related text prompts in DrawBench since these are not comparable inputs for models in different languages. 
For text rendering in Chinese, we 
also have discussed in detail in Section~\ref{sec:risk}. 
We also remove the MS-COCO category in ERNIE-ViLG, because MS-COCO has been used in the automatic evaluation, and the prompts are relatively simple for current text-to-image models, especially when evaluating the models' ability to understand complex scene.
Note that there are two similar categories (i.e., \verb|Conflicting| and \verb|Counterfactual|) in DrawBench and ERNIE-ViLG that we do not align and merge. The reason is that the \verb|Conflicting| category focuses on the impossible combination of things, while \verb|Counterfactual| contains many prompts with negative descriptions, both of which are now difficult problems.

\subsection{Detailed Results on ViLG-300}\label{appx:result}
\begin{figure}[t]
    \centering
    \begin{subfigure}{\linewidth}
        \includegraphics[width=\linewidth]{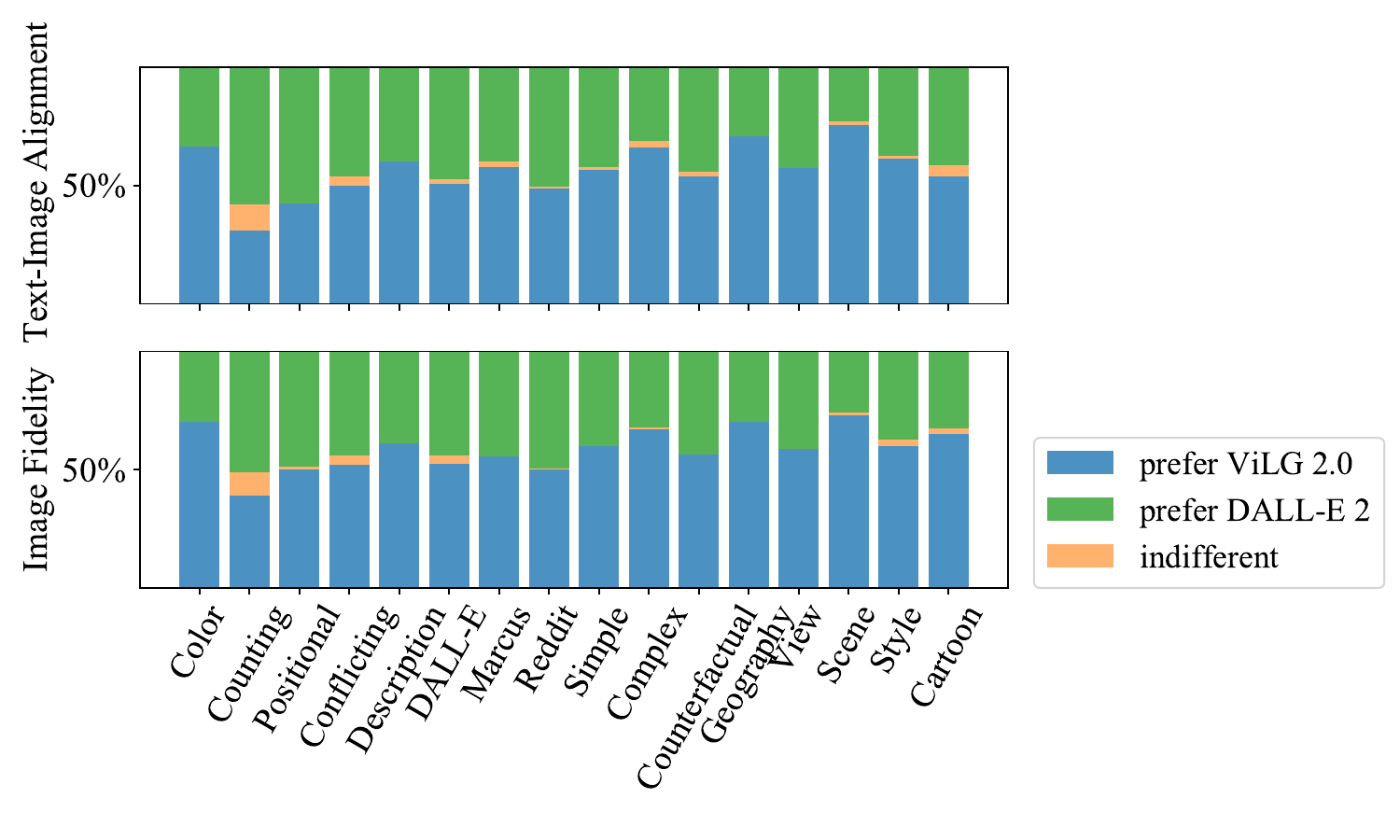}
        \caption{ERNIE-ViLG~2.0 v.s. DALL-E~2}
        \label{fig:human_detail_dalle}
    \end{subfigure}
    \begin{subfigure}{\linewidth}
        \includegraphics[width=\linewidth]{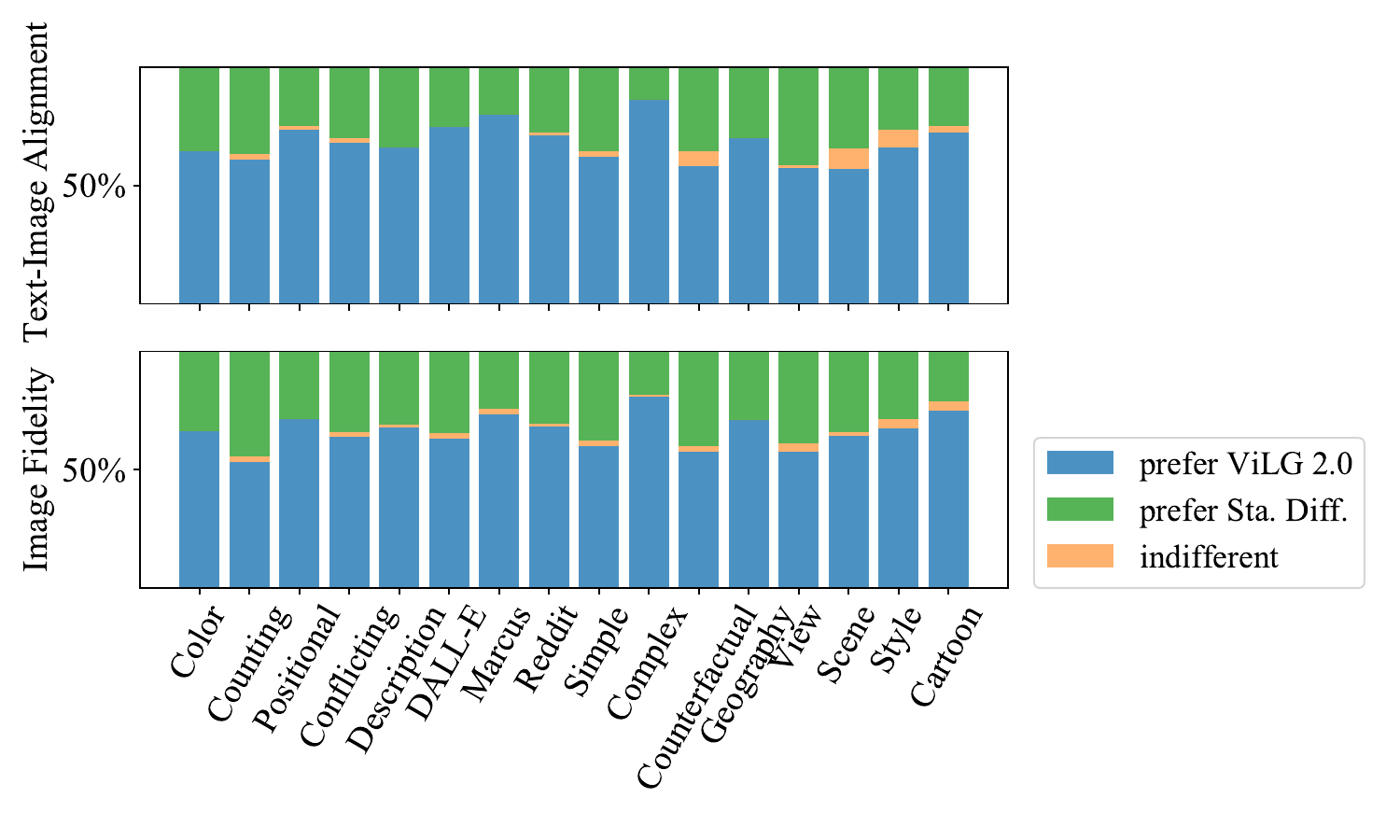}
        \caption{ERNIE-ViLG~2.0 v.s. Stable Diffusion}
        \label{fig:human_detail_sd}
    \end{subfigure}
    \caption{Detailed comparison of ERNIE-ViLG~2.0 and DALL-E~2/Stable Diffusion on ViLG-300 with human evaluation. We do not apply any filtering strategy and report the initial results here.}
    \label{fig:human_detail}
\end{figure}

\begin{figure*}[htbp]
\centering
\begin{tabular}{c@{\hskip 2.5pt}c@{\hskip 8pt}c@{\hskip 2.5pt}c}
\multicolumn{2}{c}{\scriptsize 一个绿色的杯子和一个蓝色的手机} & \multicolumn{2}{c}{\scriptsize 一个红酒杯放在一条狗上面} \\
\multicolumn{2}{c}{\scriptsize A green cup and a blue cell phone} & 
\multicolumn{2}{c}{\scriptsize A wine glass on top of a dog} \\
\multicolumn{4}{c}{\scriptsize ERNIE-ViLG~2.0} \\
{\includegraphics[width=0.18\linewidth]{figs/comp_case/00015/00015_1_vilg.jpg}} &
{\includegraphics[width=0.18\linewidth]{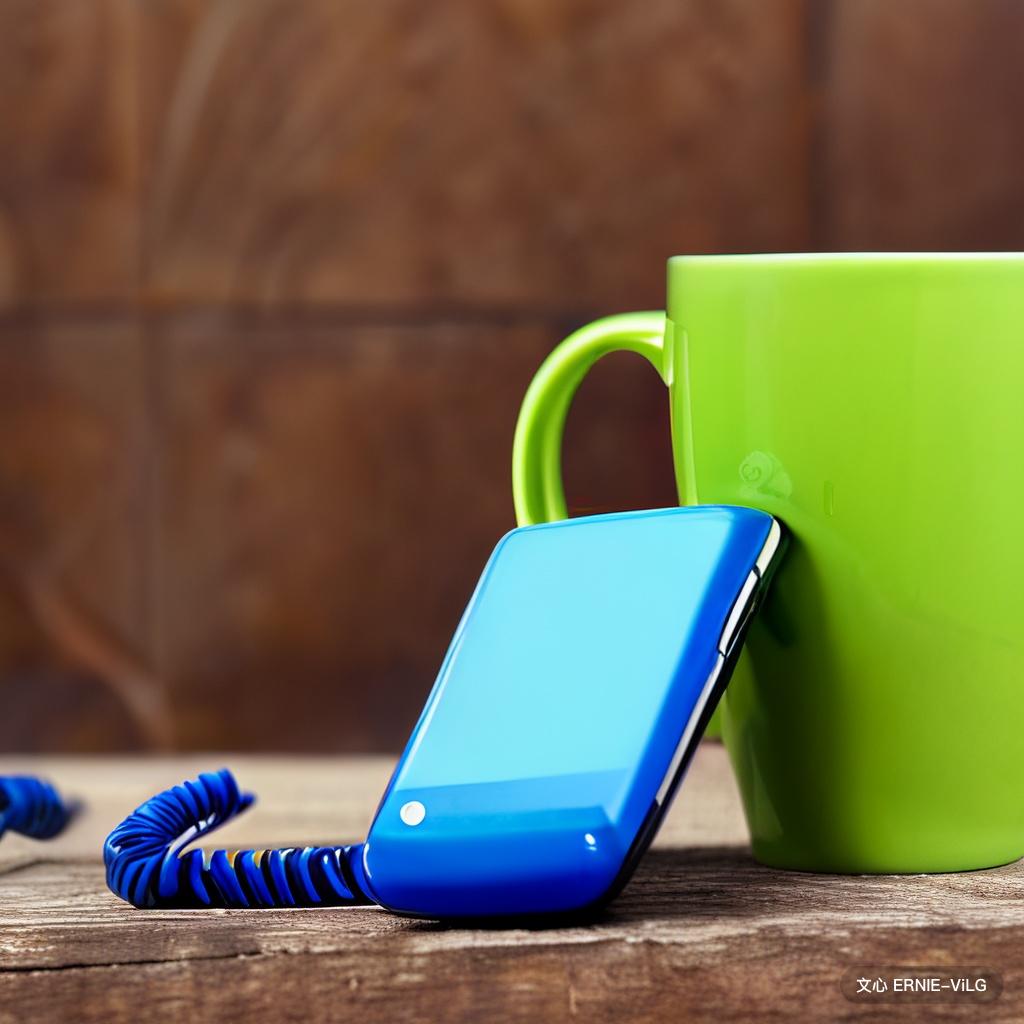}} &
{\includegraphics[width=0.18\linewidth]{figs/comp_case/00098/00098_0_vilg.jpg}} &
{\includegraphics[width=0.18\linewidth]{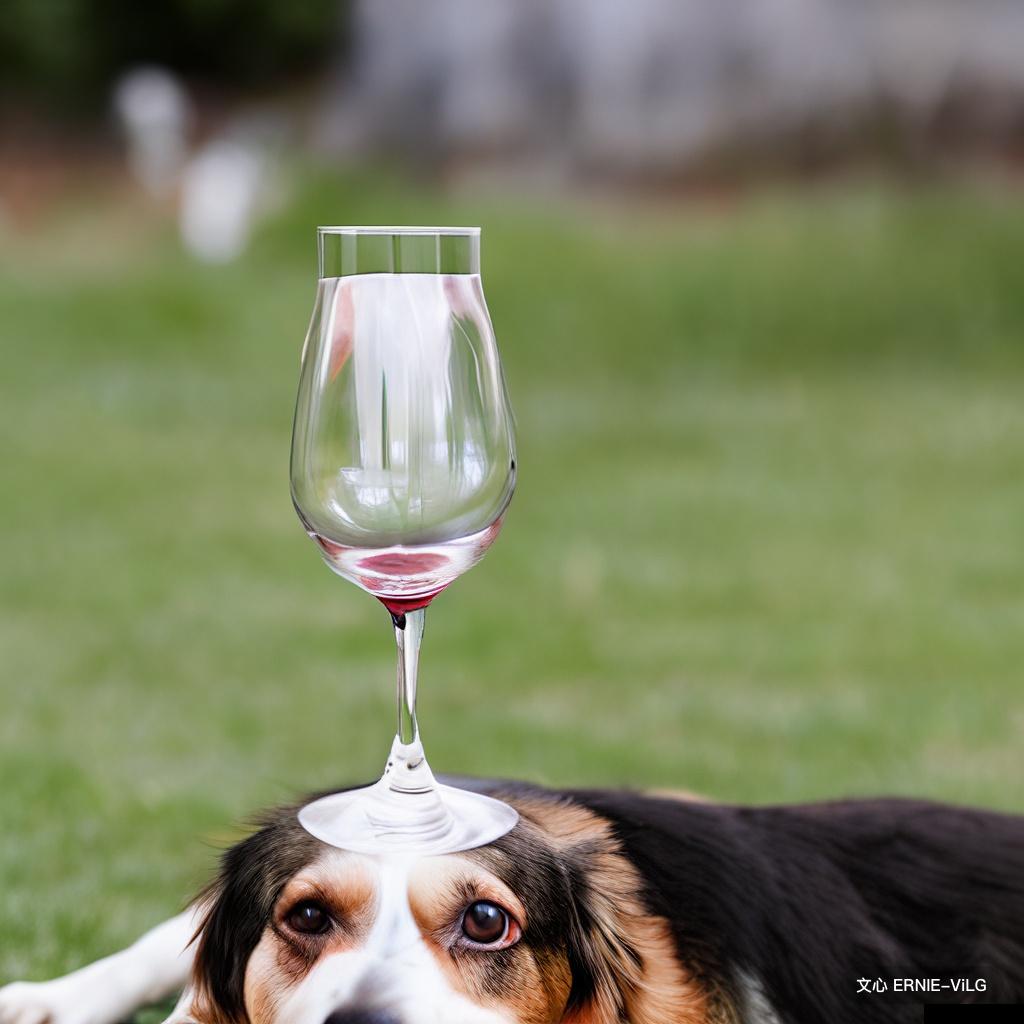}} \\
{\includegraphics[width=0.18\linewidth]{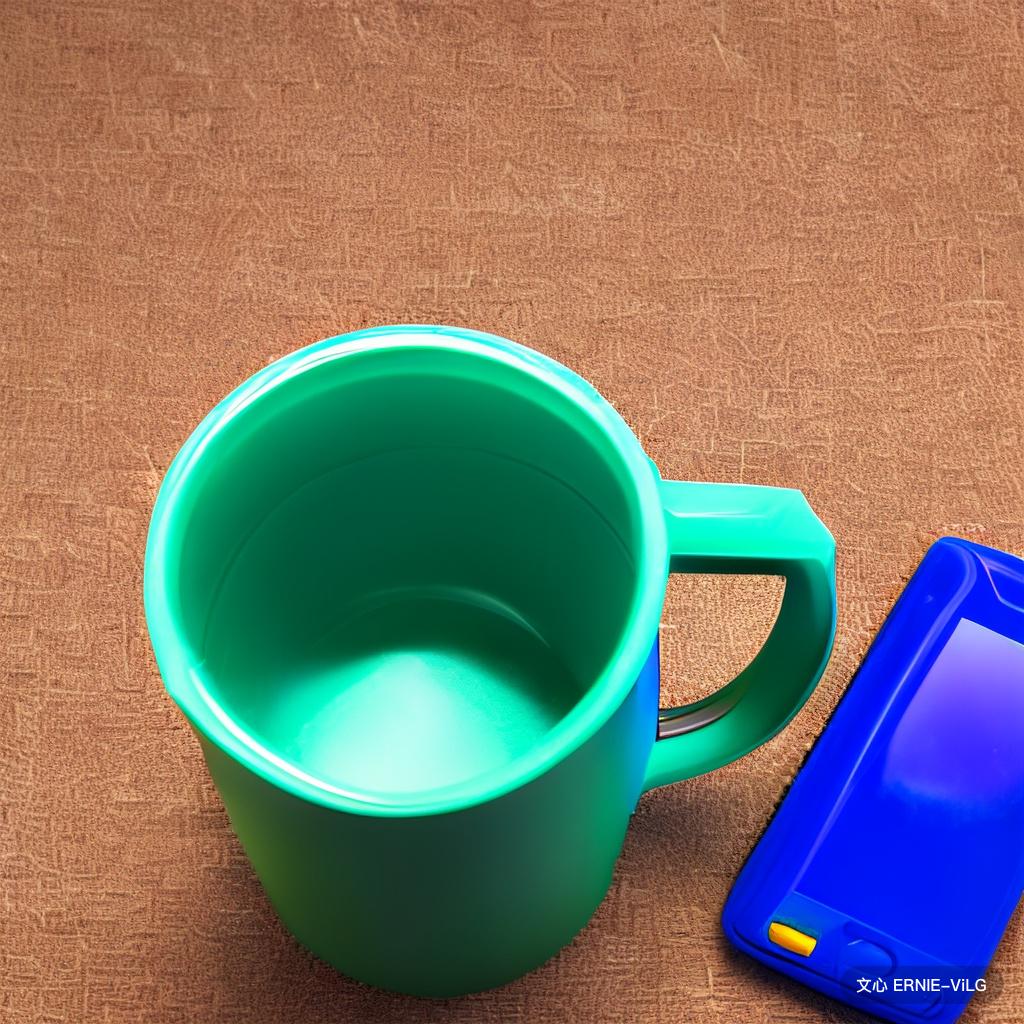}} &
{\includegraphics[width=0.18\linewidth]{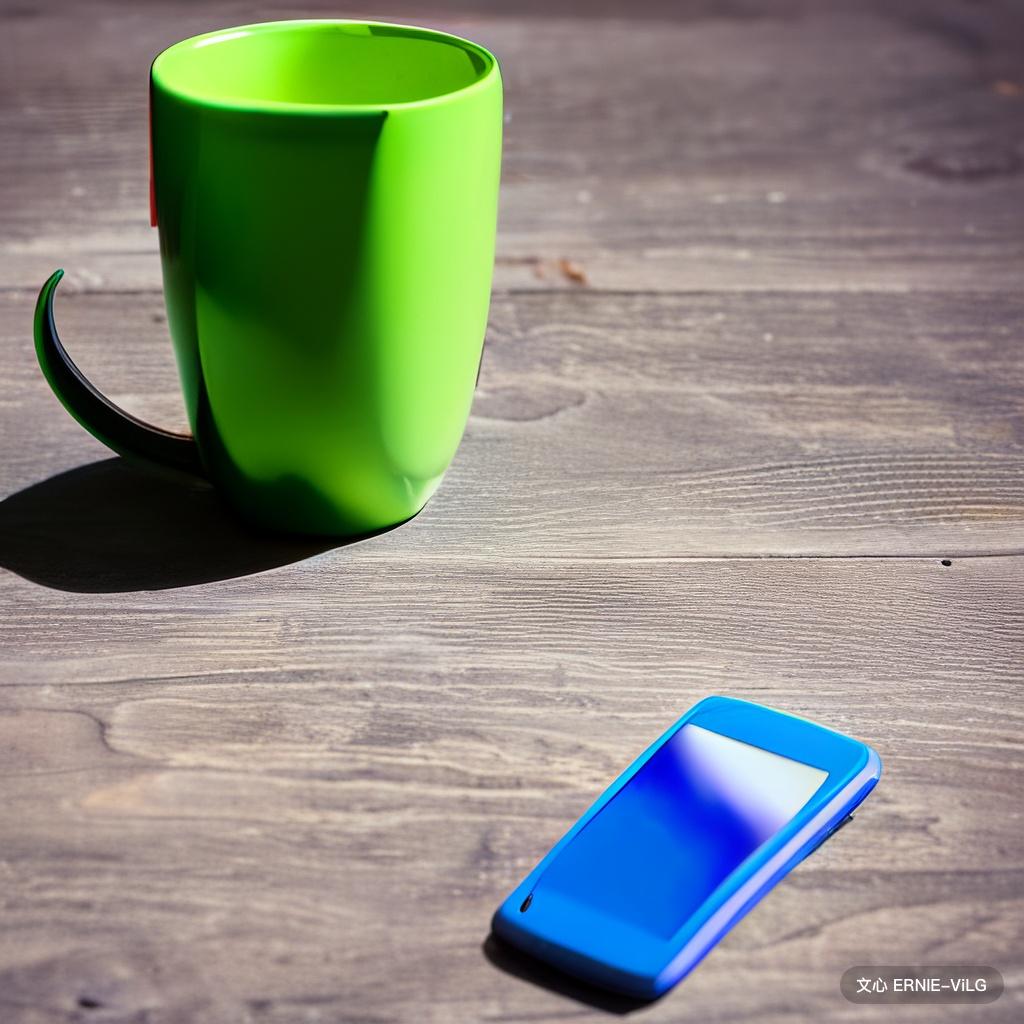}} &
{\includegraphics[width=0.18\linewidth]{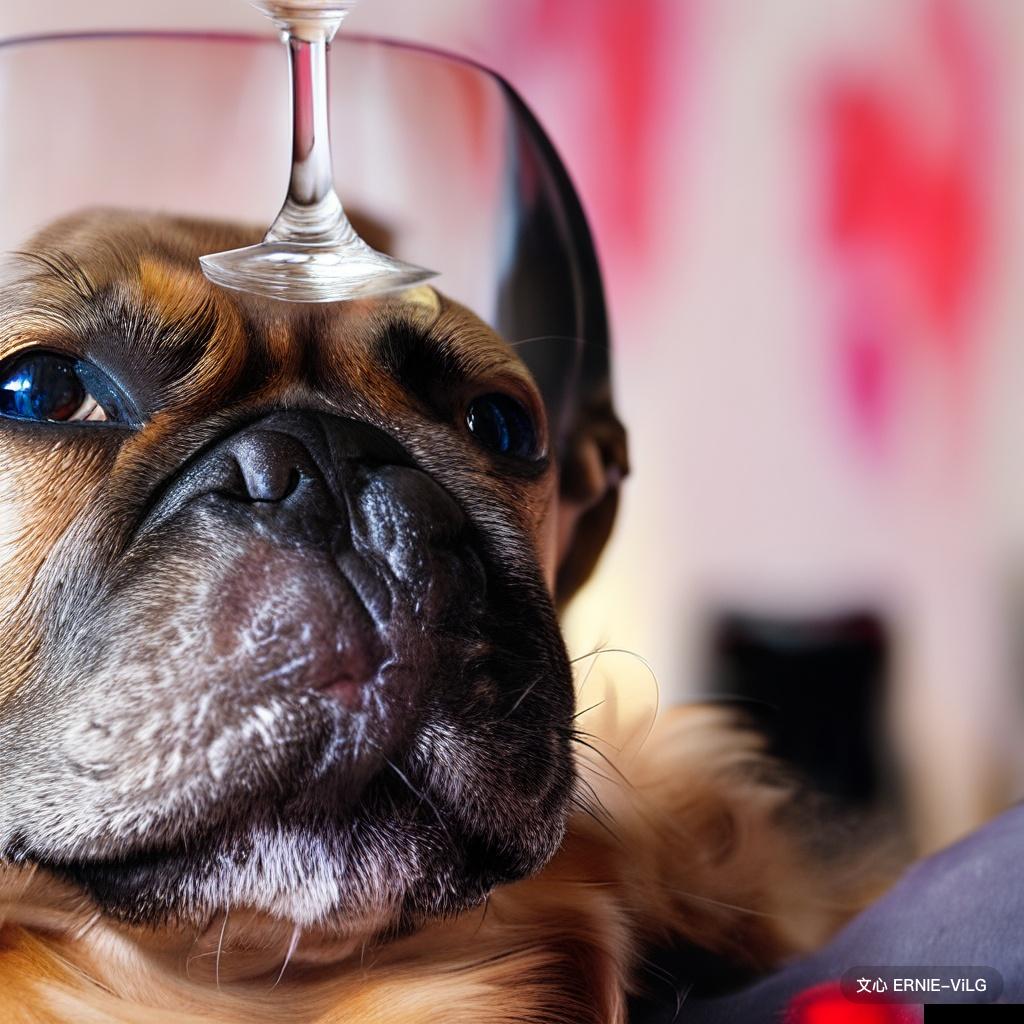}} &
{\includegraphics[width=0.18\linewidth]{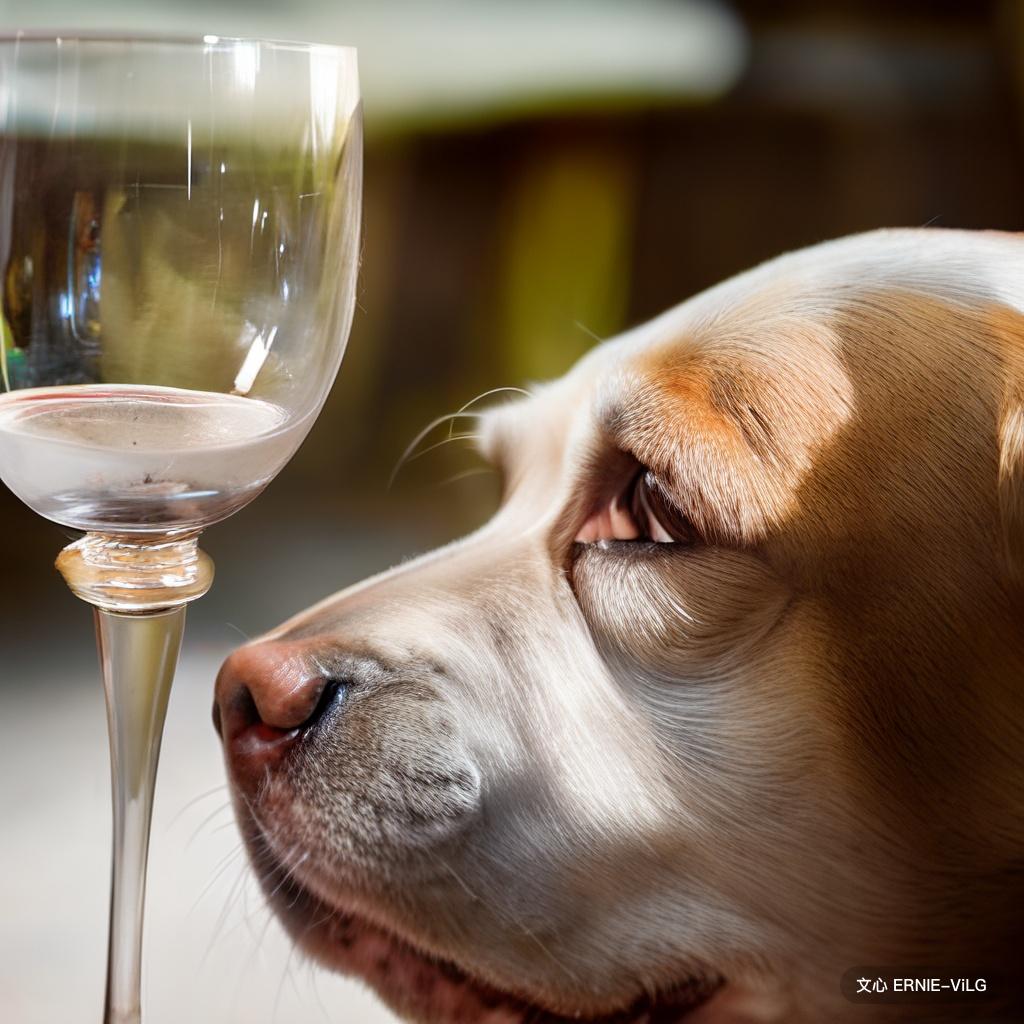}} \\
\multicolumn{4}{c}{\scriptsize DALL-E~2} \\
{\includegraphics[width=0.18\linewidth]{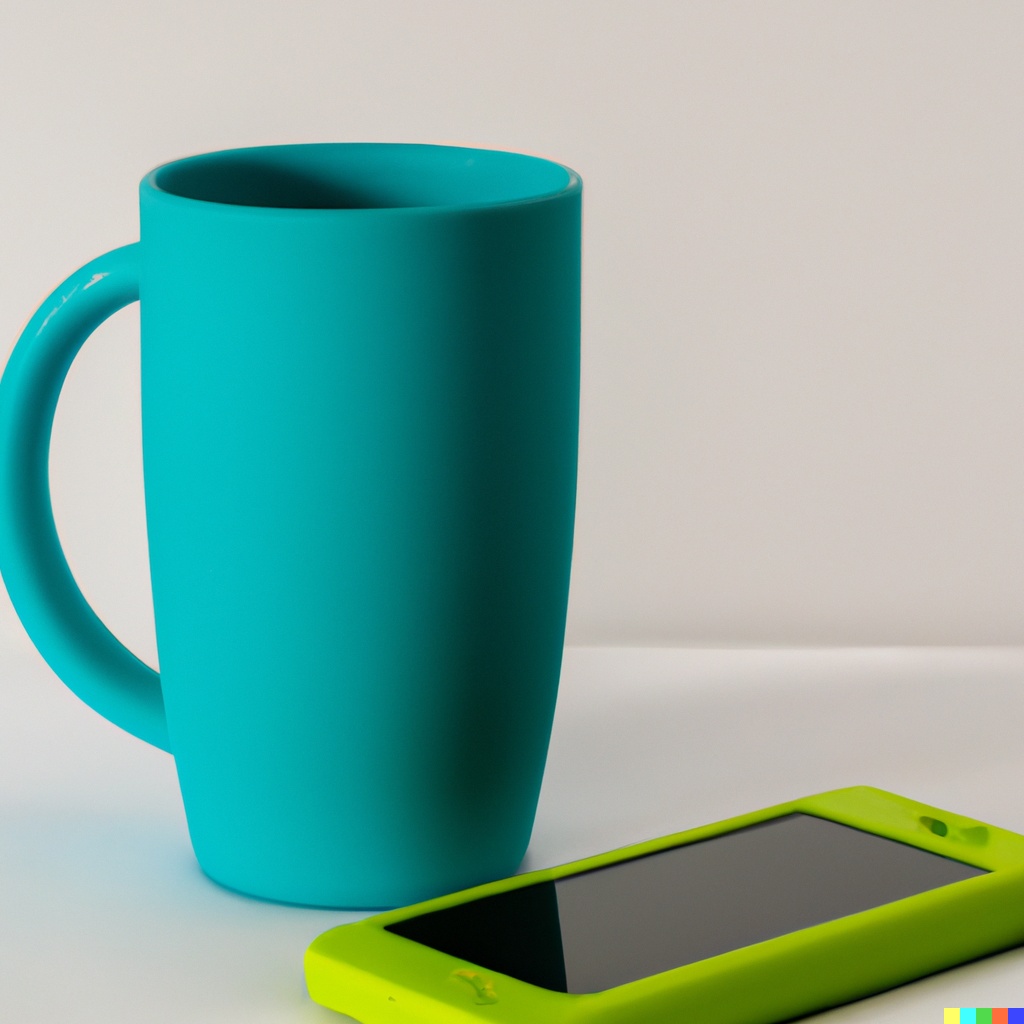}} &
{\includegraphics[width=0.18\linewidth]{figs/comp_case/00015/00015_1_dalle.jpg}} &
{\includegraphics[width=0.18\linewidth]{figs/comp_case/00098/00098_0_dalle.jpg}} &
{\includegraphics[width=0.18\linewidth]{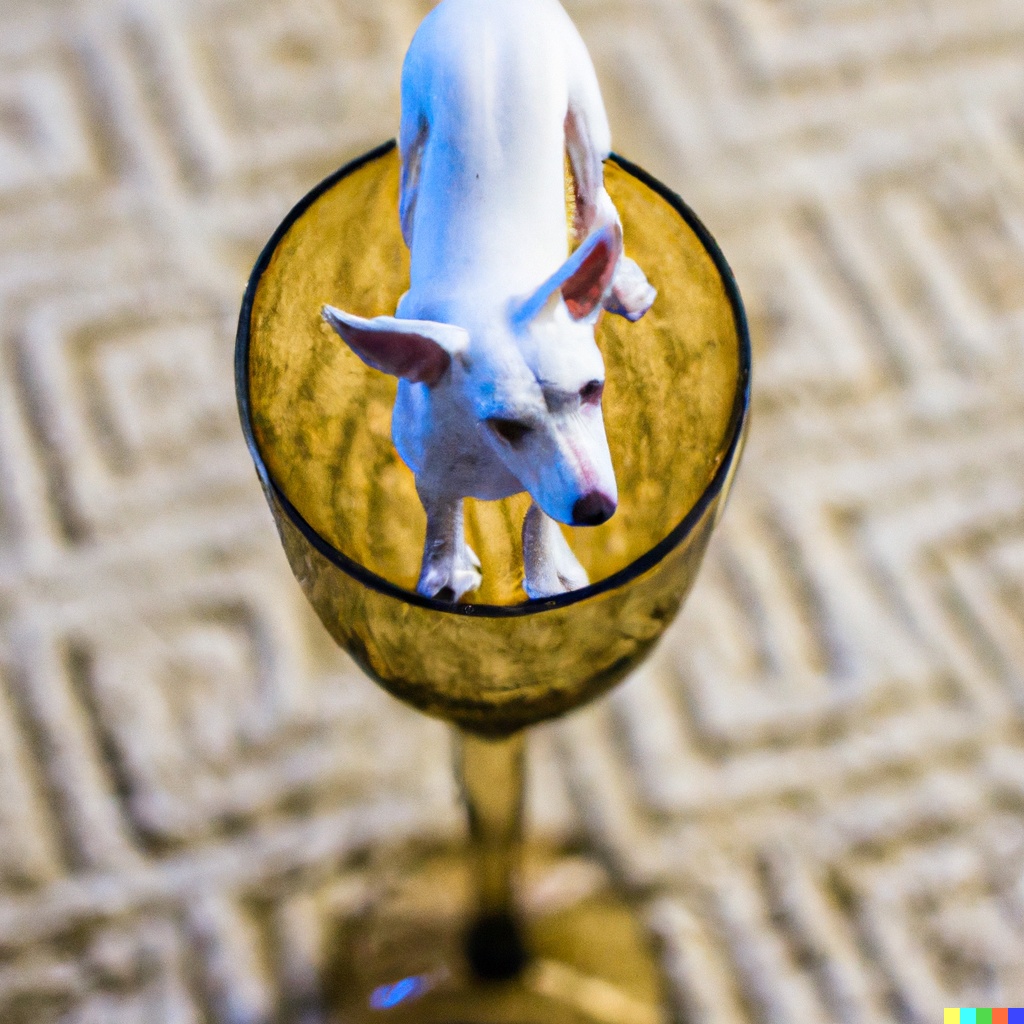}} \\
{\includegraphics[width=0.18\linewidth]{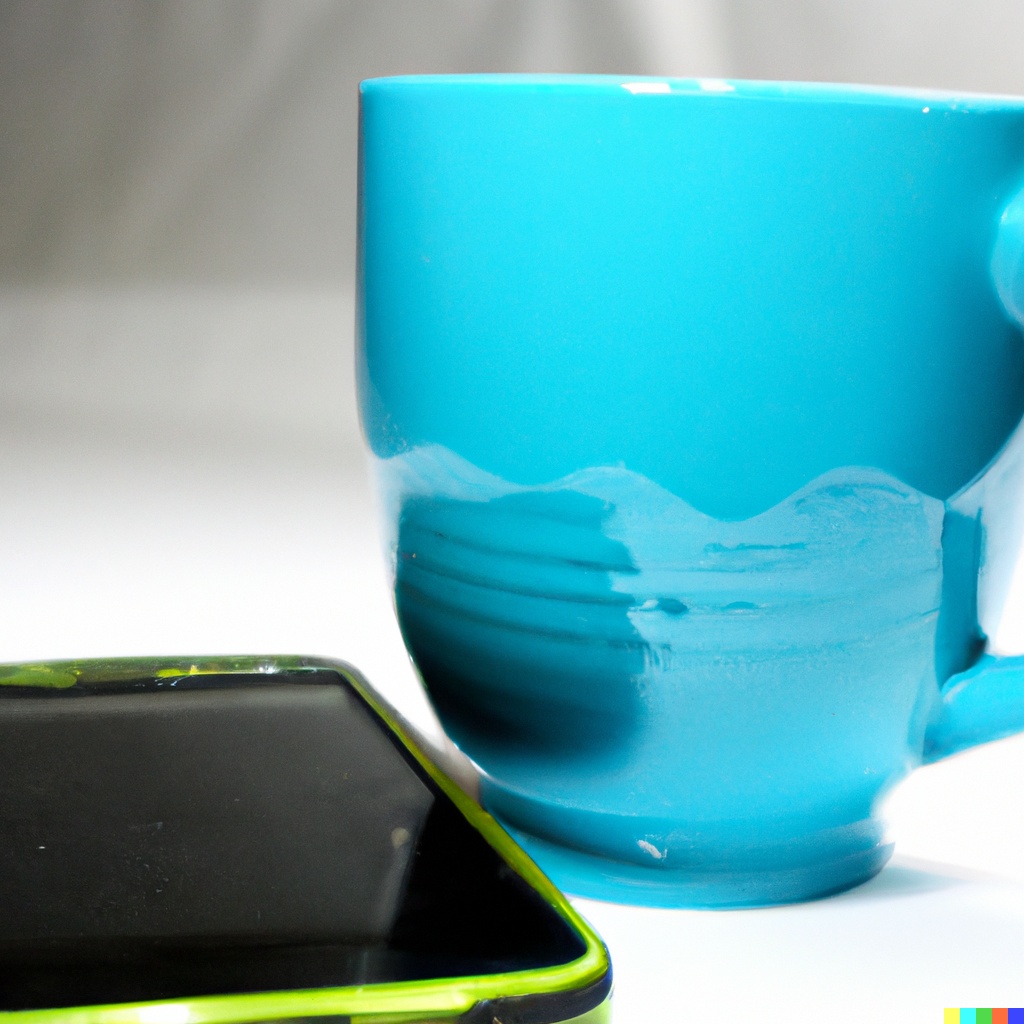}} &
{\includegraphics[width=0.18\linewidth]{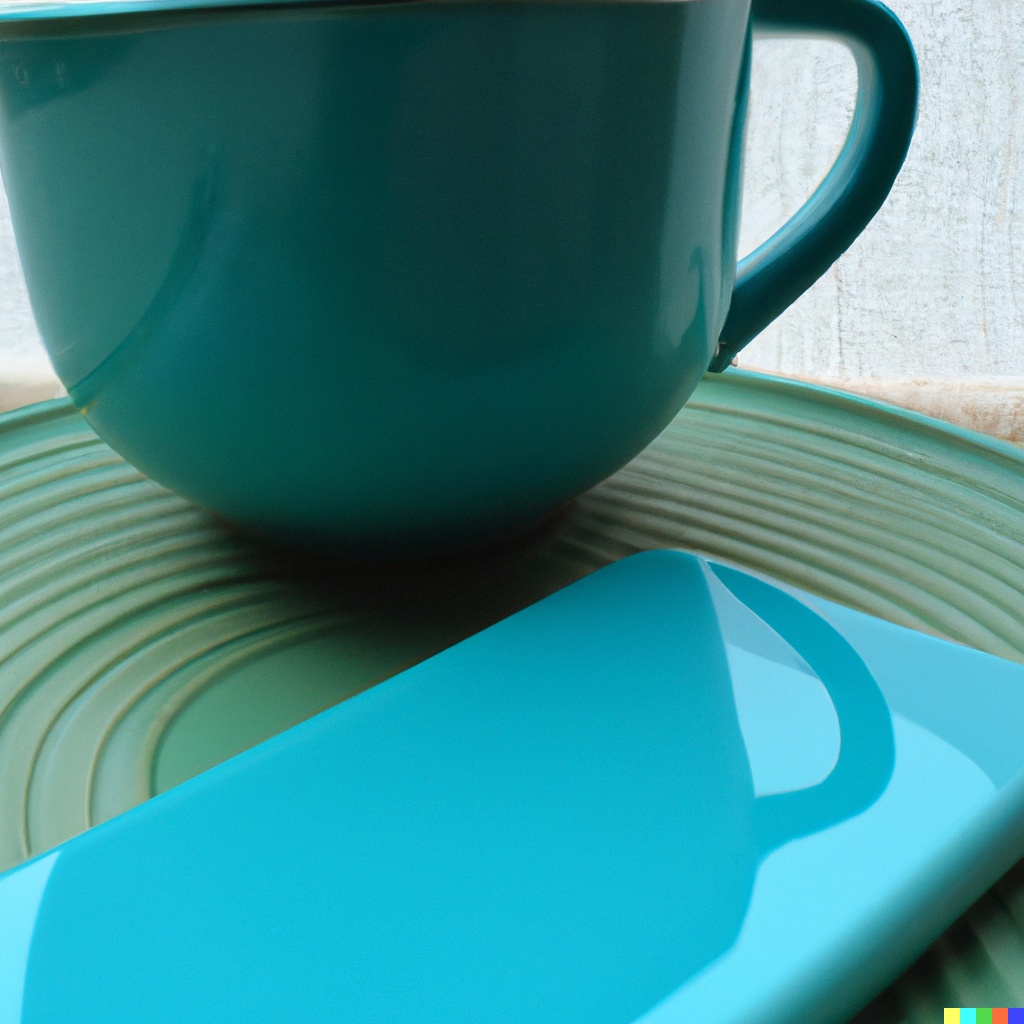}} &
{\includegraphics[width=0.18\linewidth]{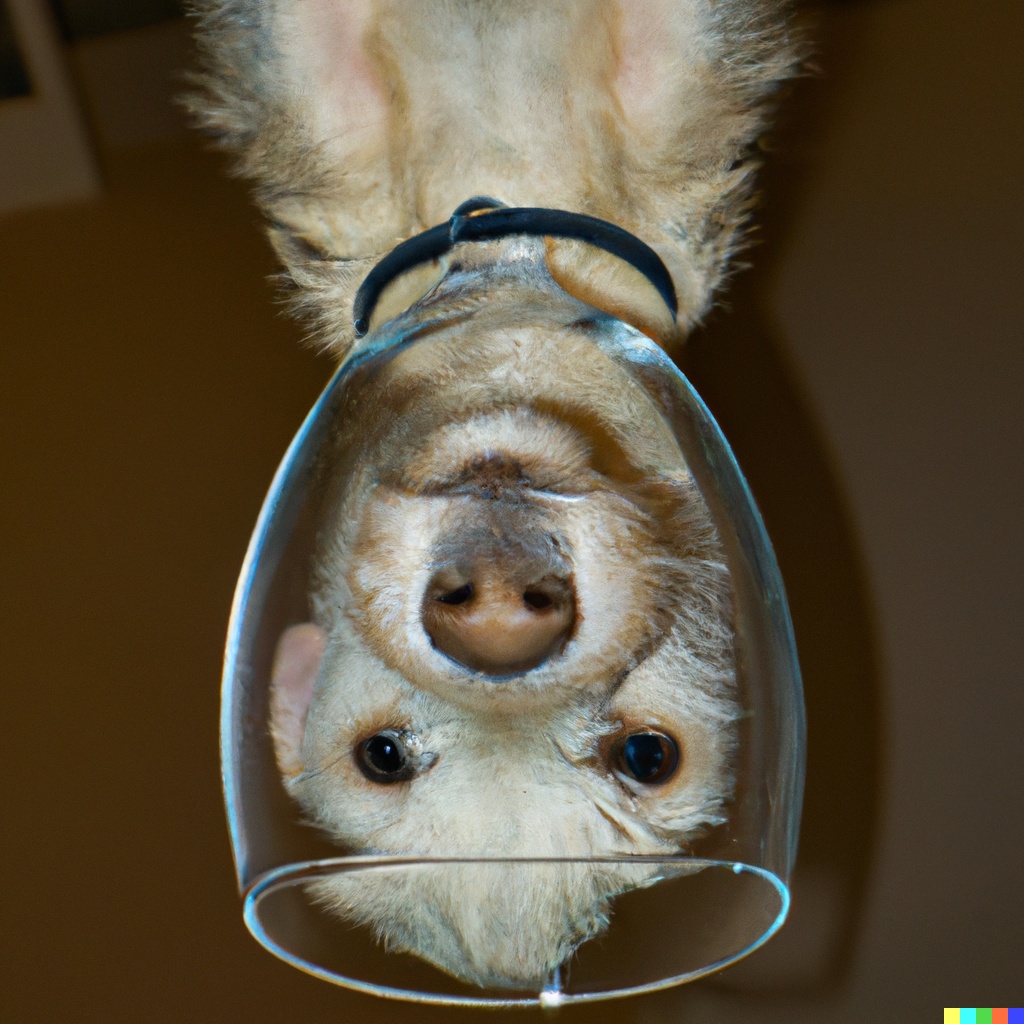}} &
{\includegraphics[width=0.18\linewidth]{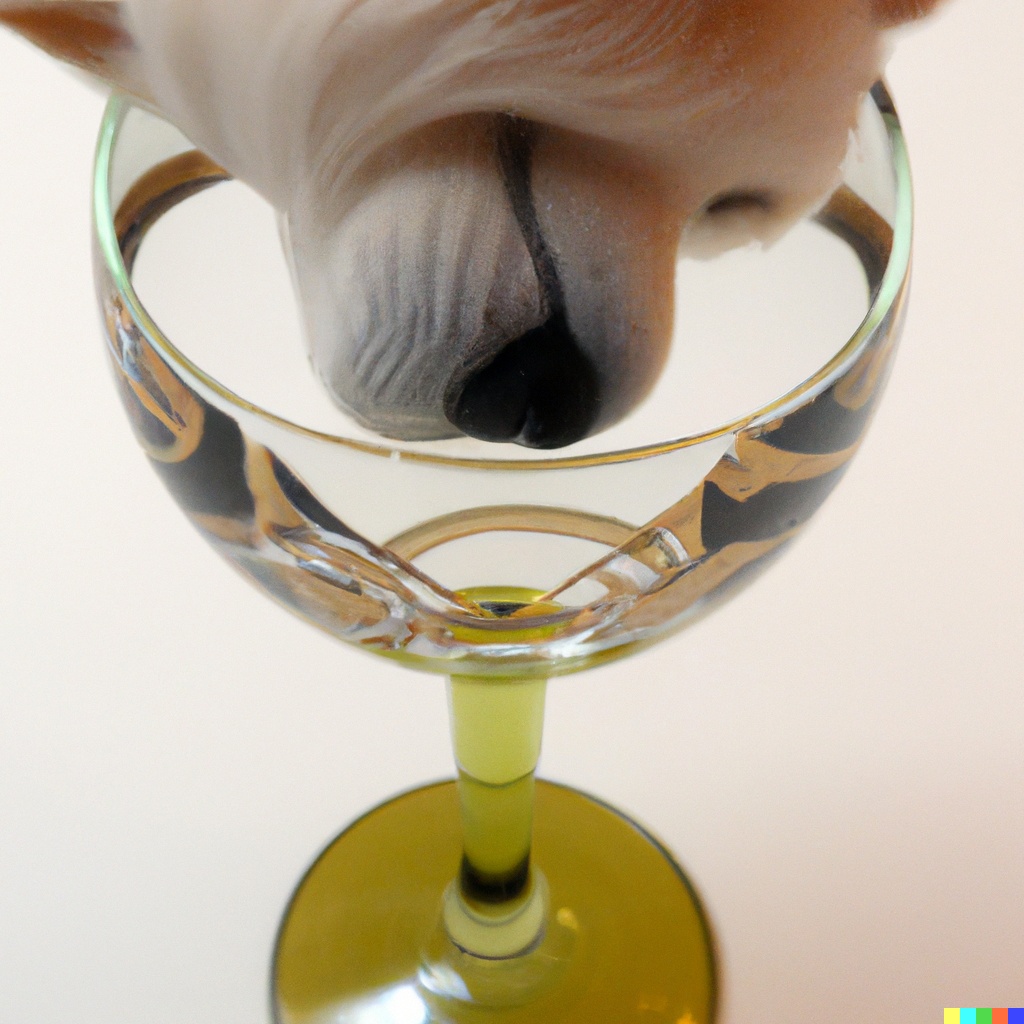}} \\
\multicolumn{4}{c}{\scriptsize Stable Diffusion} \\
{\includegraphics[width=0.18\linewidth]{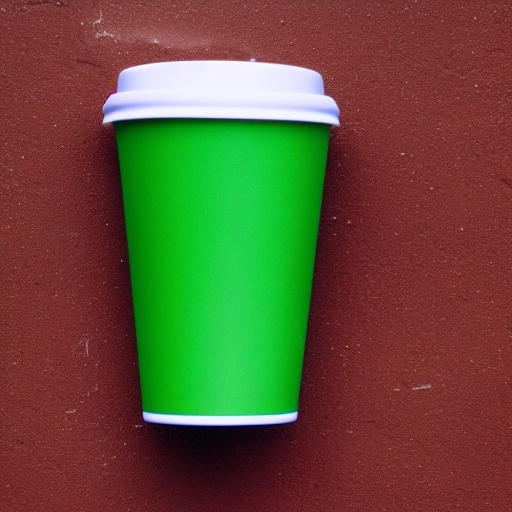}} &
{\includegraphics[width=0.18\linewidth]{figs/comp_case/00015/00015_1_sd.jpg}} &
{\includegraphics[width=0.18\linewidth]{figs/comp_case/00098/00098_0_sd.jpg}} &
{\includegraphics[width=0.18\linewidth]{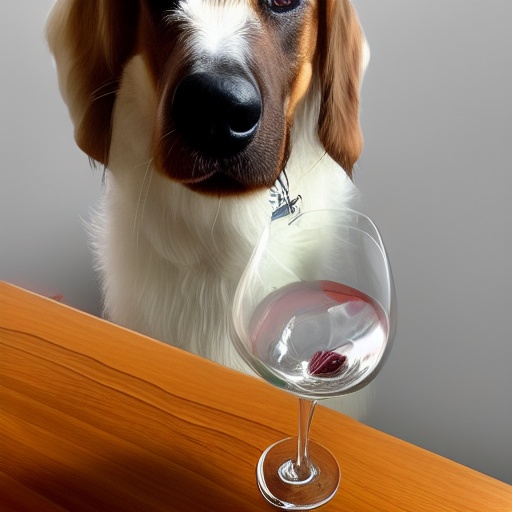}} \\
{\includegraphics[width=0.18\linewidth]{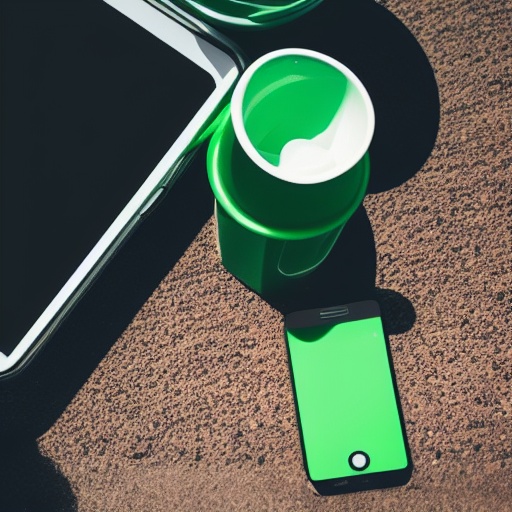}} &
{\includegraphics[width=0.18\linewidth]{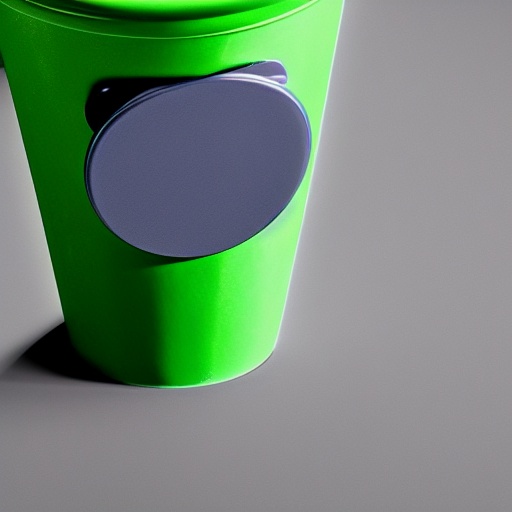}} &
{\includegraphics[width=0.18\linewidth]{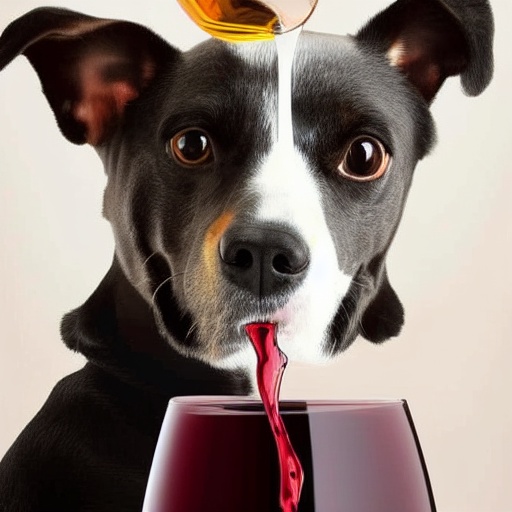}} &
{\includegraphics[width=0.18\linewidth]{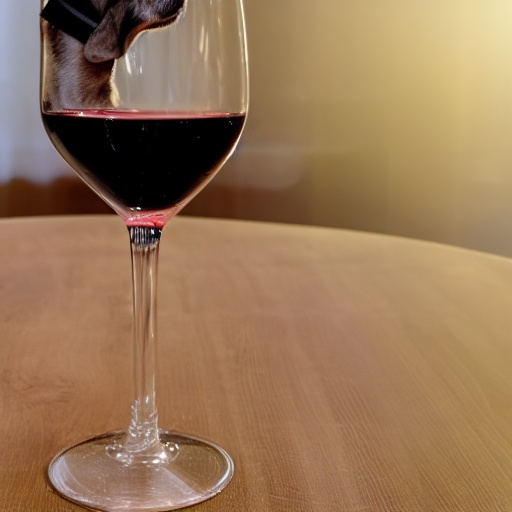}} \\
\end{tabular}
\caption{Example qualitative comparisons between ERNIE-ViLG~2.0 and DALL-E~2/Stable Diffusion on DrawBench prompts from ViLG-300.}
\label{fig:case_drawbench}
\end{figure*}

\begin{figure*}[htbp]
\centering
\begin{tabular}{c@{\hskip 2.5pt}c@{\hskip 8pt}c@{\hskip 2.5pt}c}
\multicolumn{2}{c}{\scriptsize 锅里煮着粽子和玉米} & 
\multicolumn{2}{c}{\scriptsize 樱花数字油画} \\
\multicolumn{2}{c}{\scriptsize Zongzi and corn boiled in the pot} & 
\multicolumn{2}{c}{\scriptsize Cherry blossom, digital oil painting} \\
\multicolumn{4}{c}{\scriptsize ERNIE-ViLG~2.0} \\
{\includegraphics[width=0.18\linewidth]{figs/comp_case/00190/00190_0_vilg.jpg}} &
{\includegraphics[width=0.18\linewidth]{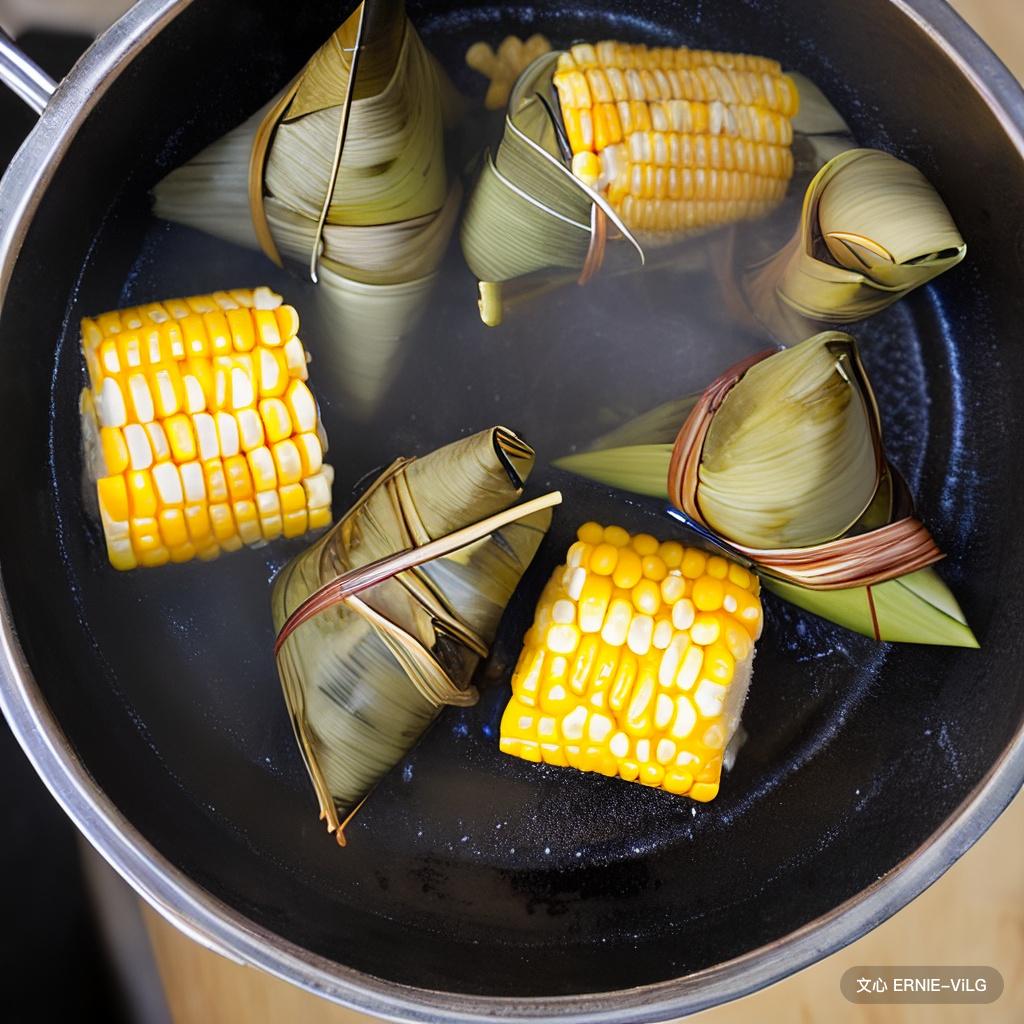}} &
{\includegraphics[width=0.18\linewidth]{figs/comp_case/00276/00276_0_vilg.jpg}} &
{\includegraphics[width=0.18\linewidth]{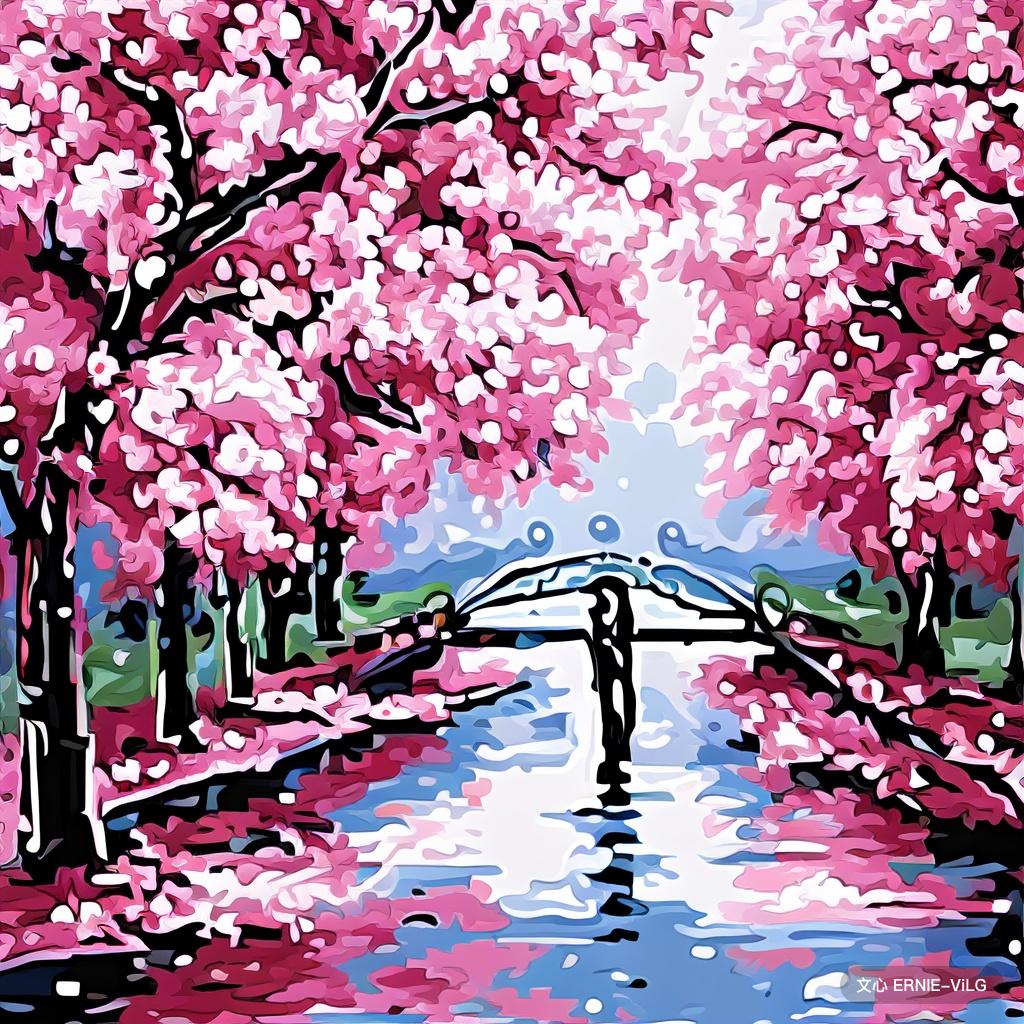}} \\
{\includegraphics[width=0.18\linewidth]{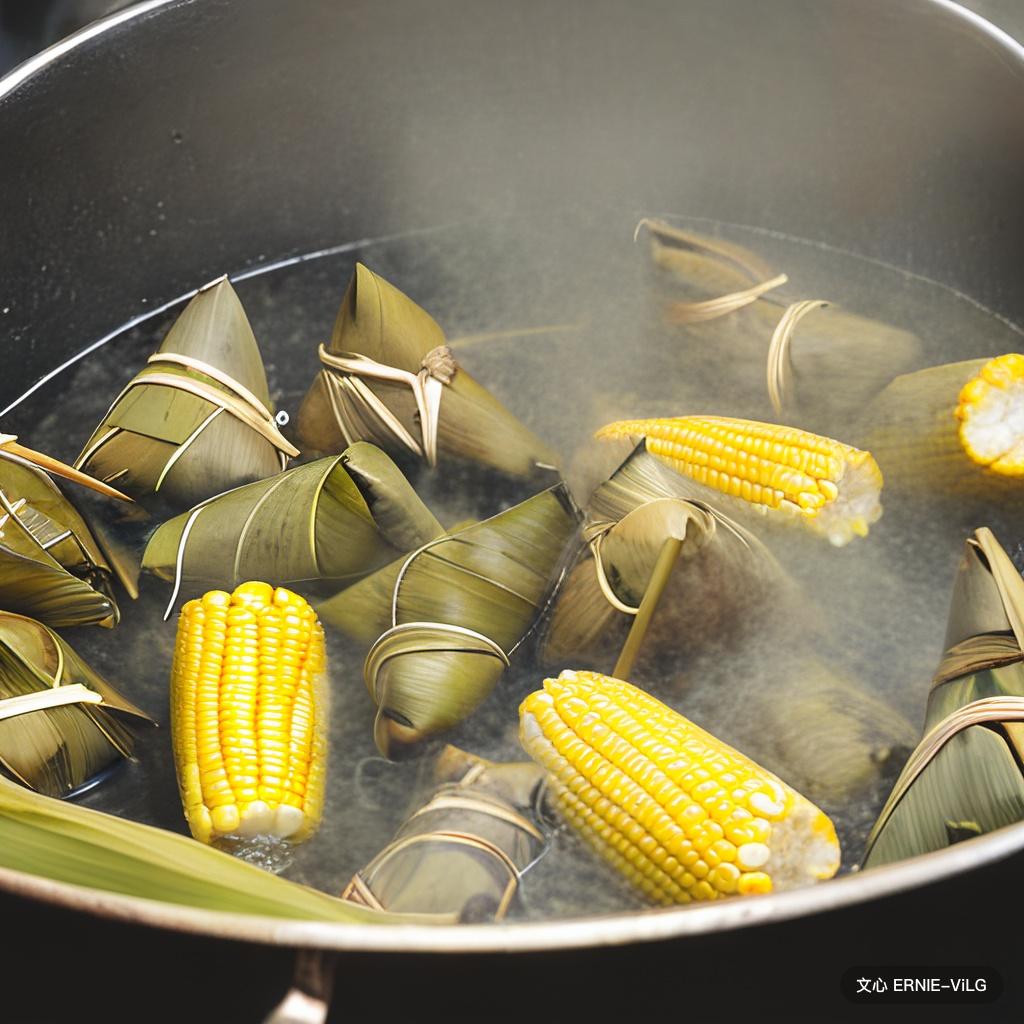}} &
{\includegraphics[width=0.18\linewidth]{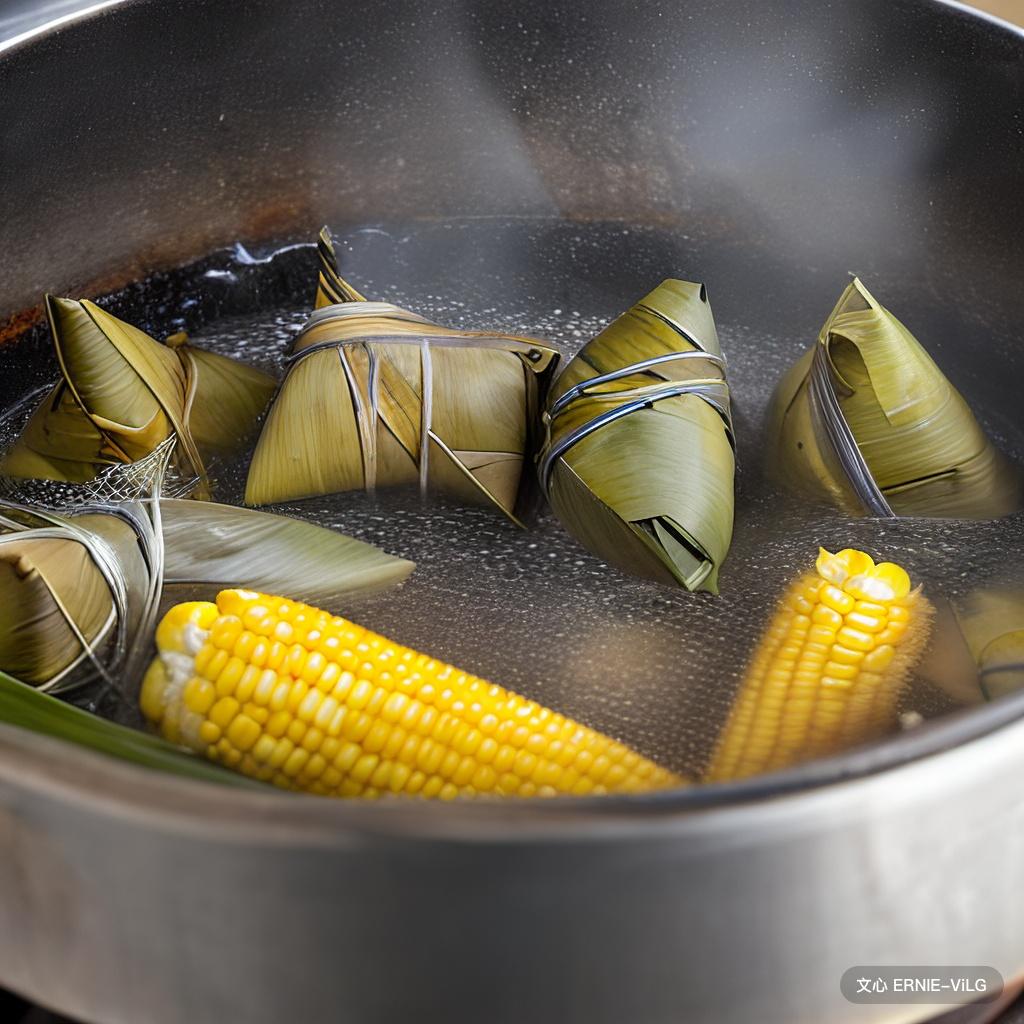}} &
{\includegraphics[width=0.18\linewidth]{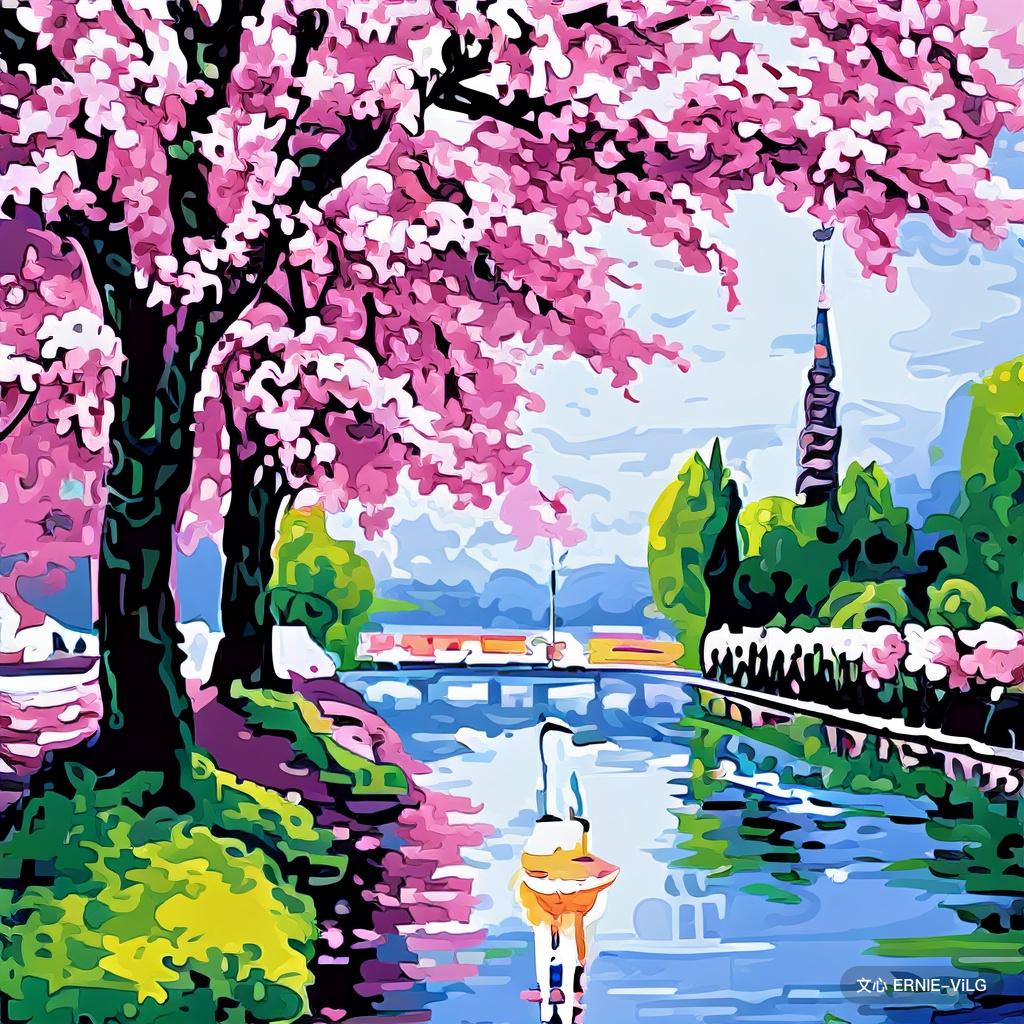}} &
{\includegraphics[width=0.18\linewidth]{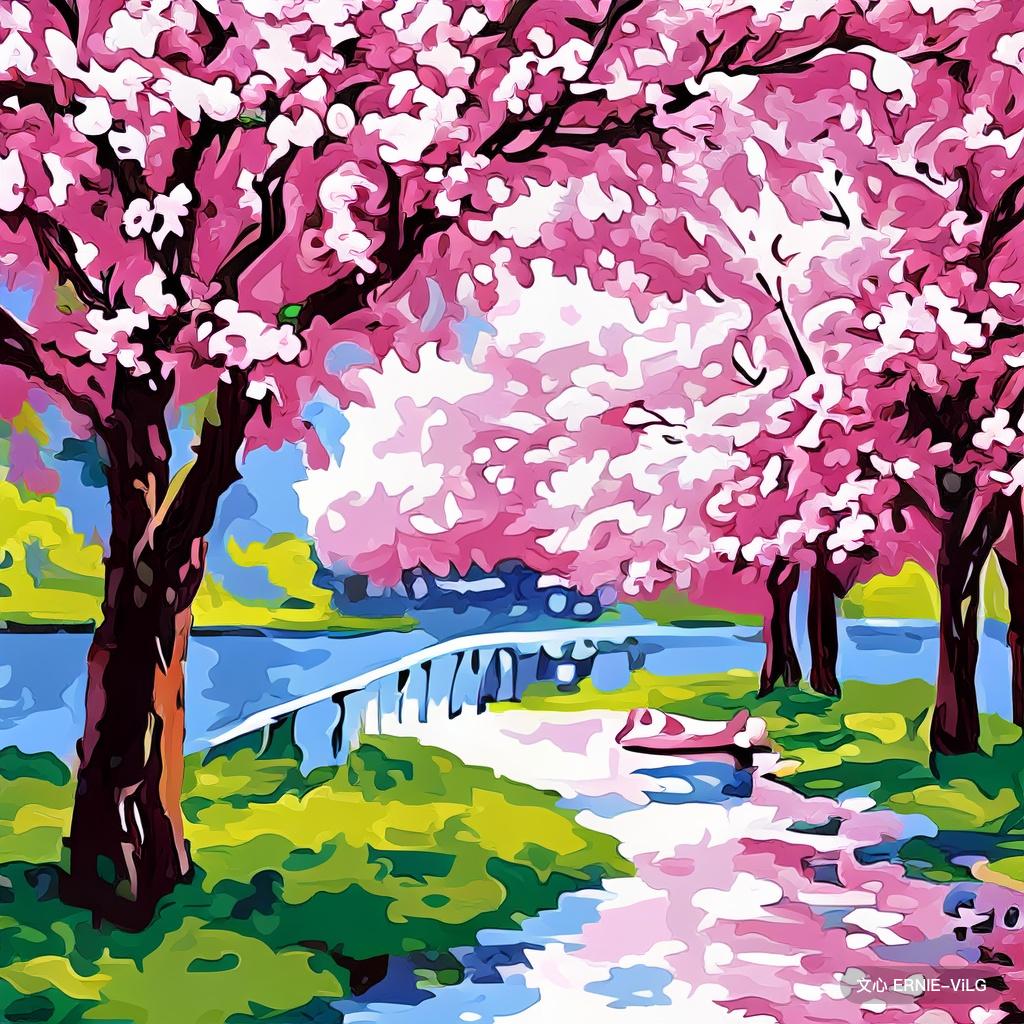}} \\
\multicolumn{4}{c}{\scriptsize DALL-E~2} \\
{\includegraphics[width=0.18\linewidth]{figs/comp_case/00190/00190_0_dalle.jpg}} &
{\includegraphics[width=0.18\linewidth]{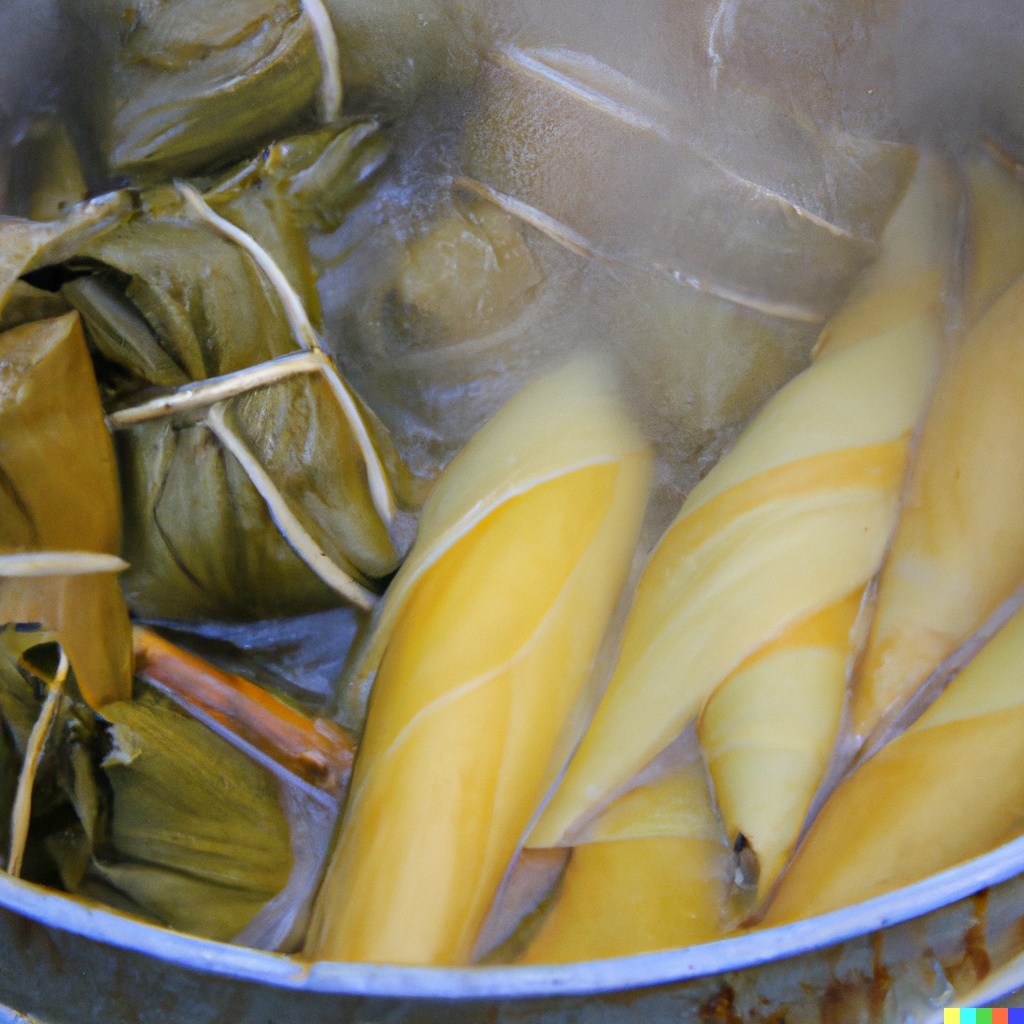}} &
{\includegraphics[width=0.18\linewidth]{figs/comp_case/00276/00276_0_dalle.jpg}} &
{\includegraphics[width=0.18\linewidth]{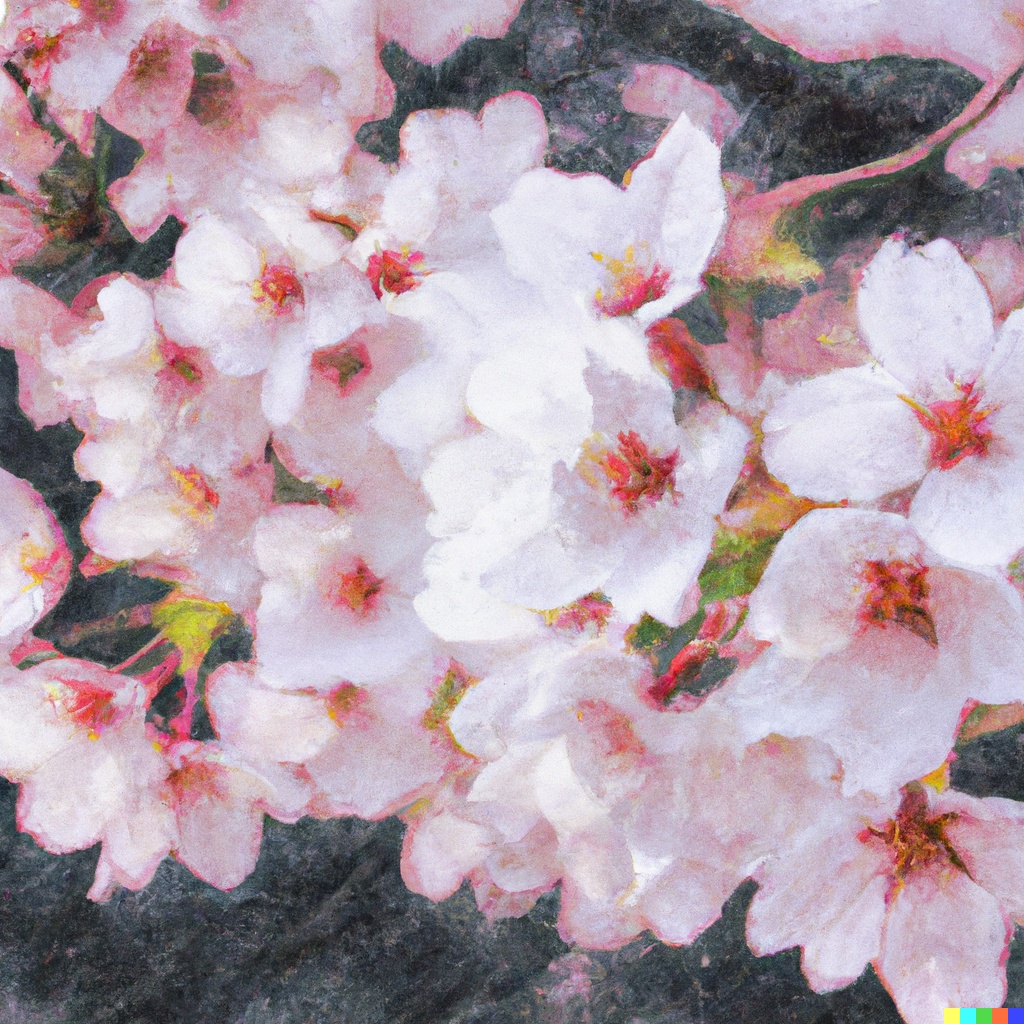}} \\
{\includegraphics[width=0.18\linewidth]{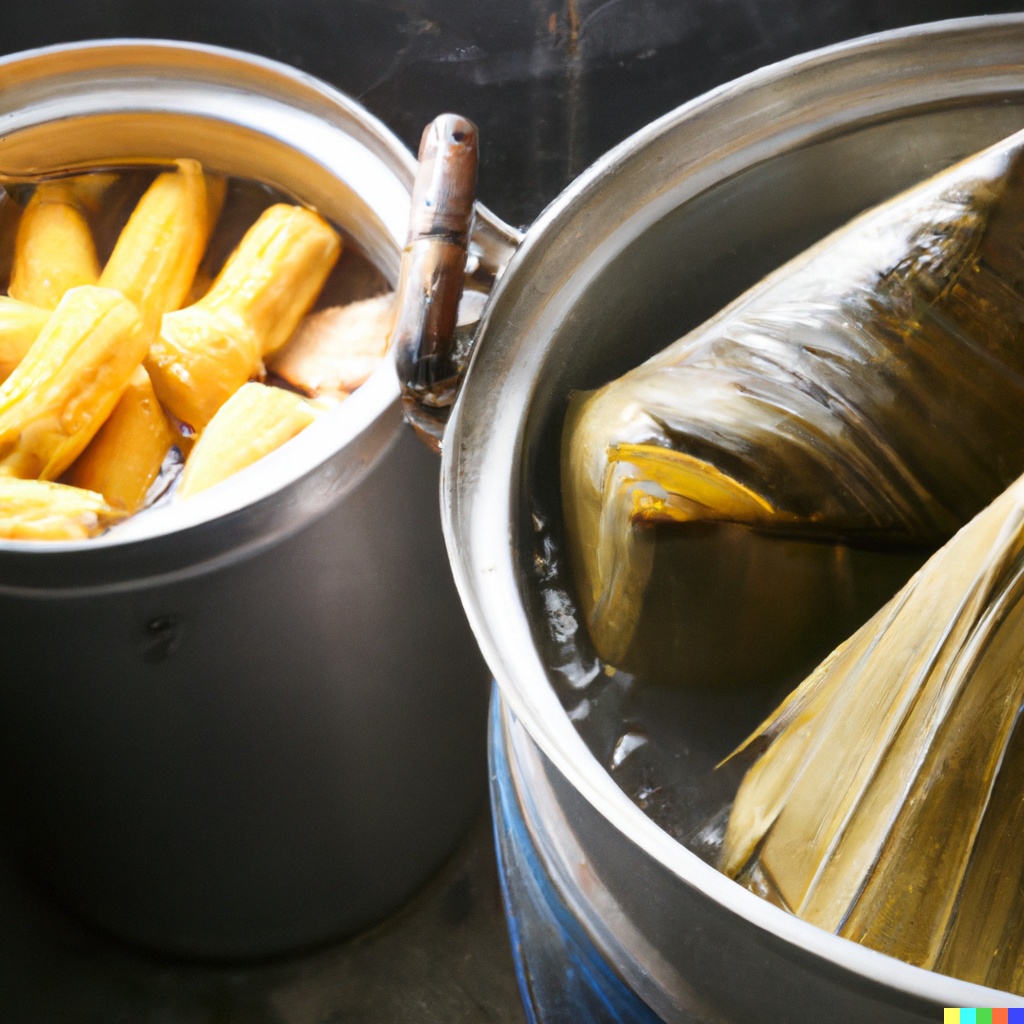}} &
{\includegraphics[width=0.18\linewidth]{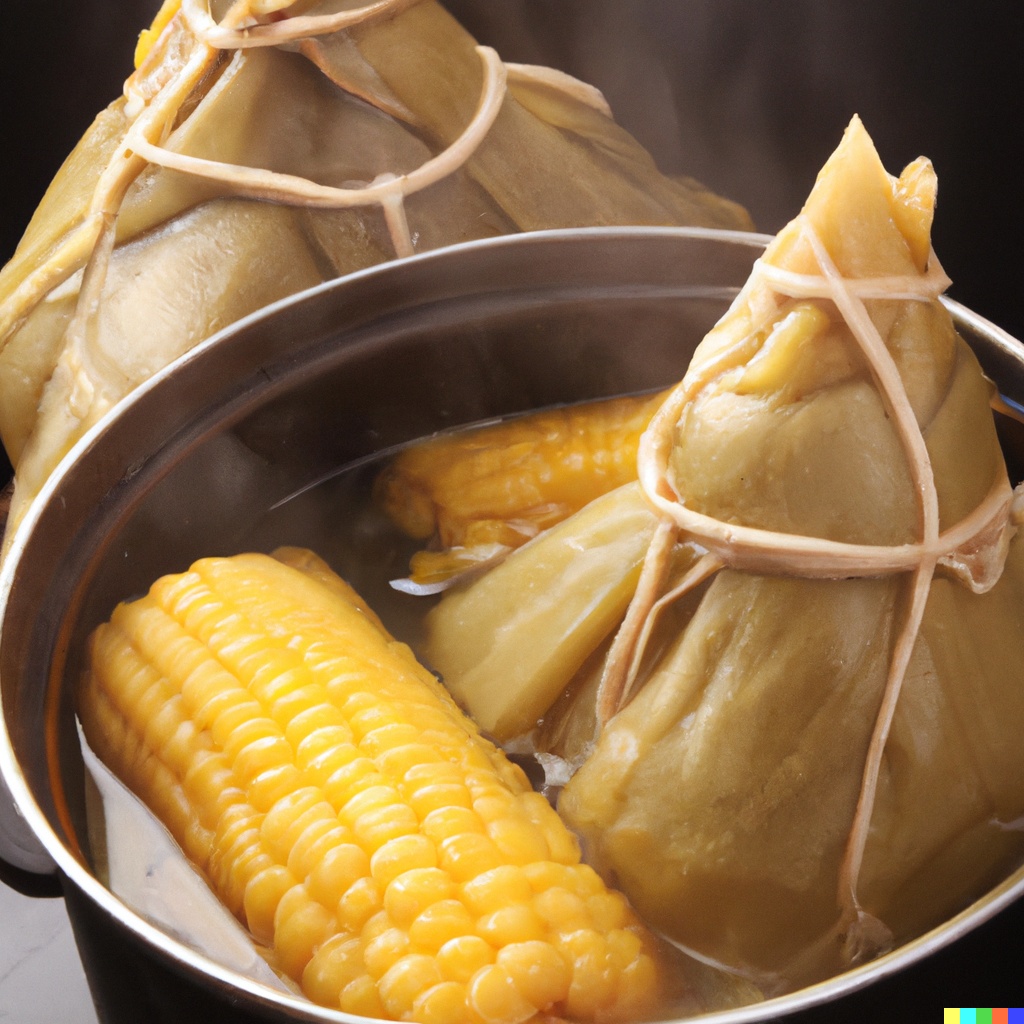}} &
{\includegraphics[width=0.18\linewidth]{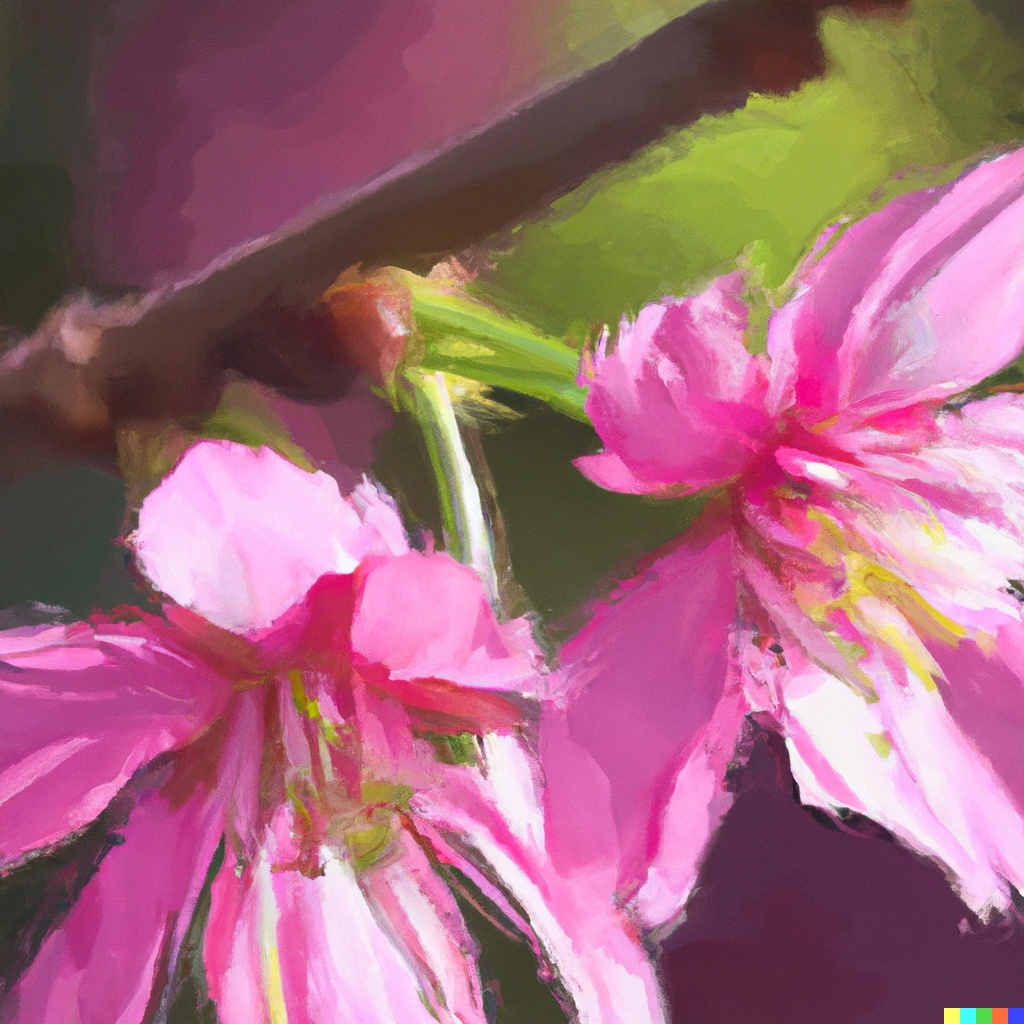}} &
{\includegraphics[width=0.18\linewidth]{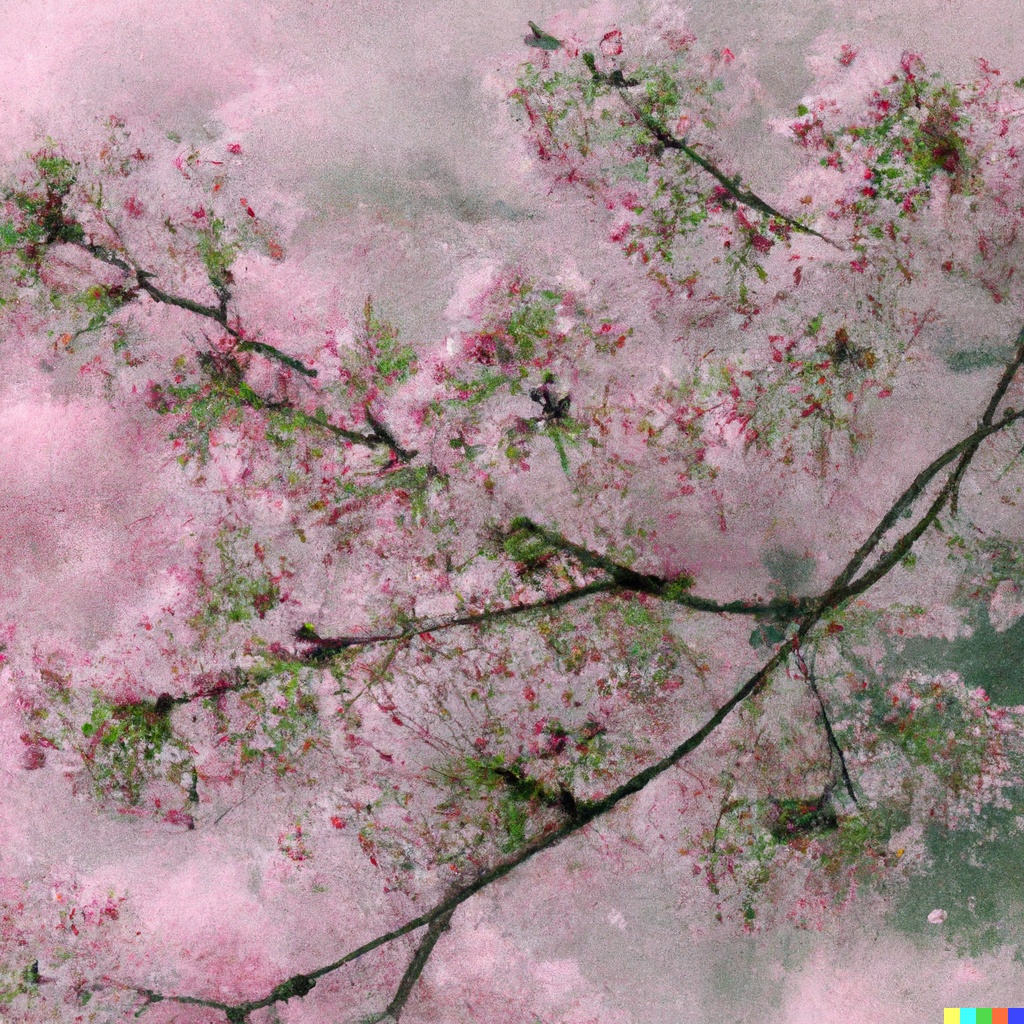}} \\
\multicolumn{4}{c}{\scriptsize Stable Diffusion} \\
{\includegraphics[width=0.18\linewidth]{figs/comp_case/00190/00190_0_sd.jpg}} &
{\includegraphics[width=0.18\linewidth]{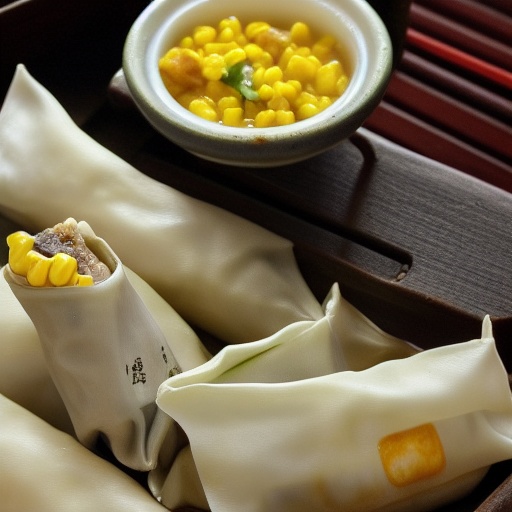}} &
{\includegraphics[width=0.18\linewidth]{figs/comp_case/00276/00276_0_sd.jpg}} &
{\includegraphics[width=0.18\linewidth]{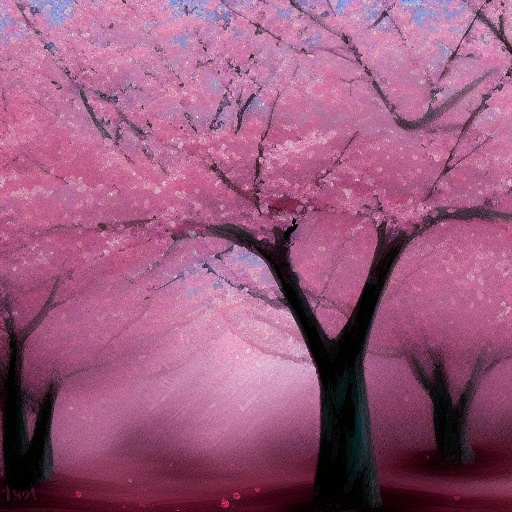}} \\
{\includegraphics[width=0.18\linewidth]{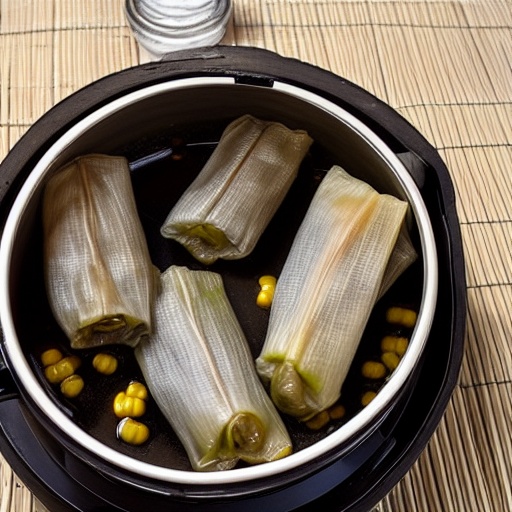}} &
{\includegraphics[width=0.18\linewidth]{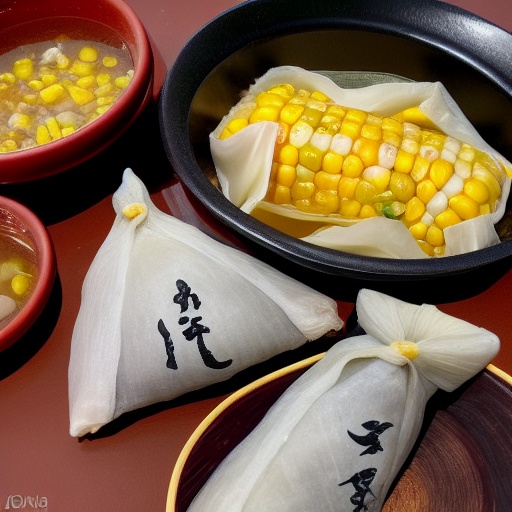}} &
{\includegraphics[width=0.18\linewidth]{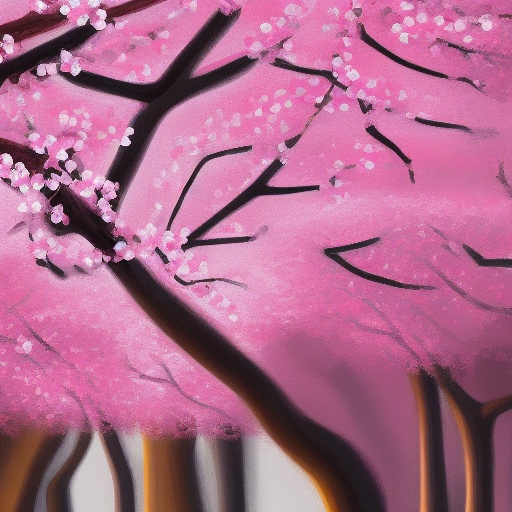}} &
{\includegraphics[width=0.18\linewidth]{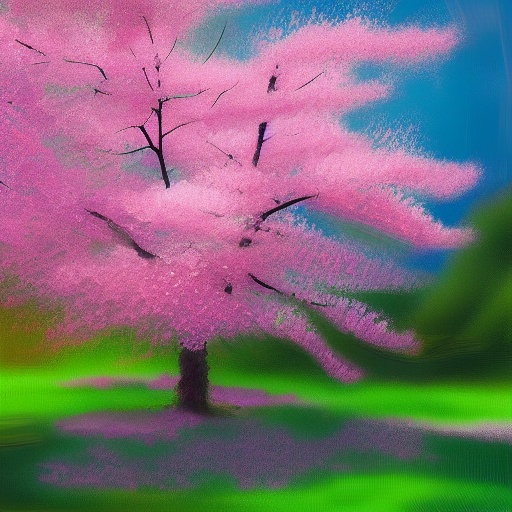}} \\
\end{tabular}
\caption{Example qualitative comparisons between ERNIE-ViLG~2.0 and DALL-E~2/Stable Diffusion on ERNIE-ViLG prompts from ViLG-300.}
\label{fig:case_ernie_vilg}
\end{figure*}

Figure~\ref{fig:human_detail} shows the detailed performance comparison between ERNIE-ViLG~2.0 and DALL-E~2/Stable Diffusion on ViLG-300, and example qualitative comparisons are shown in Figure~\ref{fig:case_drawbench} and~\ref{fig:case_ernie_vilg}.
The most important conclusion is that ERNIE-ViLG~2.0 is quite skilled in dealing with text prompts with colors and complex scenes, and also has impressive performance in many categories, such as \verb|Geography|, \verb|Scene|, and \verb|Cartoon|.
Intuitively, we attribute the excellent performance to the knowledge injection that endows the model with the ability to perceive and understand various named entities and detailed descriptions, as well as the increase in the number of parameters brought by the mixture-of-denoising-experts strategies also makes the model even more powerful.
At the same time, we also propose that further understanding of the number of objects and the relationship between them can be the focus of future text-to-image models.

\begin{figure}[t]
    \centering
        \includegraphics[width=\linewidth]{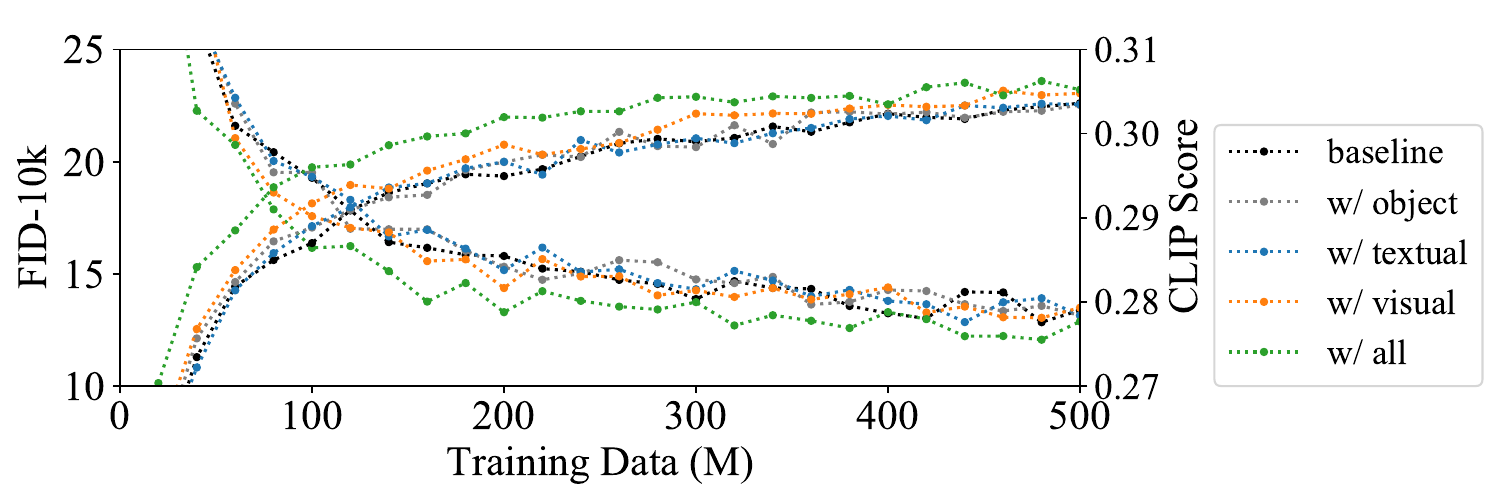}
    \caption{The convergence curves of various models with knowledge enhancement strategies. We choose guidance scale 3 and 8 to draw the curves of FID-10k and CLIP Score, respectively.}
    \label{fig:ke_conv}
\end{figure}

\begin{figure}[t]
    \centering
        \includegraphics[width=0.65\linewidth]{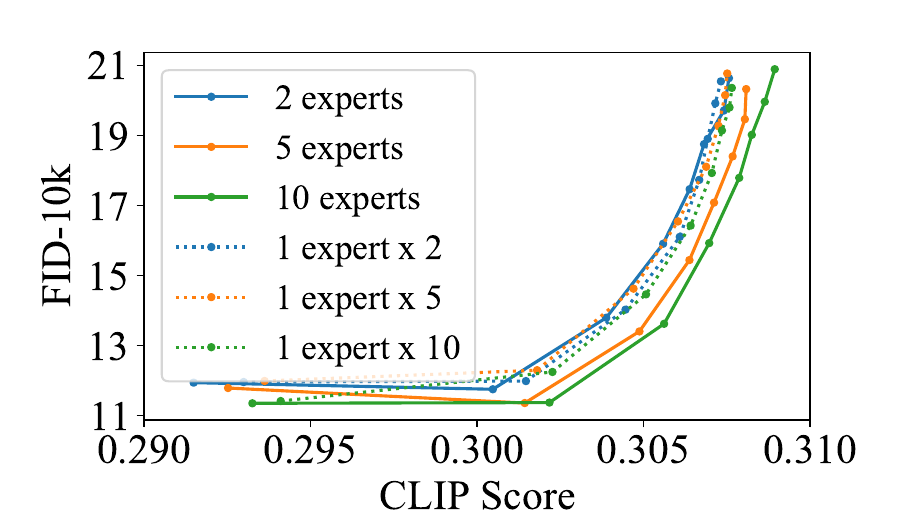}
    \caption{The performance comparison of different amount of denoising experts and training samples, and models see the same number of training samples at comparing points.}
    \label{fig:moe_eval_2}
\end{figure}

\begin{CJK*}{UTF8}{gbsn}
    \begin{figure*}[t]
        \centering
        \setlength{\tabcolsep}{6pt}
        \begin{tabular}{ccc}
            \rotatebox{90}{\scriptsize\phantom{AAAA.} Stable Diffusion} &
            \includegraphics[width=0.36\linewidth]{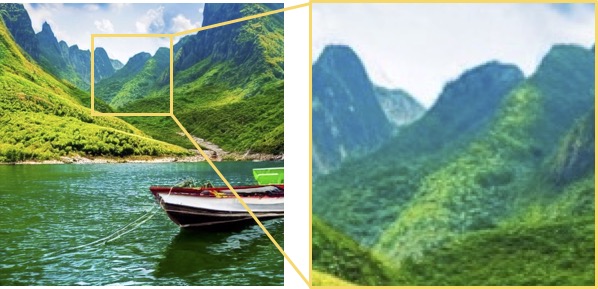} &
            \includegraphics[width=0.55\linewidth]{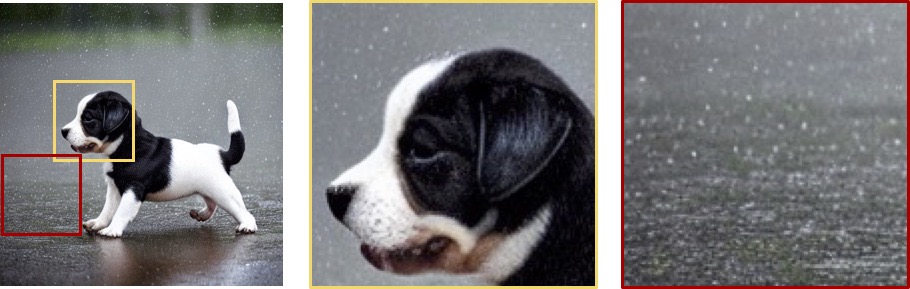} \\
            
            \rotatebox{90}{\scriptsize\phantom{AAAAA} DALLE-2} &
            \includegraphics[width=0.36\linewidth]{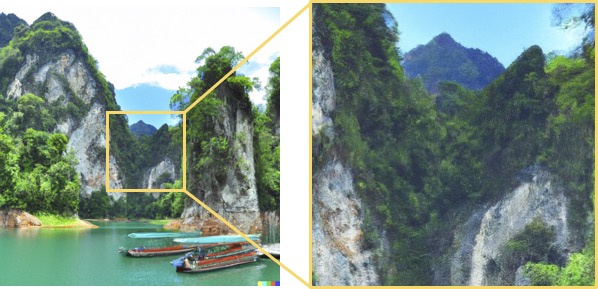} &
            \includegraphics[width=0.55\linewidth]{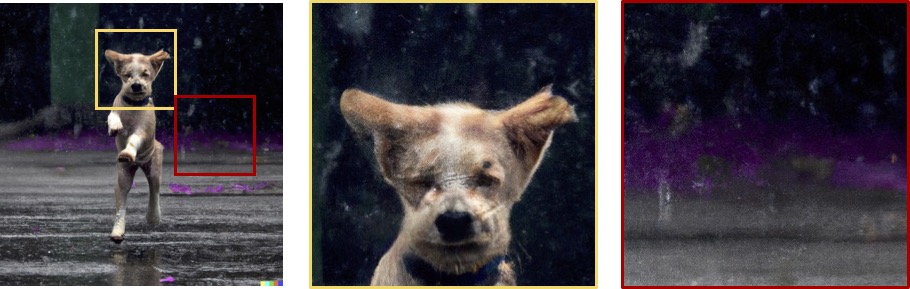} \\
            
            \rotatebox{90}{\scriptsize\phantom{AAA} ERNIE-ViLG~2.0} &
            \includegraphics[width=0.36\linewidth]{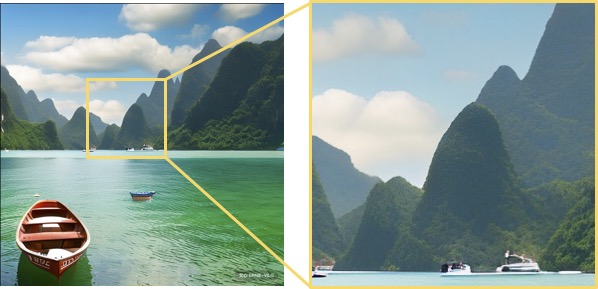} &
            \includegraphics[width=0.55\linewidth]{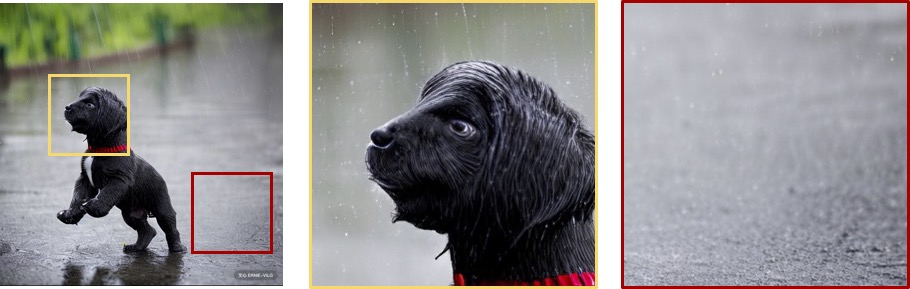} \\
    
            & \scriptsize \makecell{青山\ 绿水\ 小船\ 风景}
            & \scriptsize \makecell{雨中跳舞的小狗} \\
            & \scriptsize \makecell{Green mountains, green waters, boats, scenery}
            & \scriptsize \makecell{A puppy dancing in the rain}
        \end{tabular}
        \caption{Comparison of image quality by magnifying parts of generated images. ERNIE-ViLG~2.0 enables the generation of sharper 1024$\times$1024 images with more natural details.}
        \label{fig:resolution_case}
    \end{figure*}
\end{CJK*}

\section{Detailed Ablation Study}\label{appx:ablation}

\begin{figure*}[t]
    \centering
    \setlength{\tabcolsep}{1.5pt}
    \begin{tabular}{cccccccccc}
        \rotatebox{90}{\scriptsize\phantom{AA.} baseline} &
        \includegraphics[width=0.115\linewidth]{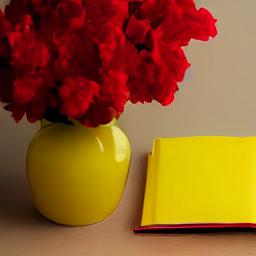} &
        \includegraphics[width=0.115\linewidth]{figs/ke-case/00017_b.jpeg} &
        \includegraphics[width=0.115\linewidth]{figs/ke-case/00027_b.jpeg} &
        \includegraphics[width=0.115\linewidth]{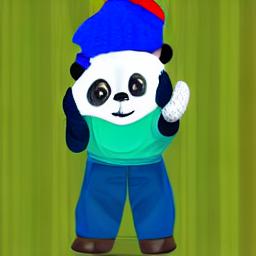} &
        \includegraphics[width=0.115\linewidth]{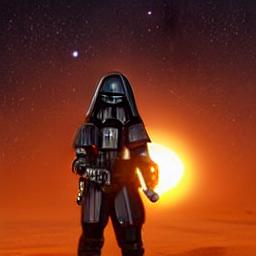} &
        \includegraphics[width=0.115\linewidth]{figs/ke-case/00205_b.jpeg} &
        \includegraphics[width=0.115\linewidth]{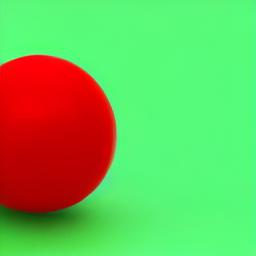} &
        \includegraphics[width=0.115\linewidth]{figs/ke-case/00231_b.jpeg} \\
        
        \rotatebox{90}{\scriptsize\phantom{AA.} w/ textual} &
        \includegraphics[width=0.115\linewidth]{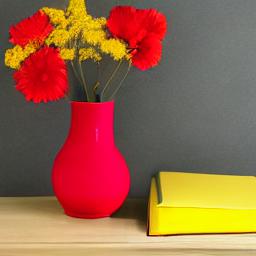} &
        \includegraphics[width=0.115\linewidth]{figs/ke-case/00017_t.jpeg} &
        \includegraphics[width=0.115\linewidth]{figs/ke-case/00027_t.jpeg} &
        \includegraphics[width=0.115\linewidth]{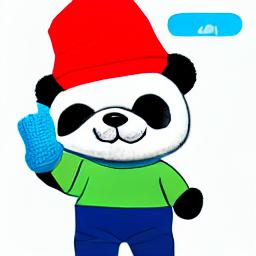} &
        \includegraphics[width=0.115\linewidth]{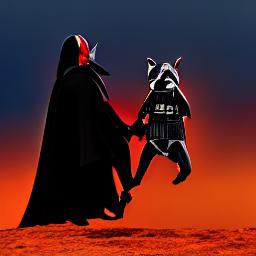} &
        \includegraphics[width=0.115\linewidth]{figs/ke-case/00205_t.jpeg} &
        \includegraphics[width=0.115\linewidth]{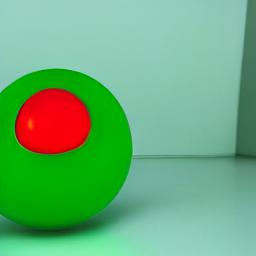} &
        \includegraphics[width=0.115\linewidth]{figs/ke-case/00231_t.jpeg} \\
        
        \rotatebox{90}{\scriptsize\phantom{AA.} w/ visual} &
        \includegraphics[width=0.115\linewidth]{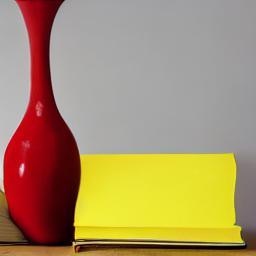} &
        \includegraphics[width=0.115\linewidth]{figs/ke-case/00017_v.jpeg} &
        \includegraphics[width=0.115\linewidth]{figs/ke-case/00027_v.jpeg} &
        \includegraphics[width=0.115\linewidth]{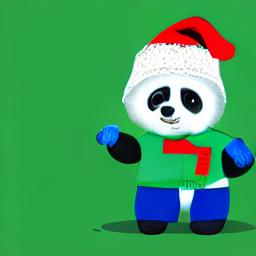} &
        \includegraphics[width=0.115\linewidth]{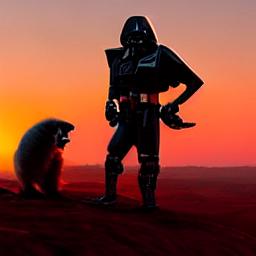} &
        \includegraphics[width=0.115\linewidth]{figs/ke-case/00205_v.jpeg} &
        \includegraphics[width=0.115\linewidth]{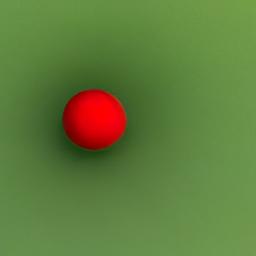} &
        \includegraphics[width=0.115\linewidth]{figs/ke-case/00231_v.jpeg} \\
        
        \rotatebox{90}{\scriptsize\phantom{AAA} w/ all} &
        \includegraphics[width=0.115\linewidth]{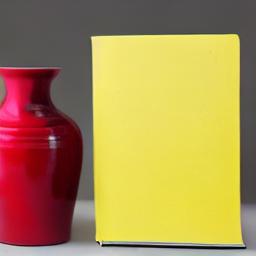} &
        \includegraphics[width=0.115\linewidth]{figs/ke-case/00017_a.jpeg} &
        \includegraphics[width=0.115\linewidth]{figs/ke-case/00027_a.jpeg} &
        \includegraphics[width=0.115\linewidth]{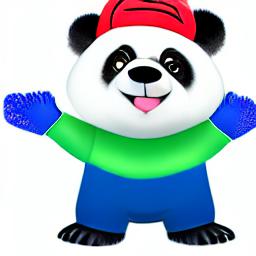} &
        \includegraphics[width=0.115\linewidth]{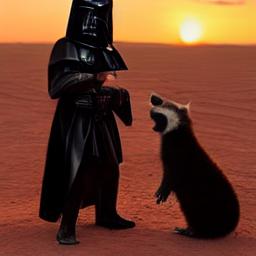} &
        \includegraphics[width=0.115\linewidth]{figs/ke-case/00205_a.jpeg} &
        \includegraphics[width=0.115\linewidth]{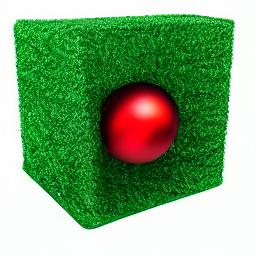} &
        \includegraphics[width=0.115\linewidth]{figs/ke-case/00231_a.jpeg} \\

        & \scriptsize \makecell{一本黄色的书 \\ 和一个红色的 \\ 花瓶}
        & \scriptsize \makecell{一辆白色的汽 \\ 车和一只红色 \\ 的羊}
        & \scriptsize \makecell{制作拿铁拉花 \\ 的熊猫}
        & \scriptsize \makecell{熊猫宝宝戴着 \\ 红帽子、蓝手 \\ 套，穿着绿衬 \\ 衫和蓝裤子的 \\ 表情符号}
        & \scriptsize \makecell{日落时分，达 \\ 斯·维德在火 \\ 星上与浣熊 \\ 玩耍 }
        & \scriptsize \makecell{客厅的沙发前 \\ 有一张木质茶 \\ 几}
        & \scriptsize \makecell{一个小红球在 \\ 一个大绿块 \\ 内部}
        & \scriptsize \makecell{身上着火的 \\ 鱼} \\ 
        & \scriptsize \makecell{A yellow book \\ and a red vase.}
        & \scriptsize \makecell{A white car and \\ a red sheep.}
        & \scriptsize \makecell{A panda making \\ latte art.}
        & \scriptsize \makecell{An emoji of a \\ baby panda \\ wearing \\ a red hat, \\ blue gloves, \\  green shirt, \\ and blue pants.}
        & \scriptsize \makecell{Darth Vader \\ playing with \\ raccoon in Mars \\ during sunset.}
        & \scriptsize \makecell{There is a \\ wooden tea \\ table in front \\ of the sofa \\ in the \\ living room}
        & \scriptsize \makecell{A small red ball \\ in a large \\ green block}
        & \scriptsize \makecell{A burning \\ fish}
    \end{tabular}
    \caption{Samples from ViLG-300 with different knowledge enhancement strategies. It can be found that the impacts of textual and visual knowledge do not seem to overlap, and the combination of them is an effective solution to facilitate accurate semantic control and high image fidelity.}
    \label{fig:ke_case}
\end{figure*}

\begin{figure*}[t]
    \centering
    \setlength{\tabcolsep}{1.5pt}
    \begin{tabular}{cccccccccc}
        \rotatebox{90}{\scriptsize\phantom{AAA} 1 expert} &
        \includegraphics[width=0.115\linewidth]{figs/mode-case/00057_1.jpg} &
        \includegraphics[width=0.115\linewidth]{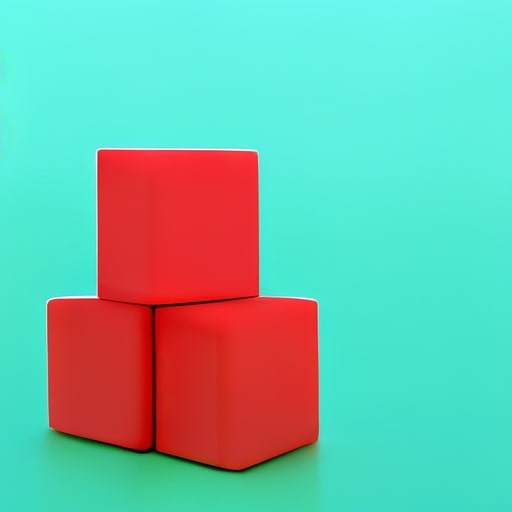} &
        \includegraphics[width=0.115\linewidth]{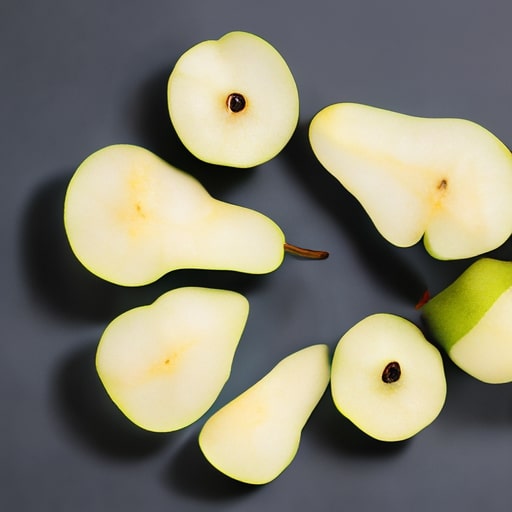} &
        \includegraphics[width=0.115\linewidth]{figs/mode-case/00100_1.jpg} &
        \includegraphics[width=0.115\linewidth]{figs/mode-case/00152_1.jpg} &
        \includegraphics[width=0.115\linewidth]{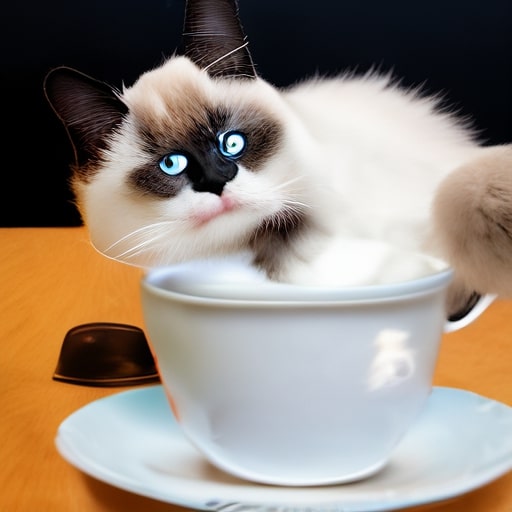} &
        \includegraphics[width=0.115\linewidth]{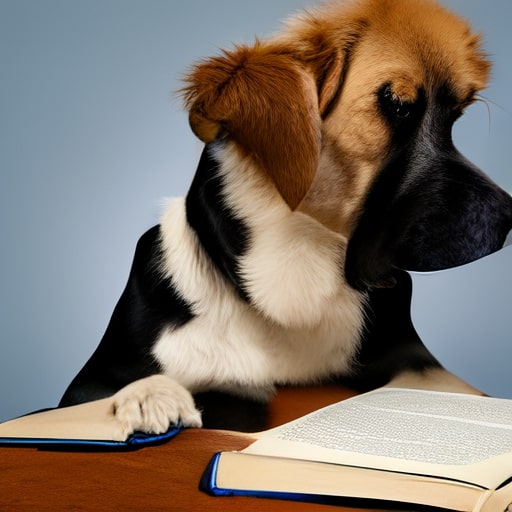} &
        \includegraphics[width=0.115\linewidth]{figs/mode-case/00237_1.jpg} \\
        
        \rotatebox{90}{\scriptsize\phantom{AA.} 2 experts} &
        \includegraphics[width=0.115\linewidth]{figs/mode-case/00057_2.jpg} &
        \includegraphics[width=0.115\linewidth]{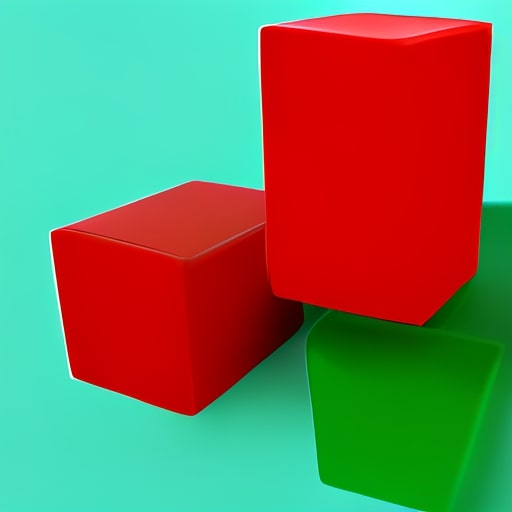} &
        \includegraphics[width=0.115\linewidth]{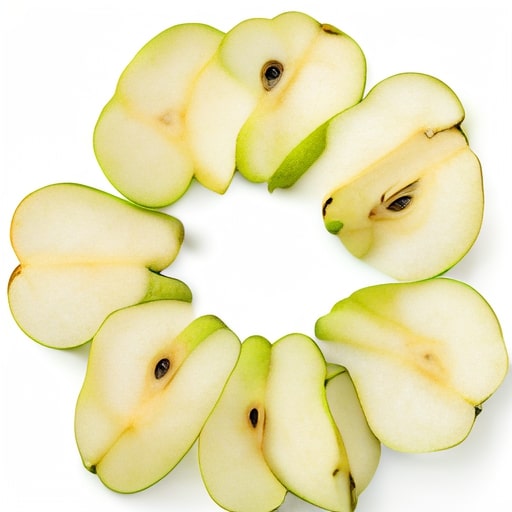} &
        \includegraphics[width=0.115\linewidth]{figs/mode-case/00100_2.jpg} &
        \includegraphics[width=0.115\linewidth]{figs/mode-case/00152_2.jpg} &
        \includegraphics[width=0.115\linewidth]{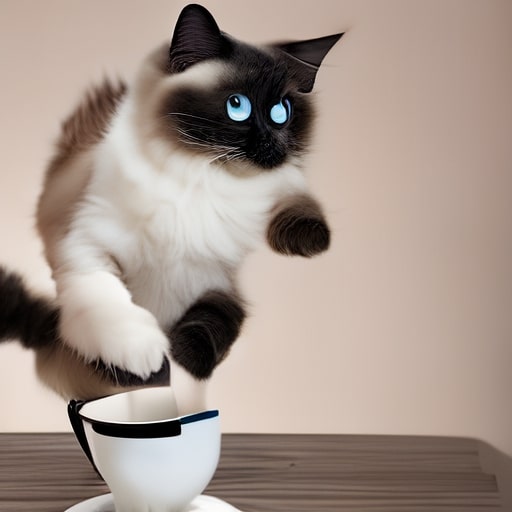} &
        \includegraphics[width=0.115\linewidth]{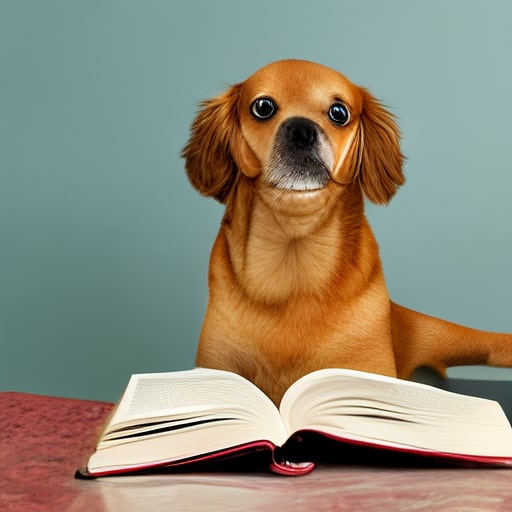} &
        \includegraphics[width=0.115\linewidth]{figs/mode-case/00237_2.jpg} \\
        
        \rotatebox{90}{\scriptsize\phantom{AA.} 5 experts} &
        \includegraphics[width=0.115\linewidth]{figs/mode-case/00057_5.jpg} &
        \includegraphics[width=0.115\linewidth]{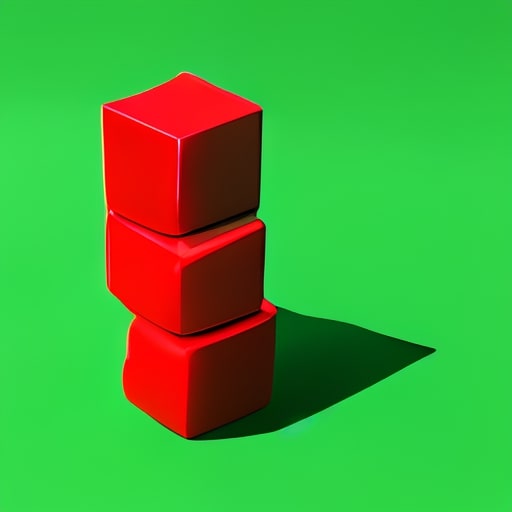} &
        \includegraphics[width=0.115\linewidth]{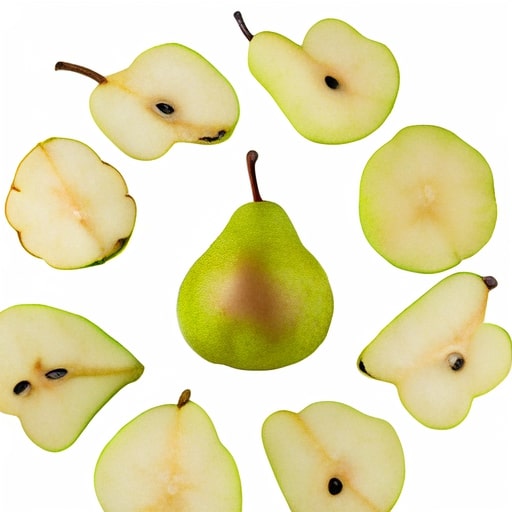} &
        \includegraphics[width=0.115\linewidth]{figs/mode-case/00100_5.jpg} &
        \includegraphics[width=0.115\linewidth]{figs/mode-case/00152_5.jpg} &
        \includegraphics[width=0.115\linewidth]{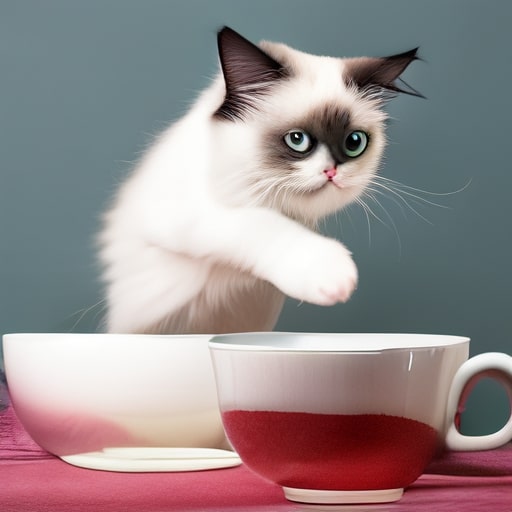} &
        \includegraphics[width=0.115\linewidth]{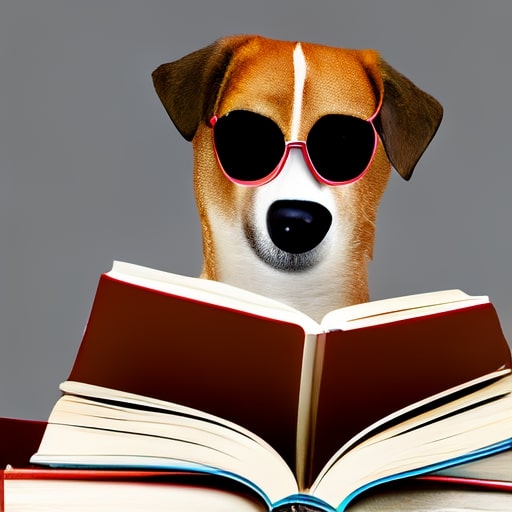} &
        \includegraphics[width=0.115\linewidth]{figs/mode-case/00237_5.jpg} \\
        
        \rotatebox{90}{\scriptsize\phantom{AA} 10 experts} &
        \includegraphics[width=0.115\linewidth]{figs/mode-case/00057_10.jpg} &
        \includegraphics[width=0.115\linewidth]{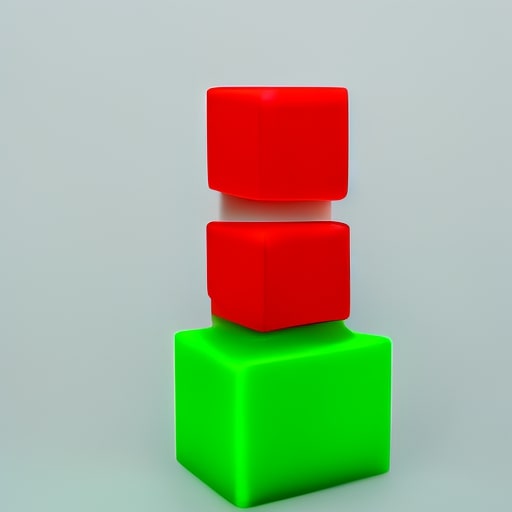} &
        \includegraphics[width=0.115\linewidth]{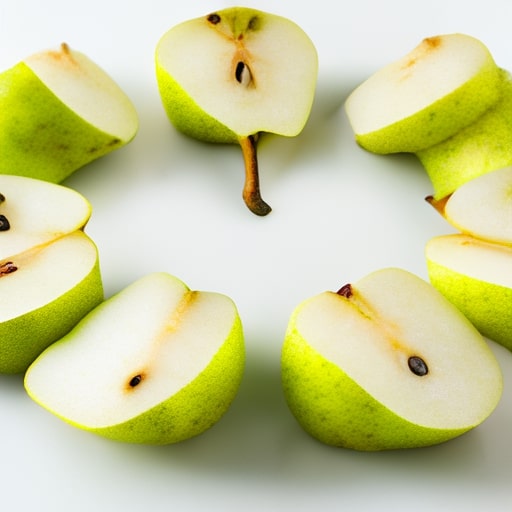} &
        \includegraphics[width=0.115\linewidth]{figs/mode-case/00100_10.jpg} &
        \includegraphics[width=0.115\linewidth]{figs/mode-case/00152_10.jpg} &
        \includegraphics[width=0.115\linewidth]{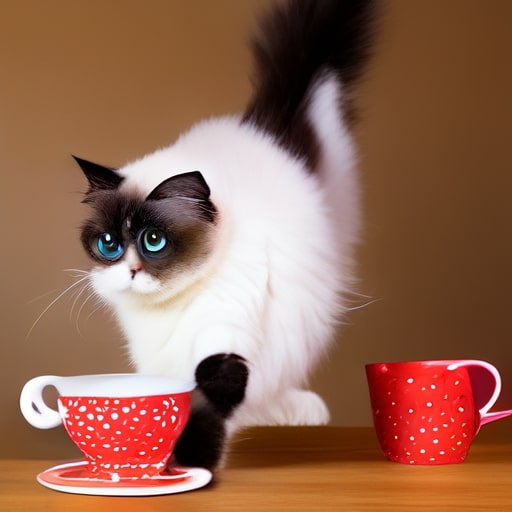} &
        \includegraphics[width=0.115\linewidth]{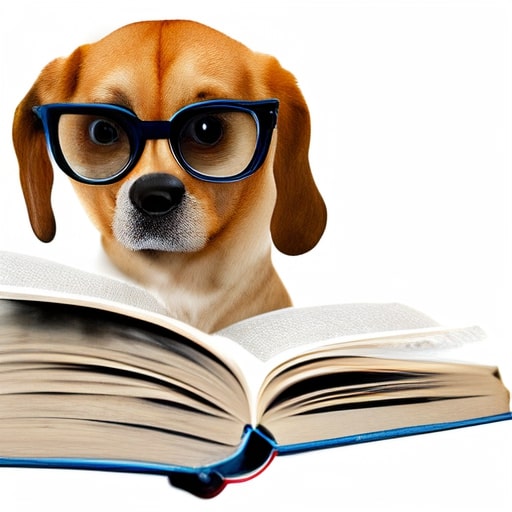} &
        \includegraphics[width=0.115\linewidth]{figs/mode-case/00237_10.jpg} \\
        
        & \scriptsize \makecell{桌子上放着一 \\ 只闹钟}
        & \scriptsize \makecell{三个立方体的 \\ 堆叠。红色在 \\ 上，红色在 \\ 中，绿色在下}
        & \scriptsize \makecell{切成七块排列 \\ 成一个环的梨}
        & \scriptsize \makecell{勺子上的伞}
        & \scriptsize \makecell{穿西服的狼}
        & \scriptsize \makecell{跳上桌子玩杯 \\ 子的布偶猫}
        & \scriptsize \makecell{一只狗在看书，\\ 很厚的书}
        & \scriptsize \makecell{狐狸脑袋的 \\ 兔子} \\ 
        & \scriptsize \makecell{A single clock \\ is sitting \\ on a table}
        & \scriptsize \makecell{A stack of 3 \\ cubes. The red \\ is on the top,  \\ The red is \\ in the middle, \\ The green is \\ on the bottom}
        & \scriptsize \makecell{A pear cut \\ into seven \\ pieces arranged \\ in a ring}
        & \scriptsize \makecell{An umbrella \\ on top of \\ a spoon}
        & \scriptsize \makecell{Wolf in \\ a suit}
        & \scriptsize \makecell{A puppet cat \\ jumping on \\ the table to \\ play with cups}
        & \scriptsize \makecell{A dog is \\ reading a \\ thick book}
        & \scriptsize \makecell{A rabbit \\ with a \\ fox's head}
    \end{tabular}
    \caption{Samples from ViLG-300 with different number of denoising experts. When increasing the experts, the most noticeable evolution is that the texture of generated image becomes more natural and photorealistic. Limited by the layout, we simplify the prompt in the second column, and the input received by model actually is ``三个立方体堆叠。一个红色立方体在顶部，放在一个红色立方体上。这个红色立方体在中间，放在一个绿色立方体上。这个绿色立方体在底部。(A stack of 3 cubes. A red cube is on the top, sitting on a red cube. The red cube is in the middle, sitting on a green cube. The green cube is on the bottom.)''.}
    \label{fig:mode_case}
\end{figure*}

Here we attach more analysis to the ablation study and more showcases in Figure~\ref{fig:ke_case},\ref{fig:mode_case}.

\subsection{Knowledge Enhancement Ablation }\label{appx:ke_ablation}

Figure~\ref{fig:ke_conv} provides the convergence curves of various models, it is obvious that the knowledge enhancement strategies significantly accelerate the convergence process of diffusion models.
Notably, at the very beginning of training, the knowledge-enhanced model reaches or even exceeds the performance that the baseline model with two times of training samples (i.e., 100M v.s. 200M, 200M v.s. 400M).

To quantitatively measure the improvement brought by each knowledge source, we calculate the CLIP score between ViLG-300 prompts and generated images\footnote{We feed ERNIE-ViLG~2.0 with Chinese prompts and calculate the CLIP score between generated images and corresponding English prompts.}. Table~\ref{tab:vilg300-clip} presents the top five categories with maximum performance gain of each strategy against baseline. It can be found that different strategies result in improvements in different categories, indicating that they help model absorb knowledge and improve text-image alignment in complementary aspects.
In addition, we also notice that CLIP could not well capture the relationships between multiple objects (e.g., counterfactual), so we leave the accurate automatic evaluation method for fine-grained semantic control as a valuable future work.

\subsection{Mixture-of-Denoising-Experts Ablation}\label{appx:moe_ablation}

To explore the impact of training samples or model size (i.e., the number of denoising experts) on performance, we train the setting of \verb|1| \verb|expert| with 400M/1B/2B samples following Section~\ref{sec:MoDE-ablation}, which aligns with the number of samples trained by \verb|2/5/10| \verb|experts|.
Figure~\ref{fig:moe_eval_2} shows that the performance of \verb|1| \verb|expert| with 400M and \verb|2| \verb|experts| with 200M each is basically equal, while the performance of \verb|1| \verb|expert| lags behind that of \verb|2| \verb|experts| as training goes on.
This shows that decoupling the denoising capability of different stages is an effective strategy, and reasonably scaling the size of U-Net is able to further boost the performance of text-to-image model.

\subsection{Comparison of Image Quality}\label{appx:definition}
Figure~\ref{fig:resolution_case} compares the image details of ERNIE-ViLG~2.0 and baseline models by zooming in small regions of generated images.
Technically, both ERNIE-ViLG~2.0 and Stable Diffusion generate image latent representation with diffusion models conditioned on text. While Stable Diffusion only produces 512$\times$512 sized images, ERNIE-ViLG~2.0 could directly output images with 1024$\times$1024 resolution. Therefore, the magnified parts of ERNIE-ViLG~2.0 are clearer than those of Stable Diffusion.
As for DALL-E~2, it employs cascaded generation by first producing 64$\times$64 images with text and then scaling it up to 1024$\times$1024 resolution with two super-resolution models.
Although it generates images of the same resolution as ERNIE-ViLG~2.0, the output of DALL-E~2 sometimes contains unnatural textures, such as fluffy trees and rain drops in the magnified regions.
Contrary to DALL-E~2, the textures of our model's outcome are more natural and photorealistic.

\begin{CJK*}{UTF8}{gbsn}
\end{CJK*}

\begin{CJK*}{UTF8}{gbsn}
\end{CJK*}

\begin{CJK*}{UTF8}{gbsn}
\end{CJK*}

\begin{CJK*}{UTF8}{gbsn}
\end{CJK*}

\end{document}